%% file: main.tex
\documentclass[a4paper,fleqn]{cas-sc}

\usepackage[numbers]{natbib}
\usepackage{amsmath}
\usepackage{amsthm}
\newtheorem{definition}{Definition}
\usepackage{amsfonts}
\usepackage{amssymb}
\usepackage{float}
\usepackage{algorithm}
\usepackage{algorithmic}
\usepackage[tight,footnotesize]{subfigure}
\usepackage{multirow}
\usepackage{adjustbox}
\usepackage{booktabs}
\usepackage{appendix}
\usepackage{placeins}

\usepackage{color, colortbl}
\definecolor{Gray}{gray}{0.9}

\usepackage{xcolor}

\newcommand{\figref}[1]{Figure~\ref{#1}}
\renewcommand{\tabref}[1]{Table~\ref{#1}}
\newcommand{\obs}[0]{o}

\newcommand{\state}[0]{s}
\newcommand{\State}[0]{S}
\newcommand{\action}[0]{u}
\newcommand{\Action}[0]{U}
\newcommand{\admissibleset}[0]{S_{\varphi}}
\newcommand{\safeset}[0]{S_{h}}

\def\tsc#1{\csdef{#1}{\textsc{\lowercase{#1}}\xspace}}
\tsc{WGM}
\tsc{QE}

\begin{document}
\let\WriteBookmarks\relax
\def\floatpagepagefraction{1}
\def\textpagefraction{.001}

\shorttitle{How RTA impacts RL Training and Performance}    

\shortauthors{Hamilton, Dunlap, Johnson, Hobbs}  

\title[mode = title]{Ablation Study of How Run Time Assurance Impacts the Training and Performance of Reinforcement Learning Agents}  




\affiliation[inst1]{organization={Vanderbilt University Electrical and Computer Engineering},
            addressline={2201 West End Ave}, 
            city={Nashville},
            postcode={37235}, 
            state={TN},
            country={USA}}

\affiliation[inst2]{organization={Parallax Advanced Research},
            addressline={4035 Colonel Glenn Hwy}, 
            city={Beavercreek},
            postcode={45431}, 
            state={OH},
            country={USA}}
            
\affiliation[inst3]{organization={Autonomy Capability Team (ACT3) at the Air Force Research Laboratory},
            city={Wright-Patterson AFB},
            postcode={45433}, 
            state={OH},
            country={USA}}


\author[inst1]{Nathaniel Hamilton}[orcid=0000-0002-7147-1964]
\cormark[1]


\ead{nathaniel_hamilton@outlook.com}

\ead[url]{http://nphamilton.github.io/}

\author[inst2]{Kyle Dunlap}[orcid=0000-0002-5869-1380]
\author[inst1]{Taylor T Johnson}[orcid=0000-0001-8021-9923]
\author[inst3]{Kerianne L Hobbs}[orcid=0000-0001-5215-2231]


\cortext[1]{Corresponding author}

\begin{abstract}
Reinforcement Learning (RL) has become an increasingly important research area as the success of machine learning algorithms and methods grows. To combat the safety concerns surrounding the freedom given to RL agents while training, there has been an increase in work concerning \emph{Safe Reinforcement Learning} (SRL). However, these new and safe methods have been held to less scrutiny than their unsafe counterparts. For instance, comparisons among safe methods often lack fair  evaluation across similar initial condition bounds and hyperparameter settings, use poor evaluation metrics, and cherry-pick the best training runs rather than averaging over multiple random seeds. In this work, we conduct an \emph{ablation study} using evaluation best practices to investigate the impact of \emph{run time assurance} (RTA), which monitors the system state and intervenes to assure safety, on effective learning. By studying multiple RTA approaches in both on-policy and off-policy RL algorithms, we seek to understand which RTA methods are most effective, whether the agents become dependent on the RTA, and the importance of reward shaping versus safe exploration in RL agent training. Our conclusions shed light on the most promising directions of SRL, and our evaluation methodology lays the groundwork for creating better comparisons in future SRL work.
\end{abstract}

\begin{highlights}
    \item Training RL agents with run time assurance always runs the risk of forming dependence 
    \item Explicit simplex run time assurance approaches are the most conducive for learning 
    \item On-policy RL algorithms see a greater benefit from training with run time assurance 
    \item A well-defined reward function is imperative for training successful RL agents 
\end{highlights}


\maketitle



\section{Introduction}

Reinforcement Learning (RL) and Deep Reinforcement Learning (DRL) are fast-growing fields with growing impact, spurred by success in agents that learn to beat human experts in games like Go \cite{silver2016mastering} and Starcraft \cite{starcraft2019}. 
However, these successes are predominantly limited to virtual environments. 
An RL agent learns a behavior policy that is optimized according a reward function. The policy is learned through interacting with/in the environment, making training on real-world hardware platforms prohibitively expensive and time-consuming. Additionally, RL allows agents to learn via trial and error, exploring \emph{any behavior} during the learning process. In many cyber-physical domains, this level of freedom is unacceptable. Consider the example of an industrial robot arm learning to place objects in a factory. Some behaviors could cause the robot to damage itself, other elements in the factory, or nearby workers. To mitigate these set-backs, most RL training is performed in a simulated environment. After the training is completed in simulation, the learned policy can then be transferred to the real world via a \emph{sim2real} transfer. However, this \textit{sim2real} transfer can result in poor performance and undesirable behavior \cite{jang2019simulation, hamilton2022zero}.

To counteract these issues, the field of \emph{Safe Reinforcement Learning} (SRL) has grown. Recent works demonstrate real-world online learning \cite{fisac2018general}, optimal performance that does not require safety checking when deployed \cite{wagener2021safe}, and SRL approaches that work better than state-of-the-art DRL approaches \cite{jothimurugan2019composable}.
Each new SRL paper claims to be the best, safest, most efficient, or least restrictive approach, but few prove these claims with valid demonstrations. 
When we tried to replicate studies using original source code, we found inconsistencies in their comparisons that made the SRL problem easier to solve. When these inconsistencies were accounted for, almost all the improvements claimed by the authors to come from the SRL method disappeared. These inconsistencies include (1) manipulating the initial conditions for only the SRL agents, so the RL methods have to learn how to solve the problem from a greater range of initial conditions, (2) forcing the ``unsafe'' RL agents to learn how to recover from unrecoverable unsafe conditions\footnote{An example of an unrecoverable unsafe condition would include crashing the vehicle being controlled, while a recoverable unsafe condition might include violating a set speed limit. While it may be possible to recover from a crash in a low fidelity simulation, it likely is not possible in reality. Therefore, when a crash occurs, the simulation should be terminated instead of allowed to continue. Otherwise, the agents trained to continue after such an event have to learn the optimal policy for that continuation in addition to the original task.}, and (3) tuning hyperparameters for their SRL methods while leaving the regular RL methods hyperparameters at a baseline. Any one of these inconsistencies can lead to results skewed in the SRL method's favor. Furthermore, many demonstrations fail to repeat trials across multiple random seeds. Because RL is a stochastic process, showing results from one random seed is not representative of the true performance of the algorithm. Only presenting the results of a singular trial allows for results to be selectively chosen to show a large improvement over existing methods. The work in \cite{henderson2018deep, agarwal2021deep} highlights the importance of running experiments across at least 5 random seeds and averaging the results and showing the performance range in order to prove the trend of increased efficiency.

These issues in SRL publications bring rise to the need for better comparative studies and more \emph{ablation studies}. An ablation study involves singling out and removing individual components of a complex system to understand their impact on the system as a whole. Ablation studies are used to determine causality and can prove which aspects of a system are actually the most important. 
%
%
In this work, we outline a better standard for comparing SRL approaches as we conduct a thorough ablation study on SRL approaches that use \textit{Run Time Assurance} (RTA), an approach that monitors the output of the control policy for unsafe control actions and intervenes by modifying the output to assure system safety. RTA can be applied during training 
and after the training is complete.

\textbf{Our contributions.} This paper presents an in-depth investigation on how RTA configuration and usage choices impact RL training and final agent performance. The key contributions of this paper are as follows. 
\begin{enumerate}
    \item Evaluation across four different RTA approaches in addition to training with no RTA.
    \item Evaluation of five different RTA training configurations that adapt how penalties are assigned during training and whether the RL agent has knowledge of a corrected action.
    \item Evaluation of (1) and (2) on two different classes of RL: off-policy (SAC) and on-policy (PPO).
    \item Evaluation of the true performance of each combination by training across 10 random seeds and averaging the results. 
    \item A large-scale (880 unique agents trained) experimental ablation study that covers (1), (2), and (3). 
    \item Analysis of the experimental results to provide practical insights and recommendations for training RL agents with RTA. In particular, answering these important questions:
    \begin{enumerate}
        \item (\ref{sec:results_dependency}) Do agents learn to become dependent on RTA? 
        \item (\ref{sec:results_configuration}) Which RTA configuration is most effective?
        \item (\ref{sec:results_approach}) Which RTA approach is most effective?
        \item (\ref{sec:results_policy}) Which works better with RTA, off-policy (SAC) or on-policy (PPO)? 
        \item (\ref{sec:results_shaping}) Which is more important, Reward Shaping or Safe Exploration?
    \end{enumerate}
\end{enumerate}


\section{Deep Reinforcement Learning}

\begin{figure}[htpb] 
    \centering
    \includegraphics[width=0.6\columnwidth]{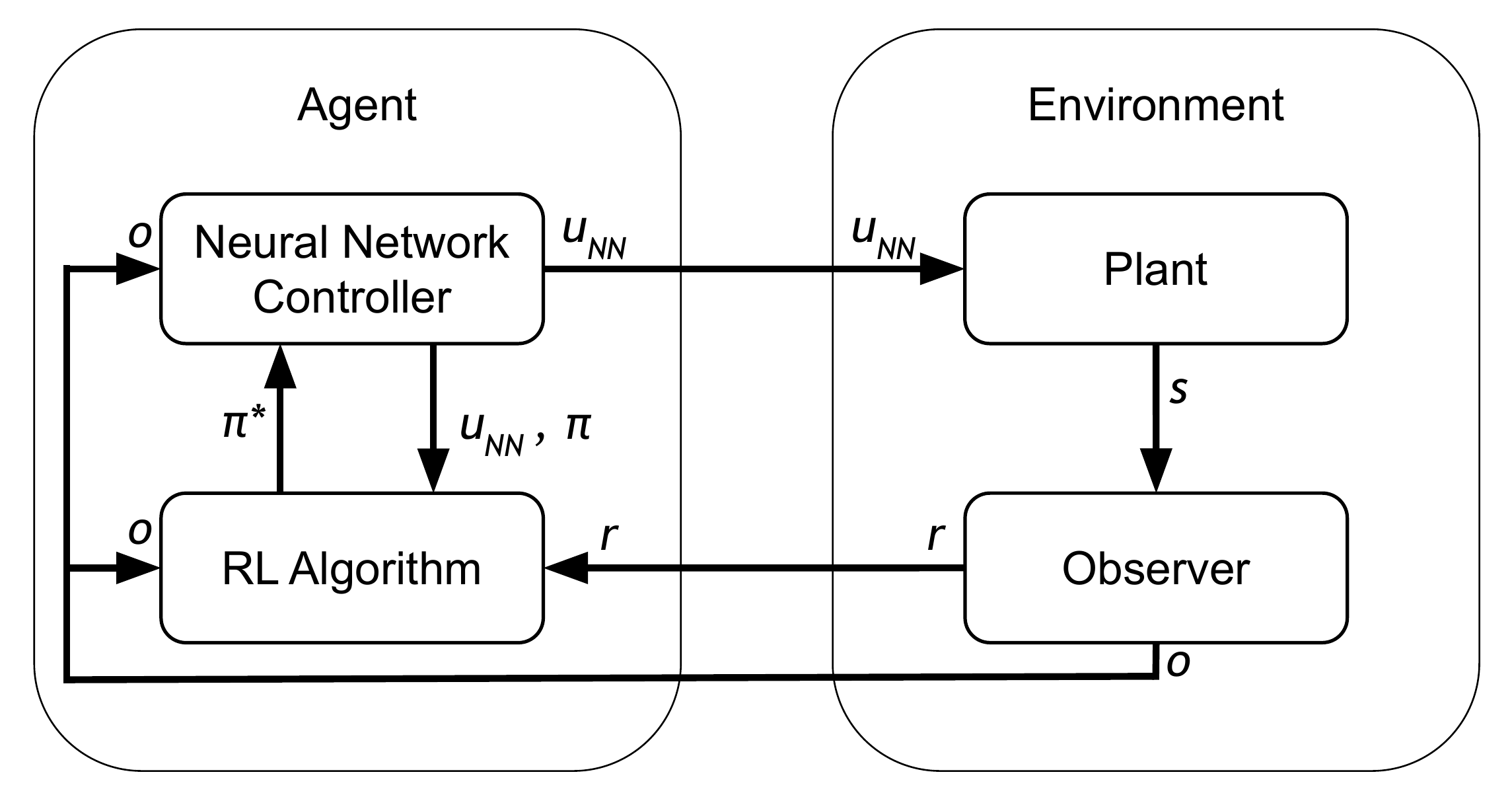}
    \caption{DRL training interactions between agent and environment without RTA}
    \label{fig:rta_off}
\end{figure}

\emph{Reinforcement Learning} (RL) is a form of machine learning in which an agent acts in an environment, learns through experience, and increases its performance based on rewarded behavior. \emph{Deep Reinforcement Learning} (DRL) is a newer branch of RL in which a neural network is used to approximate the behavior function, i.e. policy $\pi$. The basic construction of the DRL approach is shown in \figref{fig:rta_off}. The agent consists of the \emph{Neural Network Controller} (NNC) and RL algorithm, and the environment consists of a plant and observer model. The environment can be comprised of any dynamical system, from Atari simulations (\cite{hamilton2020sonic, alshiekh2018safe}) to complex robotics scenarios (\cite{brockman2016gym, fisac2018general, henderson2018deep, mania2018simple, jang2019simulation, bernini2021few}). 

Reinforcement learning is based on the \textit{reward hypothesis} that all goals can be described by the maximization of expected return, i.e. the cumulative reward \cite{silver2015}. During training, the agent chooses an action, $\action_{NN}$, based on the input observation, $\obs$. The action is then executed in the environment, updating the internal state, $\state$, according to the plant dynamics. The updated state, $s$, is then assigned a scalar reward, $r$, and transformed into the next observation vector. In all the examples shown in this work, we assume full observability, so all state information exists in the observation or can be reconstructed accurately from a single observation, i.e. the transformation by the observer is reversible. The observation is useful because it allows us to normalize the state values and change the input dimensions in order to ignore irrelevant variables and/or increase the importance of other variables by repeating them. The process of executing an action and receiving a reward and next observation is referred to as a \emph{timestep}. Relevant values, like the input observation, action, and reward are collected as a data tuple, i.e. \emph{sample}, by the RL algorithm to update the current NNC policy, $\pi$, to an improved policy, $\pi^*$. How often these updates are done is dependent on the RL algorithm.


In this work, we focus on model-free DRL algorithms, meaning the agent has no dependency on the environment model during training. Within model-free DRL algorithms, there are two main categories of training, \emph{on-policy} and \emph{off-policy}. On-policy algorithms use the learned policy to select the actions taken during training, while off-policy algorithms use a separate policy. This distinction will cause the RTA to have a varied impact on the learning process. Thus, we repeat our experiments on two state-of-the-art DRL algorithms representing these two categories of training. \emph{Proximal Policy Optimization} (PPO) is our on-policy algorithm\footnote{Other on-policy RL algorithms include A2C \cite{mnih2016asynchronous}, TRPO \cite{schulman2015trust}, and ARS \cite{mania2018simple}.} and \emph{Soft Actor-Critic} (SAC) is our off-policy algorithm\footnote{Other off-policy RL algorithms include DQN \cite{mnih2015human}, DDPG \cite{lillicrap2015continuous}, and TD3 \cite{fujimoto2018addressing}.}. 

\section{Safe Reinforcement Learning} \label{sec:SRL}
When an RL agent explores states in a video game, the consequences of making a ``wrong" move are limited. However, using RL in the real world has shown catastrophic results \cite{fisac2018general, jang2019simulation}. The field of \emph{Safe Reinforcement Learning} (SRL) was developed in response to RL's use on cyber-physical systems domain that interact with the real world in complex scenarios. In Garc\'{i}a and Fern\'{a}ndez's comprehensive survey of SRL from 2015, they categorized the approaches into two main categories or styles: (1) modification of the optimality criterion and (2) modification of the exploration process \cite{garcia2015comprehensive}. In this work, we refer to these categories under the more general terms: (1) \emph{reward shaping} and (2) \emph{safe exploration}. Additionally, we introduce an emerging category of approaches, (3) \emph{adversarial training/retraining}. Each are described in more detail in this section. 



\subsubsection{Reward Shaping}

Reward shaping, the process of crafting a well-designed, optimal reward function,  
is essential for all forms of DRL since a poorly designed reward function can lead to unexpected and/or ineffective behavior \cite{hamilton2020sonic}. Within SRL, reward shaping is used to reformulate the problem as a \textit{Constrained Markov Decision Processes} (CMDP) \cite{altman1999constrained}. Instead of optimizing performance according to a singular reward function, performance is optimized according to a task-oriented reward and a safety-focused cost \cite{achiam2017constrained,wachi2020safe,ding2021provably,hasanzadezonuzy2020learning,jothimurugan2019composable,satija2020constrained}, 
so the agent learns a high-performing, safe policy. 

\subsubsection{Safe Exploration}

Safe exploration approaches, which are often geared towards hardware deployment, ensure the agent remains $100\%$ safe throughout the duration of training. Furthermore, this approach can be redesigned for deployment, ensuring the future safety of a static neural network that has completed training. 
Safe exploration techniques can be further broken down into the following three categories.

\begin{enumerate}
    \item \textbf{Preemptive Shielding} where the action set the agent is allowed to choose from is preemptively reduced to only allow safe actions \cite{alshiekh2018safe, wu2019shield}.
    \item \textbf{Safe-by-Construction} in which verification techniques are used, often on an abstraction of the learned policy, to verify safe behavior before being allowed to explore and develop further \cite{neary2021verifiable, zhang2021safe, wang2021verification}. Alternatively, correct-by-construction can also be applied to a shielded RL solution \cite{alshiekh2018safe}.
    \item \textbf{Run Time Assurance (RTA) methods} filter the agent's desired actions, $\action_{NN}$, to assure safety. In some cases, a monitor and/or decision module is used to determine whether the desired action provided by the learning agent is safe. In the event the agent's desired action is deemed unsafe, a different action that is determined to be safe is substituted and sent to the plant \cite{hunt2021hscc, wagener2021safe, xiong2021scalable, li2021safe, fisac2018general, fulton2018safe, desai2019soter, murugesan2019formal, cheng2019end, zhao2020learning}. 
\end{enumerate}

\begin{figure}[t]
    \centering
    \includegraphics[width=0.75\columnwidth]{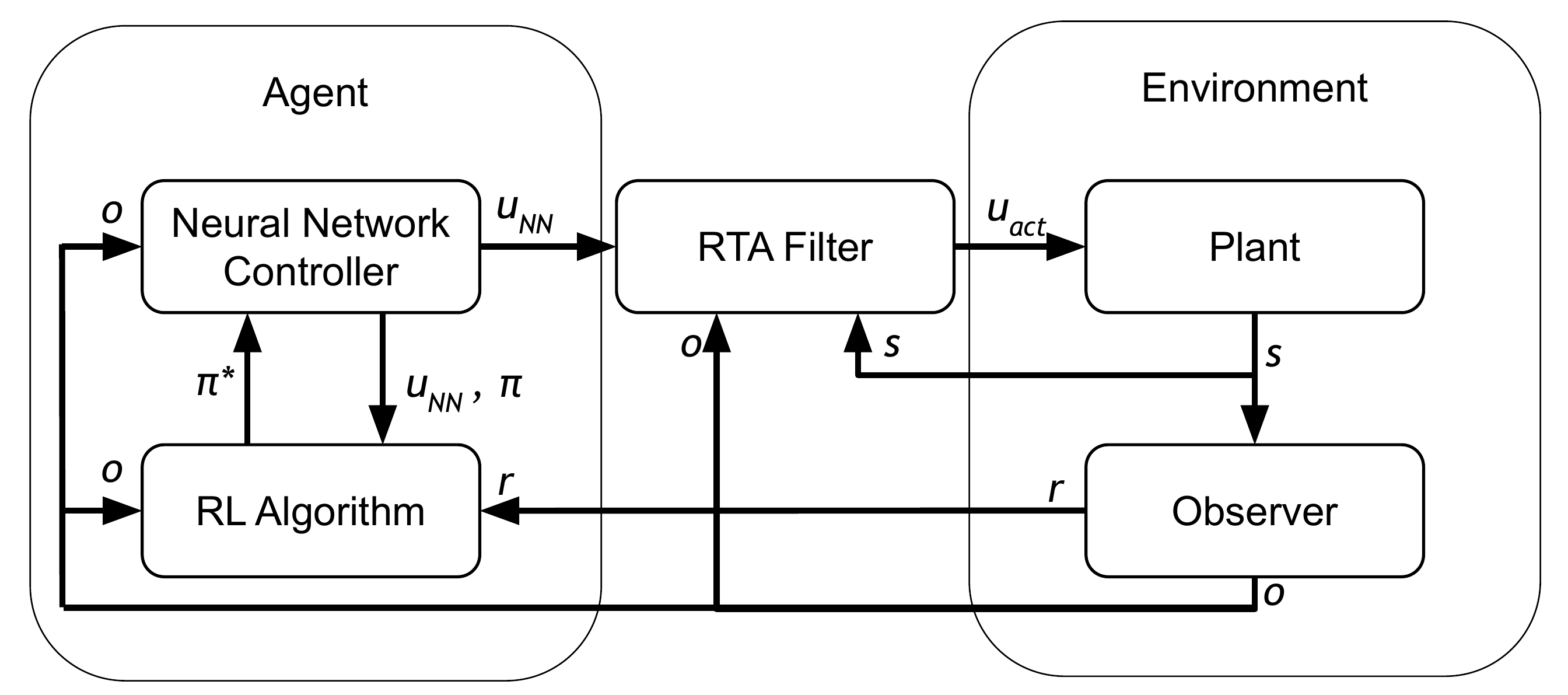}
    \caption{DRL training interactions between the agent and the environment with RTA.}
    \label{fig:rta_on}
\end{figure}

In this work, we focus solely on RTA methods for ensuring safe exploration, since they work across more examples with fewer scalability issues. An example of a general setup for safe exploration via RTA is shown in \figref{fig:rta_on}.

\subsubsection{Adversarial Training/Retraining}

The newest category of SRL approaches, \emph{Adversarial Training/Retraining}, focuses on identifying unsafe behavior in the agent and then generating data to learn from and correct that behavior \cite{phan2020neural, wang2020falsification, yang2021neural}. Most of the papers that use this approach focus on retraining an agent that already performs well in the environment. However, the approach can also be applied to an untrained network at the cost of requiring more training time.

\section{Run Time Assurance} \label{sec:RTA}
One of the main contributions of this work is investigating how the RL training process is impacted by RTA approaches that filter unsafe control inputs to preserve system safety. 
For this paper, we focus on dynamical system plant models sampled discretely given by $\state_{t+1} = f(\state_t, \action_t)$ where $\state_t \in \State$ is the state of the plant at timestep $t$, $\State \subseteq \mathbb{R}^n$ is the real-valued state space, $\action_t \in \Action$ is the control input to the plant at timestep $t$, with $\Action \subseteq \mathbb{R}^m$ the action space, and $f$ is a function describing the state evolution from current state and control action. 



For the dynamical system, inequality constraints $\varphi_i(\state): \mathbb{R}^n\to \mathbb{R}$, $\forall i \in \{1,...,M\}$ can be used to define a set of $M$ safety constraints, where the constraint is satisfied when $\varphi_i(\state) \geq 0$. The admissible set $\admissibleset \subseteq \State$, which is defined as the set of states where all constraints are satisfied, is then given by,
\begin{equation}
    \admissibleset := \{\state \in \State  \mid  \varphi_i(\state) \geq 0, \forall i \in \{1,...,M\} \}.
\end{equation}

\begin{definition}
Safety and/or safe operation is achieved by always remaining within the admissible set, i.e. not violating any specified constraints. In the examples provided in this work, safety is defined on a finite time horizon, such that the operation is considered safe if $\forall t \in [t_0, T], \state_t \in \admissibleset$. However, the ending time bound, $T$ can be set to infinity for other systems that operate in perpetuity.
\end{definition}

For RTA to ensure safe operation, we need to define a stricter subset of states to further constrain operations, known as the \emph{control invariant} safe set, $\safeset$. By operating in this stricter defined set, we avoid scenarios that can arise near the boundary of the admissible set, $\admissibleset$ where, no matter the action executed, the next state will be outside the admissible set. 

\begin{definition}
The control invariant safe set, $\safeset$, is a subset of the admissible set, $\admissibleset$, where $\forall \state \in \safeset, \exists \action \in \Action, f(\state, \action) \in \admissibleset$. 
\end{definition} 


In this work, we first focus on two classes of RTA monitoring approaches, \textit{explicit} and \textit{implicit}, which define $\safeset$ differently.
Explicit approaches use a pre-defined $\safeset$, to determine when RTA intervention is necessary. To define $\safeset$ explicitly, we first define a set of $M$ control invariant inequality constraints $h_i(\state):\mathbb{R}^n\to\mathbb{R}, \forall i \in \{1,...,M\}$, where the constraints are satisfied when $h_i(\state) \geq 0$.  $\safeset$ is then given by,
\begin{equation} \label{eq: explicitly_defined_safe_set}
    \safeset := \{\state \in \State \mid h_i(\state)\geq 0, \forall i \in \{1,...,M\} \}.
\end{equation} 

Implicit approaches use a defined backup control policy and the system dynamics to compute trajectories, which are used to determine when intervention is necessary. Implicitly, the $\safeset$ is defined as,
\begin{equation} \label{eq: implicitly_defined_safe_set}
    \safeset := \{\state \in \State \mid \forall k \in [t_0, T], \phi_k^{u_{\rm b}}(\state)\in\admissibleset \}, 
\end{equation}
where $\phi_k^{u_{\rm b}}$ represents a prediction of the state $\state$ for $k$ timesteps under the backup control policy $\action_{\rm b}$. Because computing trajectories can be computationally expensive, explicit approaches tend to be more efficient. However, implicit approaches can be easier to implement since they do not require a precise definition of the control invariant safe set, which is difficult to define without being overly conservative.

Additionally, we split the RTA monitoring approaches further with two classes of intervention, \emph{simplex} and \emph{Active Set-Invariance Filter} (ASIF). The simplex approach switches from the primary control to a pre-defined backup controller if the system is about to leave the control invariant safe set \cite{rivera1996architectural}. The backup controller is usually less efficient at the desired control task, but meets desired safety and/or human-machine teaming constraints. One possible implementation for a simplex RTA filter is constructed as follows,

\noindent \rule{1\columnwidth}{0.7pt}
\noindent \textbf{Simplex Filter}
\begin{equation}
\begin{array}{rl}
u_{\rm act}(\state)=
\begin{cases} 
\action_{\rm NN}(\obs(\state)) & {\rm if}\quad \phi_k^{\action_{\rm NN}}(\state) \in \safeset  \\ 
u_{\rm b}(\state)  & {\rm otherwise}
\end{cases}
\end{array}\label{eq:switching}
\end{equation}
\noindent \rule[7pt]{1\columnwidth}{0.7pt}
Here, $\phi_k^{\action_{\rm NN}}(\state)$ represents the predicted state if $\action_{\rm NN}$ is applied for $k$ discrete time intervals. 

ASIF approaches use barrier constraints to minimize deviations from the primary control signal while assuring safety \cite{gurriet2020applied}. One possible implementation for an ASIF RTA filter is constructed using a quadratic program as follows,

\noindent \rule{1\columnwidth}{0.7pt}
\noindent \textbf{Active Set-Invariance Filter}
\begin{equation}
\begin{split}
\action_{\rm act}(\state, \action_{\rm NN})=\text{argmin} & \left\Vert \action_{\rm NN}-\action_{\rm b}\right\Vert\\
\text{s.t.} \quad & BC_i(\state, \action_{\rm b})\geq 0, \quad \forall i \in \{1,...,M\}
\end{split}\label{eq:optimization}
\end{equation}
\noindent \rule[7pt]{1\columnwidth}{0.7pt}
Here, $BC_i(\state, \action_{\rm b})$ represents a set of $M$ barrier constraints \cite{ames2019control} used to assure safety of the system. The purpose of barrier constraints is to enforce Nagumo's condition \cite{nagumo1942lage} and ensure $\dot{h}_i(\state)$ is never decreasing $\forall i \in \{1,...,M\}$ along the boundary of $\safeset$. The function argmin finds the value of $\action_{\rm b}$ closest to $\action_{\rm NN}$ that still satisfies the barrier constraints. In this way, ASIF approaches apply the minimal change necessary to keep the system within $\admissibleset$ at each timestep.

Using these defined approaches, we categorize our experiments in this paper according to the four derived classes of RTA monitoring approaches: \emph{Explicit Simplex}, \emph{Implicit Simplex}, \emph{Explicit ASIF}, and \emph{Implicit ASIF}.
\section{Experiments}
In order to answer the questions posed in the introduction, we have designed 880 experiments across multiple environments, RTA configurations, RTA approaches, and random seeds\footnote{Code is currently undergoing the public release process and will be made available as soon as it is completed.}. 

For each environment and DRL algorithm, we use established hyperparameters to limit the impact of tuning. We use 10 random seeds to generate our traces \cite{henderson2018deep}. The evaluations run during training halt and freeze the NNC for the duration of the evaluation in order to better represent the performance of the agent at that point. After training is completed, the final learned policy is evaluated on the task 100 times to better approximate the expected performance if deployed. All evaluations are done in environments with and without the RTA active in order to identify any dependence on the RTA forming.

\input{configurations.tex}

\input{environments.tex}

\input{hparams.tex}

\section{Results and Discussion}
\label{sec:results}
In this section, we try to answer the questions posed in the introduction by analyzing the overarching trends found in our experiments. Including all the results collected would make this paper exceedingly long. Thus, we include select results that highlight the trends we found and provide all the results in Appendix~\ref{app:all_results}

\subsection{Do agents learn to become dependent on RTA?}
\label{sec:results_dependency}

\textbf{Answer:} Sometimes. Training RL agents with run time assurance \textit{always} runs the risk of forming dependence. Furthermore, this phenomenon is more prevalent in our on-policy results than our off-policy results.

An agent is dependent on the RTA if the RTA is necessary for safe and successful behavior during deployment. We can determine if an agent has learned to be dependent on the RTA by evaluating performance with and without the RTA. 
We can identify when an agent has learned to form a dependence if the return and success drop significantly when evaluated without the RTA. If the agent is independent of the RTA, the performance metrics should be consistent when evaluated with and without the RTA.

\begin{table}
    \centering
    \caption{This table shows final policy evaluation results across all test environments trained using the PPO algorithm with the Implicit Simplex RTA approach. 
    We show the recorded performance measured by the reward function (Return) and whether the agent was successful at completing the task (Success) 
    Rows highlighted in gray indicate a learned dependency.}
    \begin{tabular}{llccc}
    \toprule
      Environment & Configuration & RTA & Return & Success \\
    \midrule
Pendulum & RTA no punishment & on & 987.84 $\pm$ 10.86 & 1.00 $\pm$ 0.00 \\
& & off & 987.59 $\pm$ 10.38 & 1.00 $\pm$ 0.00 \\
Pendulum & RTA punishment & on & 987.57 $\pm$ 11.20 & 1.00 $\pm$ 0.00 \\
& & off & 987.85 $\pm$ 11.18 & 1.00 $\pm$ 0.00 \\

\rowcolor{Gray}
2D Spacecraft Docking & RTA no punishment & on &  2.02 $\pm$ 0.39 & 0.92 $\pm$ 0.27 \\
\rowcolor{Gray}
& & off & -22.39 $\pm$ 15.39 & 0.34 $\pm$ 0.47 \\
\rowcolor{Gray}
2D Spacecraft Docking & RTA punishment & on &  1.82 $\pm$ 0.60 & 0.73 $\pm$ 0.44 \\
\rowcolor{Gray}
& & off &  -17.99 $\pm$ 5.16 & 0.46 $\pm$ 0.50\\

3D Spacecraft Docking & RTA no punishment & on & -33.29 $\pm$ 22.18 & 0.50 $\pm$ 0.50 \\
& & off &  -23.51 $\pm$ 8.54 & 0.44 $\pm$ 0.50 \\
3D Spacecraft Docking & RTA punishment & on &    2.04 $\pm$ 0.39 & 0.89 $\pm$ 0.31 \\
& & off &    2.07 $\pm$ 0.34 & 0.92 $\pm$ 0.27 \\
    \end{tabular}
    \label{tab:rta_dependence_shortened}
\end{table}

We use gray to highlight examples where agents learned to form a dependency in \tabref{tab:rta_dependence_shortened}, which contains the final policy evaluations of our on-policy results across all the environments using the implicit simplex RTA approach. 
Note that not all agents in the highlighted rows (differentiated by the random seed used for training) learned a dependency, as evidenced by the increased standard deviation about the mean values. This further reinforces our answer that \emph{sometimes} agents learn to become dependent on the RTA they are trained with. Instead of ``always'' or ``never,'' whether the agent learns to become dependent on the RTA is a matter of chance, i.e. which random seed is used. Mania et al. show a great visualization in \cite{mania2018simple} of just how large an impact the random seed has on whether the agent learns a successful policy. The same is true here. If we had selected different random seeds, we would likely see different results for which agents learn to become dependent. However, it is the case that this can only happen if the agent is trained with RTA, and we have seen it is less likely to occur if the agent is punished for using the RTA as in the \textit{RTA punishment} configuration.  Note that the impact of the level/scale of punishment on whether dependence forms is left for future work.

\begin{figure}[htpb]
\centering
\subfigure[PPO evaluated with RTA Average Return]{\includegraphics[width=0.35\linewidth]{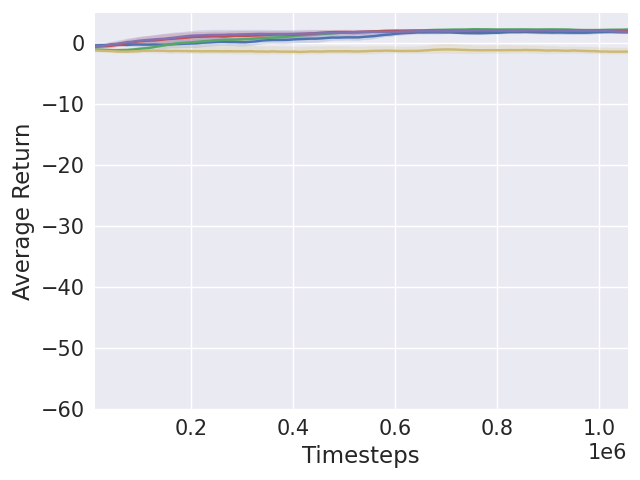}}\qquad
\subfigure[PPO evaluated with RTA Average Success]{\includegraphics[width=0.35\linewidth]{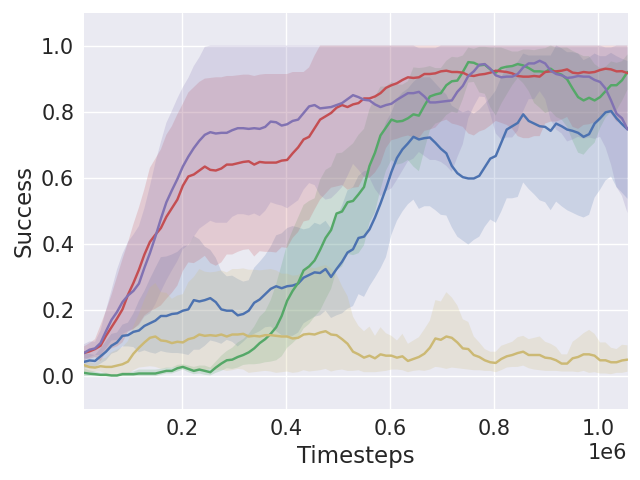}}\\
\subfigure[PPO evaluated without RTA Average Return]{\includegraphics[width=0.35\linewidth]{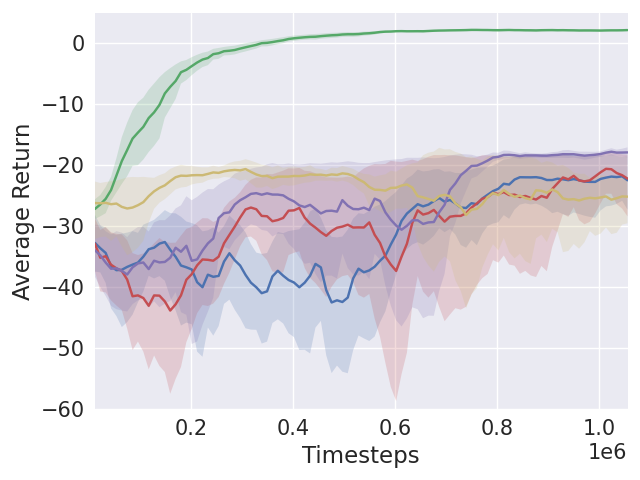}}\qquad
\subfigure[PPO evaluated without RTA Average Success]{\includegraphics[width=0.35\linewidth]{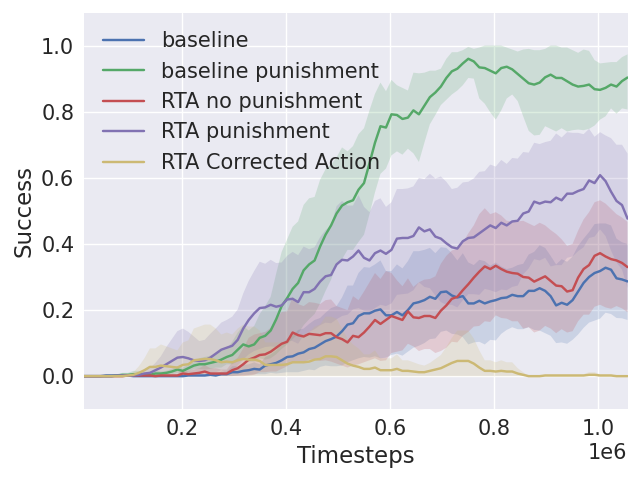}}
\caption{Results collected from experiments run in the 2D Spacecraft Docking environment with an implicit simplex RTA. Each curve represents the average of 10 trials, and the shaded region is the $95\%$ confidence interval about the mean. The large difference in return and success that is recorded with (a \& b) and without (c \& d) RTA shows that all agents trained with RTA learned to depend on it.}
\label{fig:docking2d_ppo_imp_sim_dependence}
\end{figure}

To further demonstrate issues with agents forming a dependence on RTA, observe the drop in performance and success between evaluating with RTA (a \& b) and without RTA (c \& d) in \figref{fig:docking2d_ppo_imp_sim_dependence}. While the RTA helps all the agents reach success throughout the training process, the agents trained with RTA (\textit{RTA no punishment}, \textit{RTA punishment}, and \textit{RTA Corrected Action}) do not maintain that same level of performance when evaluated without the RTA. In contrast, the \textit{baseline punishment} agents learn successful behavior that works with and without the RTA.

\subsection{Which RTA configuration is most effective?}
\label{sec:results_configuration}

\textbf{Answer:} \emph{Baseline punishment} is the most effective. However, if safe exploration is necessary, \emph{RTA punishment} is the most effective.

\begin{figure}[htb] 
\centering
\subfigure[2D Docking, PPO evaluated without RTA Average Return]{\includegraphics[width=0.35\linewidth]{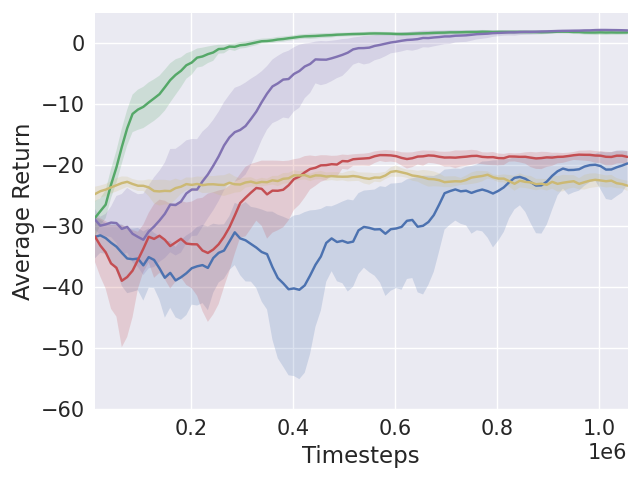}}\qquad
\subfigure[2D Docking, PPO evaluated without RTA Average Success]{\includegraphics[width=0.35\linewidth]{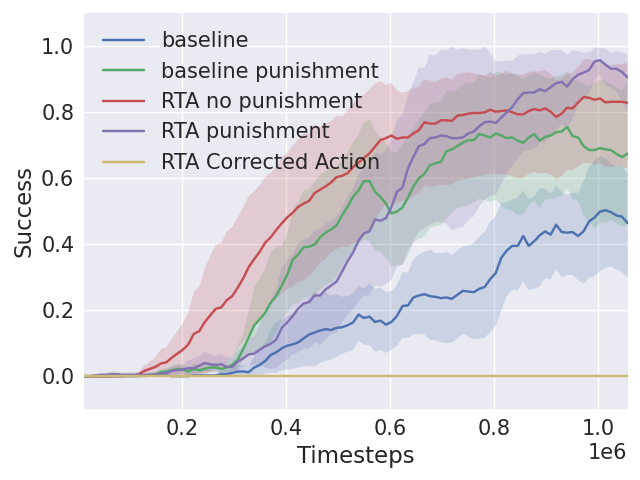}}\\
\subfigure[3D Docking, PPO evaluated without RTA Average Return]{\includegraphics[width=0.35\linewidth]{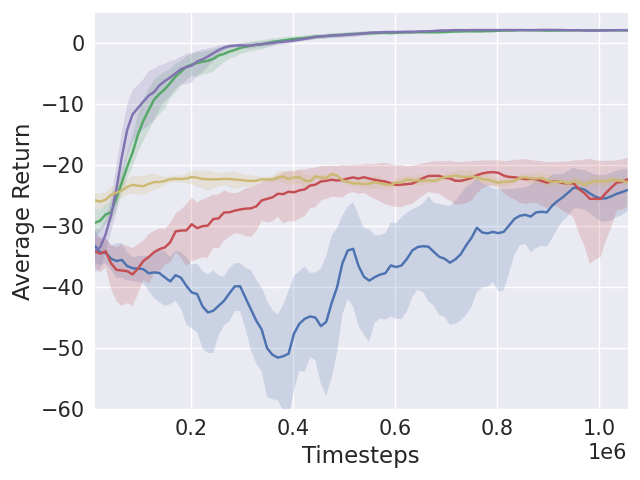}}\qquad
\subfigure[3D Docking, PPO evaluated without RTA Average Success]{\includegraphics[width=0.35\linewidth]{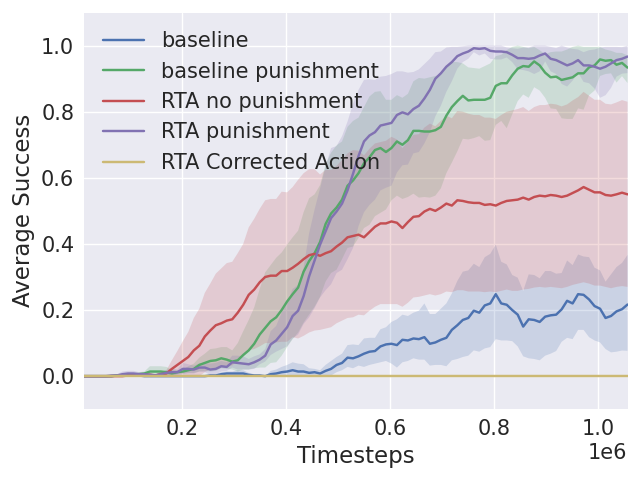}}
\caption{Results collected from experiments run in the 2D (a \& b) and 3D (c \& d) Spacecraft Docking environment with an explicit simplex RTA. Each curve represents the average of 10 trials, and the shaded region is the $95\%$ confidence interval about the mean. All plots show the \textit{baseline punishment} and \textit{RTA punishment} configurations learn at a similar rate and converge to similar levels of success and return.}
\label{fig:config_example}
\end{figure}

The most effective RTA configuration is the one that consistently trains the best performing agent evaluated without RTA. In the case of a tie and the final performance is comparable across multiple configurations, the best configuration is the one that learns the optimal performance quicker, i.e. requiring fewer samples. Across all our experiments, the \emph{baseline punishment} configuration was consistently among the best performing agents. The next best performer was the \emph{RTA punishment} configuration, which often outperformed the \emph{baseline punishment} configuration in our PPO experiments, but did not do as well in all of our SAC experiments. We discuss why this might be the case in Section~\ref{sec:results_policy}. To demonstrate this conclusion, we show the training curves in \figref{fig:config_example} from our experiments training agents across all configurations in both the 2D and 3D Spacecraft Docking environments using the PPO algorithm and the explicit simplex RTA approach.

In these particular examples, \figref{fig:config_example}, both \emph{baseline punishment} and \emph{RTA punishment} have similar training curves that converge about the same return and success. This is similar across most of our experiments, except in some experiments when \emph{RTA punishment} has a noticeably lower return because a dependence on the RTA formed.


\subsection{Which RTA approach is most effective?}
\label{sec:results_approach}

\textbf{Answer:} The explicit simplex is the most effective RTA approach for training agents that consistently perform well and do not learn to depend on the RTA to maintain safety.

\begin{figure}[htbp] 
\centering
\subfigure[PPO Average Return evaluated without RTA in the 2D Spacecraft Docking environment]{\includegraphics[width=0.35\linewidth]{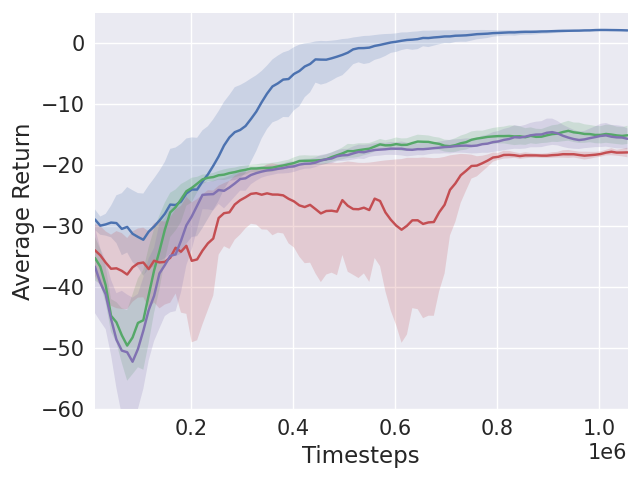}}\qquad
\subfigure[PPO Average Success evaluated without RTA in the 2D Spacecraft Docking environment]{\includegraphics[width=0.35\linewidth]{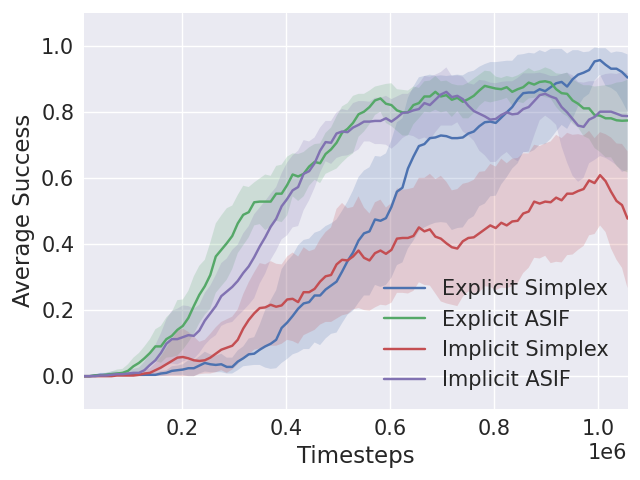}}\\
\subfigure[PPO Average Return evaluated without RTA in the 3D Spacecraft Docking environment]{\includegraphics[width=0.35\linewidth]{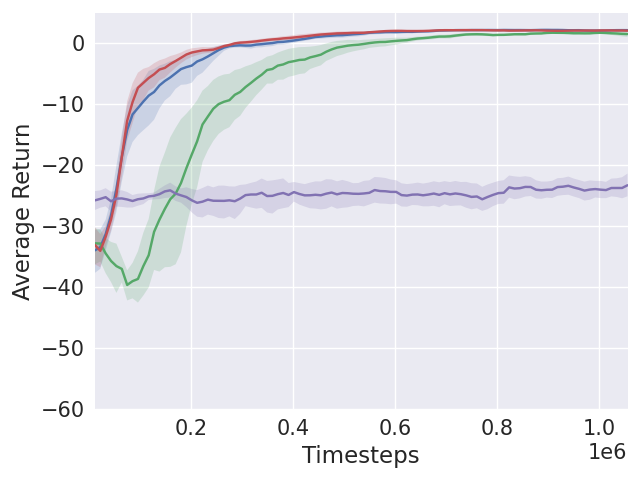}}\qquad
\subfigure[PPO Average Success evaluated without RTA in the 3D Spacecraft Docking environment]{\includegraphics[width=0.35\linewidth]{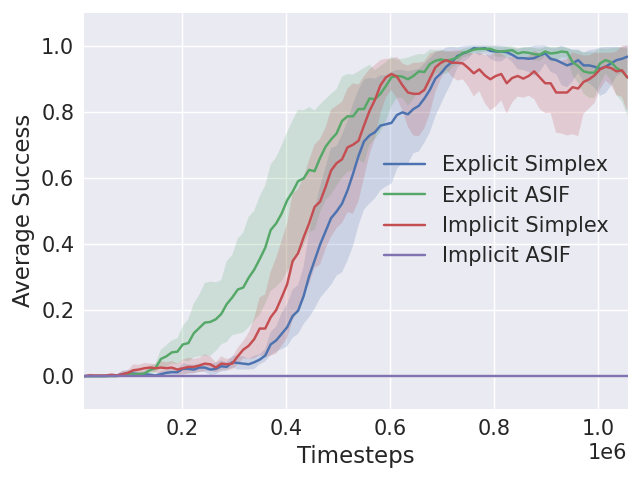}}
\caption{Training curves collected from experiments run in the 2D and 3D Spacecraft Docking environments, training with PPO in the \emph{RTA punishment} configuration across all four RTA approaches. Each curve represents the average of 10 trials, and the shaded region is the $95\%$ confidence interval about the mean.}
\label{fig:rta_approach_ex}
\end{figure}

\figref{fig:rta_approach_ex} shows the training curves for PPO agents trained in our 2D and 3D Spacecraft Docking environments with four different RTA approaches. All the training curves represent the PPO agents trained in the \emph{RTA punishment} configuration and evaluated without the RTA. The curves broadly show ASIF RTA's guide the agent to success earlier on, but at the cost of increased sample complexity. The simplex approaches instead have a reduced sample complexity achieving a higher return sooner, which then leads to a greater chance of success. 

We attribute these results to the differences between simplex and ASIF approaches. With simplex, the RTA does not intervene until the last moment, which allows for more agent-guided exploration, leading to more unique data samples. More unique data samples leads to a better approximation of the value- and/or Q-function, which reduces sample complexity. In contrast, ASIF approaches apply minimal corrections intended to guide the agent away from boundary conditions. This applies a greater restriction on agent-guided exploration, which can lead to more duplicated data samples.


The implicit RTA approaches were less effective than the explicit approaches and were less consistent. In the 2D Spacecraft Docking environment, both ASIF training curves had similar return and success. However, in the 3D Spacecraft Docking environment, the explicit ASIF curve maintained the trend of earlier success with reduced return while the implicit ASIF curve failed to improve throughout the entire training process. Similarly, in the 3D Spacecraft Docking environment, the simplex curves had similar return and success, but in the 2D Spacecraft Docking environment the implicit simplex curve showed a large drop in both return and success.

Therefore, we reason explicit RTA approaches are better for training. Additionally, simplex approaches lead to a better performing agent in the long run.

\subsection{Which works better with RTA, off-policy (SAC) or on-policy (PPO)?}
\label{sec:results_policy}

\textbf{Answer:} On-policy methods see a greater benefit from training with RTA.

Our results showed PPO sees a greater benefit from training with RTA than SAC. This is likely a result of how the methods approach the \emph{exploration versus exploitation} problem. Too much exploitation, using only known information (i.e. the current policy) too strictly, and the agent may never find the optimal policy. However, too much exploration and the agent may never learn what the goal is, particularly if the rewards are sparse. In general, on-policy methods leverage more exploitation than off-policy methods through their use of the learned policy. 

\begin{figure}[htpb]
\centering
\subfigure[SAC Average Return evaluated with RTA]{\includegraphics[width=0.35\linewidth]{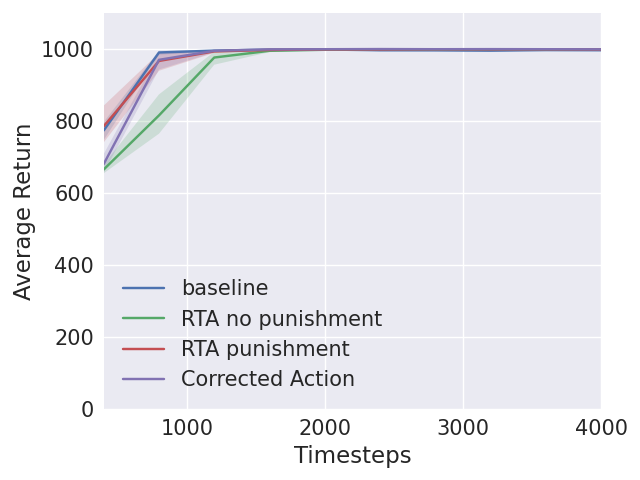}}\qquad
\subfigure[PPO Average Return evaluated with RTA]{\includegraphics[width=0.35\linewidth]{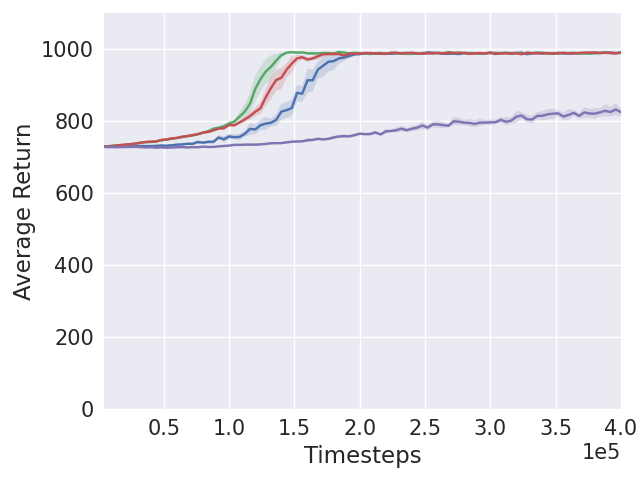}}\\
\subfigure[SAC Average Return evaluated no RTA]{\includegraphics[width=0.35\linewidth]{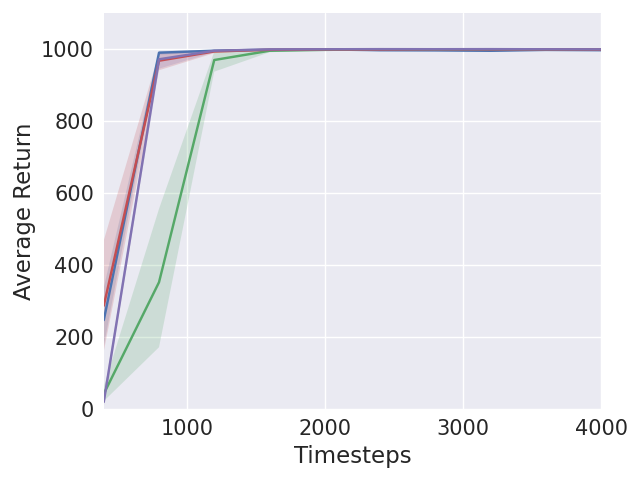}}\qquad
\subfigure[PPO Average Return evaluated no RTA]{\includegraphics[width=0.35\linewidth]{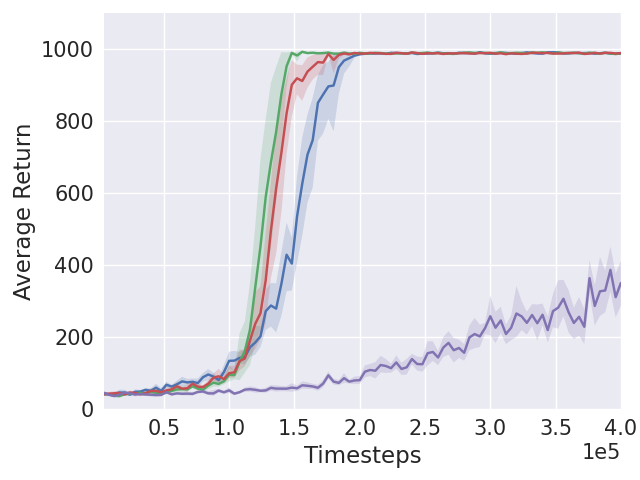}}
\caption{Results collected from experiments run in the Pendulum environment. Each curve represents the average of 10 trials, and the shaded region is the $95\%$ confidence interval about the mean. Note: maximum possible return in the environment is 1000.}
\label{fig:pendulum_results_policy}
\end{figure}

\textbf{On-policy} methods exploit the learned policy. Therefore, guiding the agent to success and away from unsafe behavior helps the agent learn that behavior. As a result, the sample complexity is reduced. \figref{fig:pendulum_results_policy} (b \& d) highlights this effect well. However, these benefits are hindered if the wrong configuration is chosen. Across almost all of our PPO experiments, the \emph{RTA Corrected Action} configuration prevented the agents from improving the learned policy as shown in \figref{fig:config_example}. This is likely a result of too much exploitation from the RTA intervening around boundary conditions, which prevented the agents from exploring other options to better define the optimal policy.

In contrast, \textbf{off-policy} methods have a larger focus on exploration. In particular, SAC maximizes entropy, assigning a higher value to unexplored state-action combinations. By restricting the actions taken near boundary conditions, the ``unsafe'' actions are never explored in actuality. Without making some distinction when the RTA intervenes makes the patterns harder to learn. This is shown in \figref{fig:bad_reward_example} where both \emph{RTA no punishment} and \emph{RTA Corrected Action} have a noticeably worse training curve than the \emph{baseline} configuration, failing to improve at all through training. We see a similar trend in \figref{fig:pendulum_results_policy} (a \& c) when SAC is used to learn the optimal policy for controlling our inverted pendulum. However, in this environment, the agent is still able to learn a successful policy.

\subsection{Which is more important, Reward Shaping or Safe Exploration?}
\label{sec:results_shaping}

\textbf{Answer:} Reward shaping is generally more important for training. While safe exploration can improve sample complexity in some cases, a well-defined reward function is imperative for training successful RL agents.

\begin{figure}[htbp] 
\centering
\subfigure[SAC evaluated with RTA Average Return]{\includegraphics[width=0.35\linewidth]{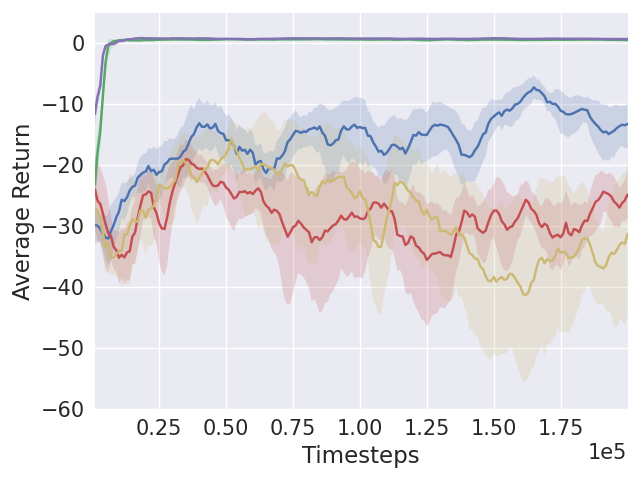}}\qquad
\subfigure[SAC evaluated with RTA Average Success]{\includegraphics[width=0.35\linewidth]{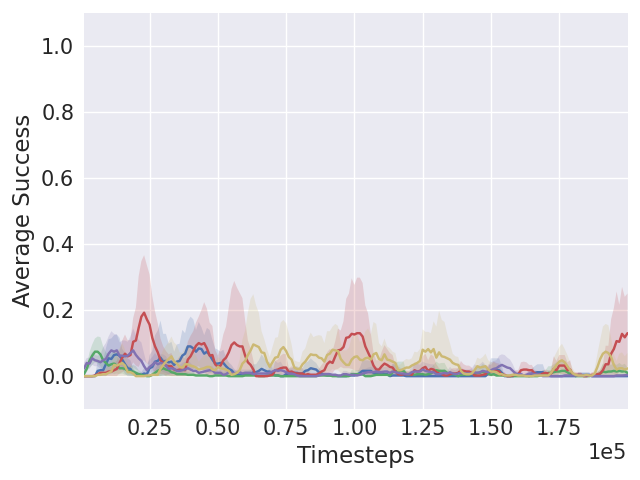}} \\
\subfigure[SAC evaluated without RTA Average Return]{\includegraphics[width=0.35\linewidth]{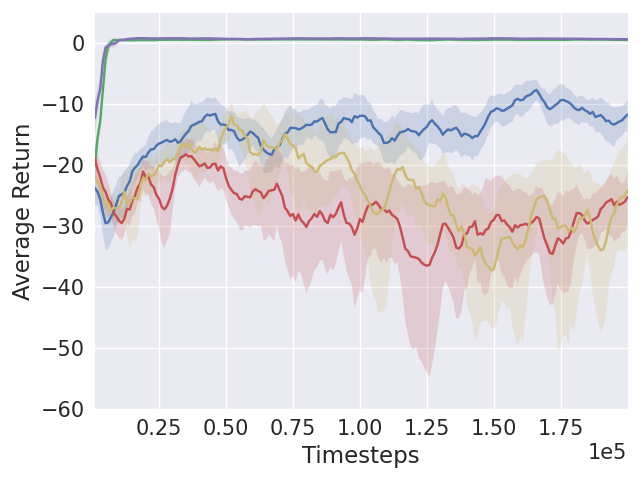}}\qquad
\subfigure[SAC evaluated without RTA Average Success]{\includegraphics[width=0.35\linewidth]{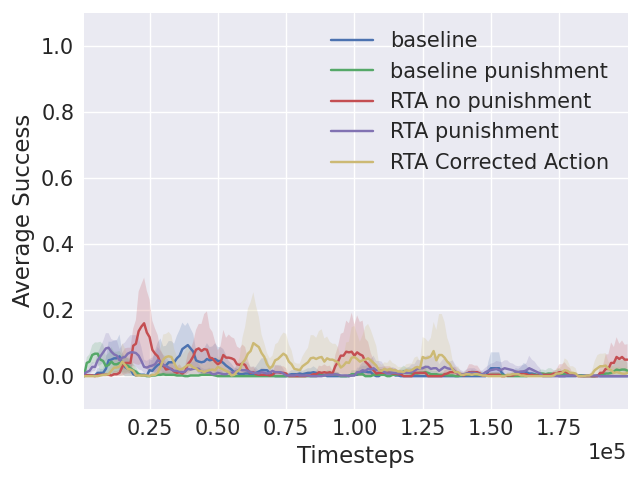}}
\caption{Results collected from experiments run in the 2D Spacecraft Docking environment with explicit simplex RTA. Each curve represents the average of 10 trials, and the shaded region is the $95\%$ confidence interval about the mean.}
\label{fig:bad_reward_example}
\end{figure}

In every experiment where reward shaping was applied, we saw a more consistent and improved training curve. For example, note in \figref{fig:config_example} that the 95\% confidence interval about \emph{baseline punishment} and \emph{RTA punishment} is smaller than \emph{baseline} and \emph{RTA no punishment} respectively. Additionally, the return and success tend to be much greater. The same trend shows in all of our experiments.

That said, safe exploration does improve sample complexity for on-policy RL, but the improvements are much greater when reward shaping is also applied in the \emph{RTA punishment} configuration. Additionally, the punishment helps prevent the agent from becoming dependent on the RTA.

However, safe exploration on its own is no substitute for a well-defined/tuned reward function, as evidenced in our SAC experiments in the docking environments shown in \figref{fig:bad_reward_example}. In these experiments, the agents with the \emph{baseline punishment} and \emph{RTA punishment} configurations quickly converged to an optimal performance, but the optimal performance did not result in success. 



\section{Conclusions and Future Work}

In conclusion, we trained 880 RL agents in 88 experimental configurations in order to answer some important questions regarding the use of RTA for training safe RL agents. Our results showed that \textbf{(1)} agents sometimes learn to become dependent on the RTA if trained with one, \textbf{(2)} \emph{baseline punishment} and \emph{RTA punishment} are the most effective configurations for training safe RL agents, \textbf{(3)} the explicit simplex RTA approach is most effective for consistent training results that do not depend on the RTA for safety, \textbf{(4)} PPO saw a greater benefit from training with RTA than SAC, suggesting that RTA may be more beneficial for on-policy than off-policy RL algorithms, and \textbf{(5)} effective reward shaping is generally more important than safe exploration for training safe RL agents.

In future work, and as more environments are released with RTA, we hope to expand this study to ensure our conclusions are representative of more complex training environments. Additionally, we would like to 
compare the effectiveness of correcting with RTA during training or retrain afterwards,
evaluate sim2real transfer of the trained RL agents in representative robotic environments, and
evaluate performance under environment and observation noise.

\section*{Acknowledgements}
This material is based upon work supported by the Department of Defense (DoD) through the National Defense Science \& Engineering Graduate (NDSEG) Fellowship Program, the Air Force Research Laboratory Innovation Pipeline Fund, and the Autonomy Technology Research (ATR) Center administered by Wright State University.
The views expressed are those of the authors and do not reflect the official guidance or position of the United States Government, the Department of Defense or of the United States Air Force. This work has been approved for public release: distribution unlimited. Case Number AFRL-2022-0550.

\bibliographystyle{cas-model2-names} 
\bibliography{references}

\newpage
\appendix
\appendixpage


\input{results.tex}

\end{document}

%% file: configurations.tex
\subsection{Run Time Assurance Configurations}
\label{app:configurations}
\begin{figure}
    \centering
    \includegraphics[width=0.4\columnwidth]{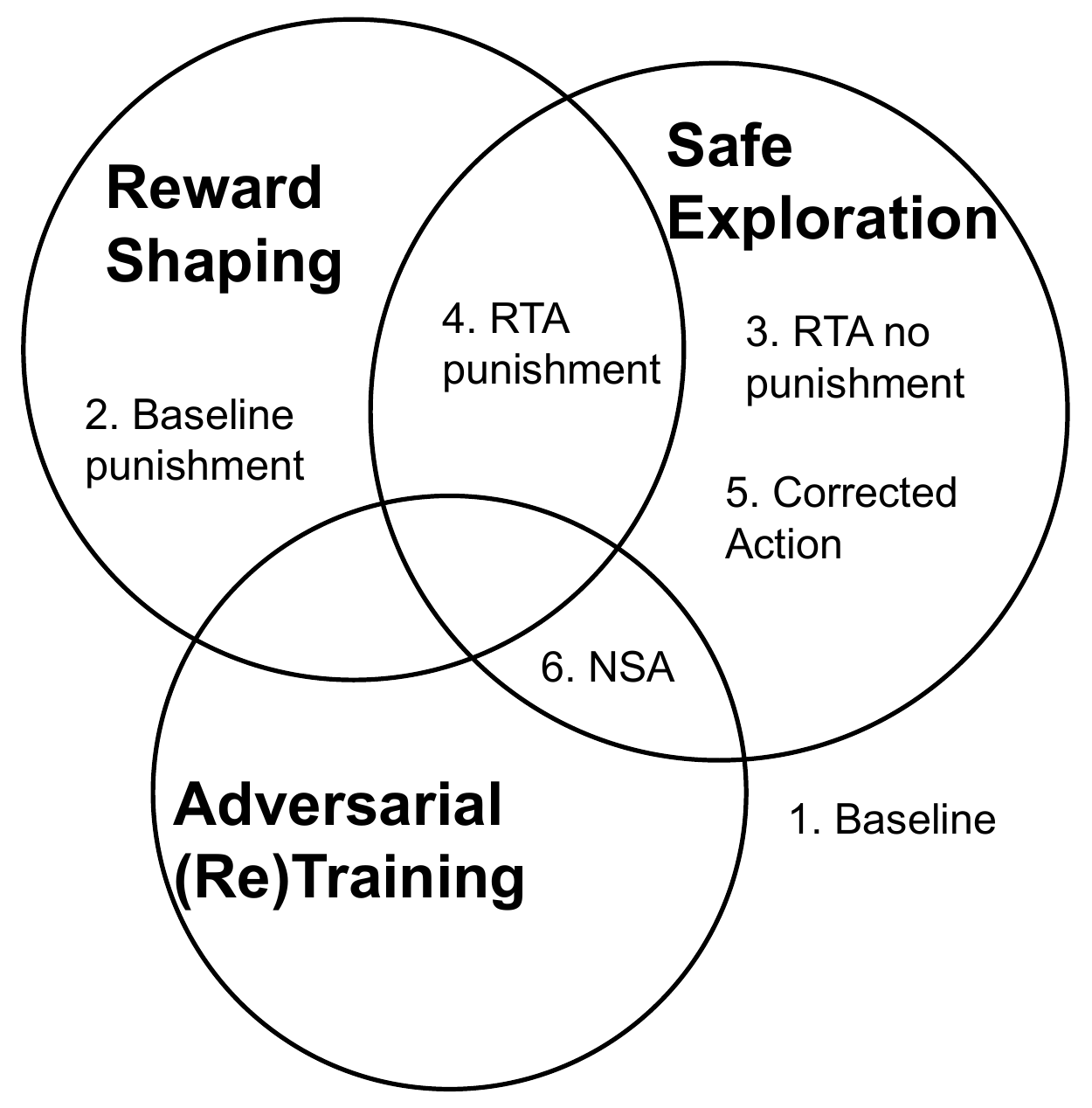}
    \caption{The RTA configurations used in our experiments represented within the three main categories of Safe Reinforcement Learning outlined in Section \ref{sec:SRL}.}
    \label{fig:app_rta_configs}
\end{figure}

In \figref{fig:rta_on}, we show a general method for including RTA in the training loop and purposefully left it vague. In the literature, there are many ways of connecting RTA. These range from treating it as an unknown feature of the environment, to using it for generating additional training data. In this work, we have separated these various ways of connecting the RTA into the 6 configurations shown in \figref{fig:app_rta_configs}. While we were only able to experiment with the first 5, the sixth is presented for future work. The configurations are listed in order of increasing complexity. Each configuration builds on the previous ones, helping us observe the impact of each addition. 

The configurations are explained in detail below. Because SAC and PPO collect different data tuples during training, we must define the configurations using different terms, $data_{SAC}$ and $data_{PPO}$, respectively.

\subsubsection{(1) Baseline (no RTA)}
This configuration, demonstrated in \figref{fig:rta_off}, is used as a baseline to compare all the RTA configurations against. In this configuration, the agent is learning according to the RL algorithm without any modifications. Note that for this comparison to be fair, the environment must be the same with no alterations to the initial set or terminal conditions.
\begin{align*}
    data_{SAC} &= \{\obs, \action_{\rm NN}, r, \obs'\} \\ 
    data_{PPO} &= \{\obs, \action_{\rm NN}, r, v, logp(\action_{\rm NN})\}
\end{align*}

$\obs$ is the input observation that led to the agent providing the output action, $\action_{\rm NN}$. $\obs'$ is the observation of the state reached after taking action $\action_{\rm NN}$ and $r$ is the reward value associated with it. $v$, the estimated value of the reached state, and $logp(\action_{\rm NN})$, the log-probability of selecting $\action_{\rm NN}$ given the current policy, are terms specific to PPO.

\subsubsection{(2) Baseline punishment}
In this configuration, we assign a negative reward, i.e. punishment $p$, if $unsafe?$ returns true, meaning at least one safety constraint was violated. This configuration adds SRL-style reward shaping to the problem. Instead of only maximizing the reward, the problem has two goals: (1) complete the task and (2) minimize the punishment, or cost, incurred from violating constraints. The remaining configurations cannot factor in this kind of punishment because they rely on safe exploration, which does not allow any violations of the safety constraints.
\begin{equation*}
    data_{SAC} =
    \begin{cases}
    \{\obs, \action_{\rm NN}, r + p, \obs'\}, & \text{if }unsafe? \\
    \{\obs, \action_{\rm NN}, r, \obs'\}, & \text{otherwise}
    \end{cases}
\end{equation*}
\begin{equation*}
    data_{PPO} = 
    \begin{cases}
    \{\obs, \action_{\rm NN}, r + p, v, logp(\action_{\rm NN})\}, & \text{if }unsafe? \\
    \{\obs, \action_{\rm NN}, r, v, logp(\action_{\rm NN})\}, & \text{otherwise}
    \end{cases}
\end{equation*}

\subsubsection{(3) RTA no punishment}
This configuration is the simplest form of safe exploration. Nothing changes from the baseline configuration, except the agent remains safe throughout the training process because of the RTA.
\begin{equation*}
    data_{SAC} =
    \begin{cases}
    \{\obs, \action_{\rm NN}, r, \obs'\}, & \text{if }intervening? \\
    \{\obs, \action_{\rm NN}, r, \obs'\}, & \text{otherwise}
    \end{cases}
\end{equation*}
\begin{equation*}
    data_{PPO} = 
    \begin{cases}
    \{\obs, \action_{\rm NN}, r, v, logp(\action_{\rm NN})\}, & \text{if }intervening? \\
    \{\obs, \action_{\rm NN}, r, v, logp(\action_{\rm NN})\}, & \text{otherwise}
    \end{cases}
\end{equation*}

\subsubsection{(4) RTA punishment}
This configuration adds an element of reward shaping to the previous configuration. Since we want the agent to learn the correct action to take in a scenario without the help of an RTA, we assign a punishment for having the RTA intervene. By adding this punishment, $p$, when the RTA intervenes, the agent should learn to make a distinction between safe and unsafe actions since safe actions will not incur a punishment.
\begin{equation*}
    data_{SAC} =
    \begin{cases}
    \{\obs, \action_{\rm NN}, r + p, \obs'\}, & \text{if }intervening? \\
    \{\obs, \action_{\rm NN}, r, \obs'\}, & \text{otherwise}
    \end{cases}
\end{equation*}
\begin{equation*}
    data_{PPO} = 
    \begin{cases}
    \{\obs, \action_{\rm NN}, r + p, v, logp(\action_{\rm NN})\}, & \text{if }intervening? \\
    \{\obs, \action_{\rm NN}, r, v, logp(\action_{\rm NN})\}, & \text{otherwise}
    \end{cases}
\end{equation*}

\subsubsection{(5) RTA Corrected Action}
In this configuration, we build on the idea of helping the agent identify the correct action to take in states near violating the safety constraints. Instead of punishing the agent for having the RTA intervene, we correct the agent's output to match that of the RTA's. In this manner, the agent only learns the actions actually taken in the environment.
\begin{align*}
    data_{SAC} &= \{\obs, \action_{\rm act}, r, \obs'\} \\
    data_{PPO} &= \{\obs, \action_{\rm act}, r, v, logp(\action_{\rm act})\}
\end{align*}

\subsubsection{(6) Neural Simplex Architecture (NSA)}
This configuration (not used in this work but planned for future work) is based on the SRL approach first published in \cite{phan2020neural}. In the authors' original implementation, NSA is used for retraining a learned policy. However, the retraining is done in an online approach, which can be easily adjusted to train a control policy from scratch.

This configuration estimates the result of taking the unsafe action and adds that estimate to the training data. This allows the agent to learn about unsafe actions without actually executing them. The additional data should help the agent develop a much better understanding of the environment and, thus, learn a more optimal policy after fewer timesteps. Currently, this configuration is limited to off-policy RL algorithms, but we are actively working on extending it for use in on-policy algorithms.
\begin{equation*}
    data_{SAC} =
    \begin{cases}
    \begin{cases}
    \{\obs, u_{b}, r, \obs'\} \text{ and } \\
    \{\obs, \action_{\rm NN}, est\_r, est\_\obs'\}
    \end{cases}, & \text{if }intervening? \\
    \{\obs, \action_{\rm NN}, r, \obs'\}, & \text{otherwise}
    \end{cases}
\end{equation*}

Here, $est\_r$ and $est\_\obs'$ are the estimated reward and estimated next observation, respectively. If an implicit RTA is used, the estimates can be computed using the internal simulation, $\phi_1^{\action_{NN}}$, which is used for determining if an action is safe.

%% file: environments.tex
\subsection{Environments}
\label{app:environments}

We ran our experiments in three environments with varying levels of complexity. By running our experiments in environments with different levels of complexity, we can observe whether the trends remain the same. If they do, then we can reasonably assume the trends will remain in even more complex environments. Currently, and to the best of our knowledge, these three environments are the only ones provided with accompanying RTA\footnote{As more RL environments are released with RTA, we hope to include them in our study to ensure our results continue to hold true.}. 

\subsubsection{Pendulum}
This environment was previously used in \cite{cheng2019end} as a good indicator of the effectiveness of SRL over standard Deep RL. We use the same initial conditions and constraints described in their work, explained below.

The goal of the agent in this environment is to use an actuator to keep the frictionless pendulum upright and within the bounds of $\pm 1rad \approx \pm 46^{\circ}$. Thus, the inequality constraint the RTA is designed to uphold can be written as,
\begin{equation}
\varphi_1(\state) := 1 - |\theta|,  
\end{equation}
where $\theta$ is the displacement angle of the pendulum measured from the upright position.

The interior plant model changes according to the discrete dynamics
\begin{align}
\label{eq:dynamics}
\begin{split}
    \omega_{t+1} &= \omega_{t} + (\frac{-3g}{2l} \sin(\theta_t + \pi) + \frac{3\action_t}{ml^2}) \Delta t \\
    \theta_{t+1} &= \theta_{t} + \omega_{t} \Delta t + (\frac{-3g}{2l} \sin(\theta_t + \pi) + \frac{3\action_t}{ml^2}) \Delta t^2,
\end{split}
\end{align}
where $g=10$, $l=1$, $m=1$, $\Delta t=0.05$, and $\action_t$ is the control from the neural network in the range $[-15, 15]$. Additionally, within the environment the pendulum's angular velocity, $\omega$, is clipped within the range $[-60, 60]$, and the angle from upright, $\theta$, is aliased within $[-\pi, \pi]$ radians. $\theta$ is measured from upright and increases as the pendulum moves clockwise. These values, $\theta$ and $\omega$, are then used to determine the input values for the neural network controller. The input observation is
\begin{equation}
    \obs = [\cos(\theta), \sin(\theta), \omega]^T.
\end{equation}



The pendulum is randomly initialized within a subset of the safe region in order to ensure it has enough time to intervene before violating the safety constraint. $\theta_0$ is randomly initialized between $\pm0.8rad$. $\omega_0$ is randomly initialized between $\pm1.0rad/s$.

In the event the safety constraint is violated, the episode is deemed a failure and immediately terminated. If the simulation were allowed to continue, the problem goal would change from simply keeping the pendulum upright, to also include learning how to swing back up. Additionally, this termination helps us identify safety violations later on in our analyses, since any safety violations would result in an episode length less than 200 timesteps.

The reward function was modified as well by adding a constant, $5$. $5$ was chosen in order to make a majority of the reward values positive. By keeping the reward positive, the agent is encouraged to not terminate the episode early. If the reward were mostly negative, the agent might learn the fastest way to terminate the episode in order to maximize the cumulative reward. The resulting reward function, Equation \ref{eq:app_reward_pendulum}, has a cumulative maximum of $1000$ instead of $0$.

\begin{equation}
\label{eq:app_reward_pendulum}
    r_t = 5 - (\theta^2_t + 0.1\omega_t^2 + 0.001\action_t^2)
\end{equation}

We use this reward function, Equation \ref{eq:app_reward_pendulum}, for all the evaluations we conducted. The punishment value used in various configurations when the RTA intervenes is $p = -1$. 


The RTA design implemented in this environment is a simple implicit simplex design. The backup controller, described by Equation \ref{eq:app_rta_pendulum}, intervenes if the desired control action is not recoverable using the backup controller. To determine whether the desired action, $\action_{NN}$ is recoverable, the desired action is simulated internally. If the simulated next state, $\bar{\state}_{t+1} = \phi_{1}^{\action_{NN}}(\state_t)$, violates the safety constraint, then the backup controller intervenes. In the event the simulated next state, $\bar{\state}_{t+1}$, is safe, a trajectory of up to 100 timesteps is simulated from $\bar{\state}_{t+1}$ using the backup controller. If any simulated state in the trajectory is unsafe, the desired action, $\action_{NN}$, is determined unsafe, and the backup controller intervenes. If the trajectory remains safe or a simulated next state is within the initial conditions, the desired action is determined to be safe, and the backup controller does not intervene. 
\begin{equation}
    \action_b(\state) = \min(\max(\frac{-32}{\pi} \theta, -15), 15)
    \label{eq:app_rta_pendulum}
\end{equation}

\subsubsection{Spacecraft Docking 2D \& 3D}
The cost of building and sending spacecraft into orbit is on the order of hundreds of millions of dollars. Therefore, it is in everyone's best interest to keep spacecraft in orbit operational and prevent collisions. Spacecraft docking is a common and challenging problem with a high risk for failure in the event an error occurs in the docking procedure and the two spacecraft collide. Here, we describe the problem with 3-dimensional dynamics, but repeated our experiments in a 2-dimensional environment where all $z$ values are held to a constant $0$.

The goal of the agent in these environments is to use mounted thrusters that move the \emph{deputy} spacecraft in the $x$, $y$, and $z$ directions to a docking region around the \emph{chief} spacecraft located at the origin. The state and observation are the same, 
\begin{equation*}
    \state = \obs = [x, y, z, \dot{x}, \dot{y}, \dot{z}]^T.
\end{equation*}
The action vector consists of the net force produced by the thrusters in each direction,
\begin{equation*}
    \action = [F_x,F_y,F_z]^T,
\end{equation*}
where each net force is a real value bounded in the range $\Action = [-1, 1]m/s^2$.

In this environment, the relative motion dynamics between the deputy and chief spacecraft are given by Clohessy-Wiltshire equations \cite{clohessy1960terminal}. These equations are a first-order approximation represented by
\begin{equation} \label{eq:app_system_dynamics}
    \state_{t+1} = A {\state_{t}} + B\action_t,
\end{equation}
where 
\begin{align}
\centering
    A = 
\begin{bmatrix} 
1    & 0 & 0    & 1   & 0  & 0 \\
0    & 1 & 0    & 0   & 1  & 0 \\
0    & 0 & 1    & 0   & 0  & 1 \\
3n^2 & 0 & 0    & 1   & 2n & 0 \\
0    & 0 & 0    & -2n & 1  & 0 \\
0    & 0 & -n^2 & 0   & 0  & 1 \\
\end{bmatrix},~
    B = 
\begin{bmatrix} 
 0 & 0 & 0 \\
 0 & 0 & 0 \\
 0 & 0 & 0 \\
\frac{1}{m} & 0 & 0 \\
0 & \frac{1}{m} & 0 \\
0 & 0 & \frac{1}{m} \\
\end{bmatrix}.
\end{align}
In these equations, $n = 0.001027rad/s$ is the spacecraft mean motion and $m = 12kg$ is the mass of the deputy spacecraft.

The agent, i.e. the deputy spacecraft, is randomly initialized in a stationary position ($\nu_{\rm H} = 0m/s$) around the chief so the distance from the chief, $d_{\rm H}$ is in the range $[100, 150]m$. From there, the deputy successfully docks if the distance between the deputy and chief, $d_{\rm H} = (x^2+y^2+z^2)^{1/2}$, is less than $20m$ and the deputy's relative speed, $\nu_{\rm H} =(\dot{x}^2+\dot{y}^2+\dot{z}^2)^{1/2}$, is less than $0.2m/s$. If the deputy is traveling faster than $0.2m/s$ within the docking region, then a crash occurs and the agent failed the task.

RTA is used in these environments to enforce a distance dependent speed limit and maximum velocity limits\footnote{More information on how and why these constraints were chosen can be found in \cite{dunlap2021Safe, dunlap2021comparing}.}. Together, these constraints keep the deputy spacecraft controllable and prevent collisions caused by the deputy approaching too fast. The distance dependent speed limit is defined as,
\begin{equation} \label{eq:ha1}
    \varphi_1(\state) := \nu_D - \nu_{\rm H} + c d_{\rm H},
\end{equation}
where $\nu_D = 0.2m/s$ defines the maximum allowable docking velocity and $c = 2n s^{-1}$ is a constant. The maximum velocity, $v_{\rm max} = 10m/s$, limits can be written as inequality constraints,
\begin{equation} \label{eq:app_ha2}
    \varphi_2(\state) := v_{\rm max}^2 - \dot{x}^2, \quad \varphi_3(\state) := v_{\rm max}^2 - \dot{y}^2, \quad
    \varphi_4(\state) := v_{\rm max}^2 - \dot{z}^2.
\end{equation}

\begin{table*}[htbp]
    \centering
    \caption{Spacecraft 2D Spacecraft Docking \& 3D reward function components}
    \begin{tabular}{lc}\hline
        \multicolumn{2}{c}{Terminal Rewards: All Configurations} \\ \hline
        Successfully Docked $(d_{\rm H} \leq 20m$ and $\nu_{\rm H}\leq 0.2 m/s)$ & +1 \\
        Crashed $(d_{\rm H} \leq 20 m$ with a velocity $\nu_{\rm H}> 0.2 m/s)$ & -1 \\
        Out of Bounds $(d_{\rm H} > 200m)$ & -1 \\
        Over Max Time/Control & -1 \\ \hline
        \multicolumn{2}{c}{Dense Reward: All Configurations} \\ \hline
        Proximity & $0.0125(\Delta d_{\rm H} )$ \\ \hline
        \multicolumn{2}{c}{Safety Rewards: Punishment Configurations} \\ \hline
        If RTA is Intervening & $-0.001$  \\
        Over Max Velocity & $-0.1 - 0.1(\nu_{\rm H} - v_{\rm max})$ \\ \hline
    \end{tabular}
    \label{tab:rewards}
\end{table*}

The reward functions for these environments are defined by sparse and dense components defined in \tabref{tab:rewards}\footnote{These values were provided by the authors of the environments during early development and do not match those published in \cite{ravaioli2022safe}. Additionally, these values were chosen with PPO as the target RL algorithm, which helps explain why SAC struggled with learning to complete the task.}. The sparsely defined terminal and safety reward components are only applied if the agent meets the specified requirements. In contrast, the dense reward component is computed after each timestep. In our experiments, the evaluation returns are computed using all the components defined in \tabref{tab:rewards}. However, during training, the safety components are ignored unless the punishment is required by the configuration.

%% file: hparams.tex
\subsection{Hyperparameters}
\label{app:hparams}
Providing the hyperparameters used in RL experiments is crucial for recreating the results. In all of our experiments, we train 10 agents using the following random seeds,
\begin{equation*}
    [1630, 2241, 2320, 2990, 3281, 4930, 5640, 8005, 9348, 9462].
\end{equation*}

Additionally, in this section, we provide the remaining hyperparameters used for training. No matter the configuration, the following hyperparameters were used. PPO hyperparameters are provided in \tabref{tab:ppo_hparams}. SAC hyperparameters are provided in \tabref{tab:sac_hparams}. 

\begin{table}[hp]
    \centering
    \caption{PPO Hyperparameters}
    \begin{tabular}{l|c|c|c}
         & Pendulum & Docking 2D & Docking 3D \\
    \midrule
        actor architecture & 64 tanh, 64 tanh, 1 linear & 64 tanh, 64 tanh, 2 linear & 64 tanh, 64 tanh, 3 linear \\
        critic architecture & 64 tanh, 64 tanh, 1 linear & 64 tanh, 64 tanh, 2 linear & 64 tanh, 64 tanh, 3 linear \\
        epoch length & 4000 & 10564 & 10564 \\
        epochs & 100 & 100 & 100 \\
        discount factor $\gamma$ & 0.0 & 0.988633 & 0.988633 \\
        clip ratio & 0.2 & 0.2 & 0.2 \\
        actor learning rate & 0.0003 & 0.001344 & 0.001344 \\
        critic learning rate & 0.001 & 0.001344 & 0.001344 \\
        updates per epoch & 80 & 34 & 34 \\
        target kl & 0.01 & 0.01 & 0.01 \\
        GAE-$\lambda$ & 0.0 & 0.904496 & 0.904496 \\
        max episode length & 200 & 1000 & 1000 \\
    \end{tabular}
    \label{tab:ppo_hparams}
\end{table}

\begin{table}[hp]
    \centering
    \caption{SAC Hyperparameters}
    \begin{tabular}{l|c|c|c}
         & Pendulum & Docking 2D & Docking 3D \\
    \midrule
        actor architecture & 64 ReLU, 64 ReLU, 1 tanh & 64 ReLU, 64 ReLU, 2 tanh & 64 ReLU, 64 ReLU, 3 tanh \\
        critic architecture & 64 ReLU, 64 ReLU, 1 ReLU & 64 ReLU, 64 ReLU, 1 ReLU & 64 ReLU, 64 ReLU, 1 ReLU \\
        epoch length & 400 & 1000 & 1000 \\
        epochs & 40 & 1000 & 1000 \\
        replay buffer size & 10000 & 10000 & 10000 \\
        discount factor $\gamma$ & 0.99 & 0.99 & 0.99 \\
        polyak & 0.995 & 0.995 & 0.995 \\
        entropy coefficient $\alpha$ & 0.2 & 0.2 & 0.2 \\
        actor learning rate & 0.001 & 0.001 & 0.001 \\
        critic learning rate & 0.001 & 0.001 & 0.001 \\
        minibatch size & 256 & 256 & 256 \\
        update after \_ step(s) & 1 & 1 & 1 \\
        max episode length & 200 & 1000 & 1000 \\
    \end{tabular}
    \label{tab:sac_hparams}
\end{table}

%% file: results.tex
\section{All Experimental Results}
\label{app:all_results}
In this section, we show the results collected from all of our experiments. The subsections are broken up according to environment and RTA approach. Experiments in the pendulum environment are in Section \ref{app:pendulum}. Experiments in the 2D Spacecraft Docking environment are in Sections \ref{app:docking2dexplicitsimplex}, \ref{app:docking2dexplicitasif}, \ref{app:docking2dimplicitsimplex}, and \ref{app:docking2dimplicitasif}. Experiments in the 3D Spacecraft Docking environment are in Sections \ref{app:docking3dexplicitsimplex}, \ref{app:docking3dexplicitasif}, \ref{app:docking3dimplicitsimplex}, and \ref{app:docking3dimplicitasif}.   

\FloatBarrier \subsection{Pendulum Implicit Simplex}
\label{app:pendulum}

\begin{figure}[ht]
\centering
\subfigure[SAC with RTA]{\includegraphics[width=0.45\linewidth]{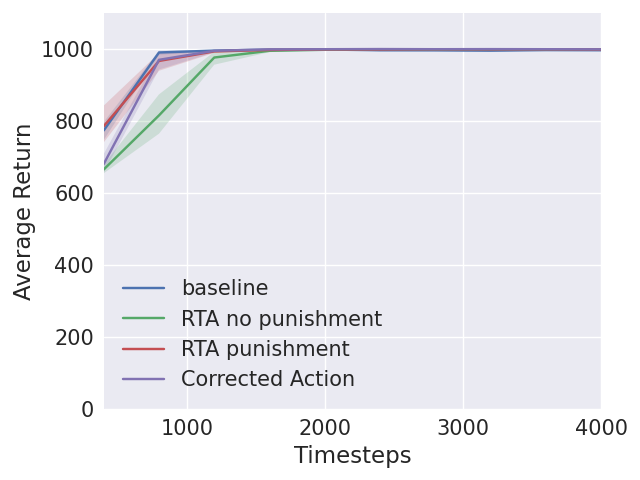}}\qquad
\subfigure[PPO with RTA]{\includegraphics[width=0.45\linewidth]{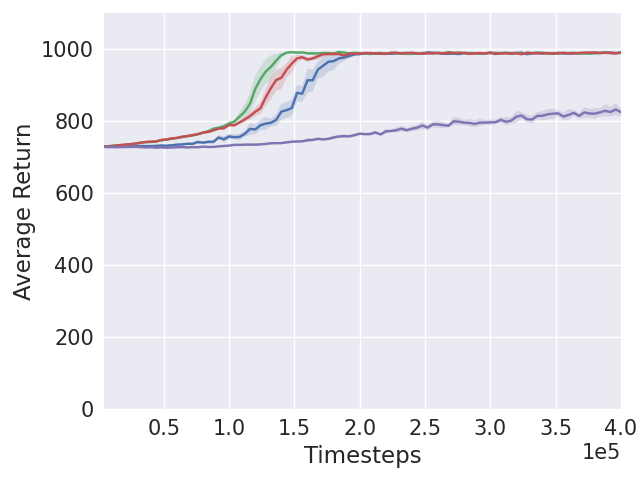}}\\
\subfigure[SAC no RTA]{\includegraphics[width=0.45\linewidth]{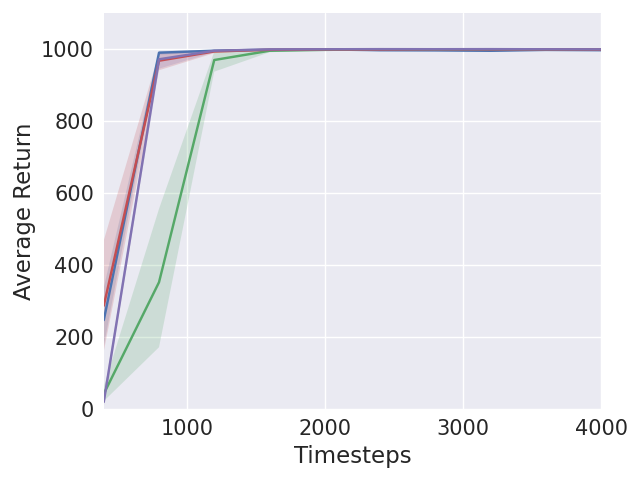}}\qquad
\subfigure[PPO no RTA]{\includegraphics[width=0.45\linewidth]{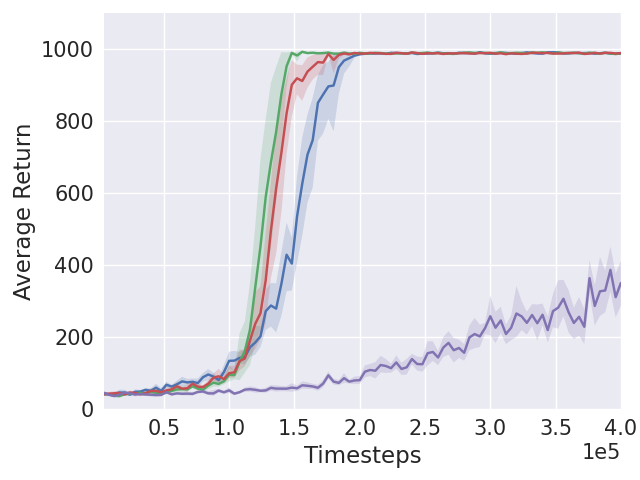}}
\caption{Results collected from experiments run in the Pendulum environment. Each curve represents the average 10 trials and the shaded region is the $95\%$ confidence interval about the mean. Note: maximum possible return in the environment is 1000.}
\label{fig:pendulum_results}
\end{figure}


\begin{table}[hb]
\caption{SAC Pendulum}
\label{tab:SAC Pendulum}
\centering
\begin{tabular}{lcccc}
    \toprule
    Configuration &   RTA &      Return &         Length &  Interventions\\
    \midrule
         baseline & on & 983.1007 $\pm$ 3.1389 & 200.0 $\pm$ 0.0 & 0.2940 $\pm$ 0.7250 \\
         & off &  982.8630 $\pm$ 3.3692 & 200.0 $\pm$ 0.0 & - \\
RTA no punishment & on & 983.9275 $\pm$ 2.0795 & 200.0 $\pm$ 0.0 & 0.0370 $\pm$ 0.2401 \\
& off & 982.8445 $\pm$ 30.7715 & 199.8070 $\pm$ 6.1001 & - \\
   RTA punishment & on & 984.1265 $\pm$ 2.0494 & 200.0 $\pm$ 0.0 & 0.0 $\pm$ 0.0 \\
   & off &  984.0623 $\pm$ 1.9408 & 200.0 $\pm$ 0.0 & - \\
 Corrected Action & on & 984.5683 $\pm$ 2.1403 & 200.0 $\pm$ 0.0 & 0.0030 $\pm$ 0.0547 \\
 & off &  984.5059 $\pm$ 2.0753 & 200.0 $\pm$ 0.0 & - \\
    \bottomrule
\end{tabular}
\end{table}

\begin{table}[hb]
\caption{PPO Pendulum}
\label{tab:PPO Pendulum}
\centering
\begin{tabular}{lcccc}
\toprule
       Configuration &   RTA &      Return &         Length &  Interventions\\
\midrule
 baseline & on & 987.9677 $\pm$ 10.9786 & 200.0 $\pm$ 0.0 &   0.0 $\pm$ 0.0 \\
 & off &  987.3375 $\pm$ 11.2540 & 200.0 $\pm$ 0.0 & - \\
RTA no punishment & on & 987.8363 $\pm$ 10.8634 & 200.0 $\pm$ 0.0 &   0.0 $\pm$ 0.0 \\
& off &  987.5939 $\pm$ 10.3785 & 200.0 $\pm$ 0.0 & - \\
   RTA punishment & on & 987.5672 $\pm$ 11.2048 & 200.0 $\pm$ 0.0 &   0.0 $\pm$ 0.0 \\
   & off &  987.8488 $\pm$ 11.1780 & 200.0 $\pm$ 0.0 & - \\
 Corrected Action & on & 850.7139 $\pm$ 51.6744 & 200.0 $\pm$ 0.0 & 27.7470 $\pm$ 11.6062 \\
 & off & 306.7720 $\pm$ 281.0874 & 64.7700 $\pm$ 56.5273 & - \\
\bottomrule
\end{tabular}
\end{table}




\FloatBarrier \subsection{2D Spacecraft Docking Explicit Simplex}
\label{app:docking2dexplicitsimplex}

\begin{figure}[ht]
\centering
\subfigure[PPO evaluated with RTA Average Return]{\includegraphics[width=0.45\linewidth]{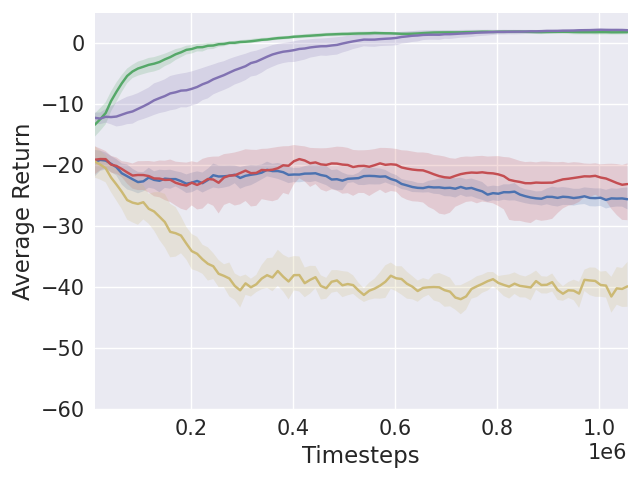}}\qquad
\subfigure[PPO evaluated with RTA Average Success]{\includegraphics[width=0.45\linewidth]{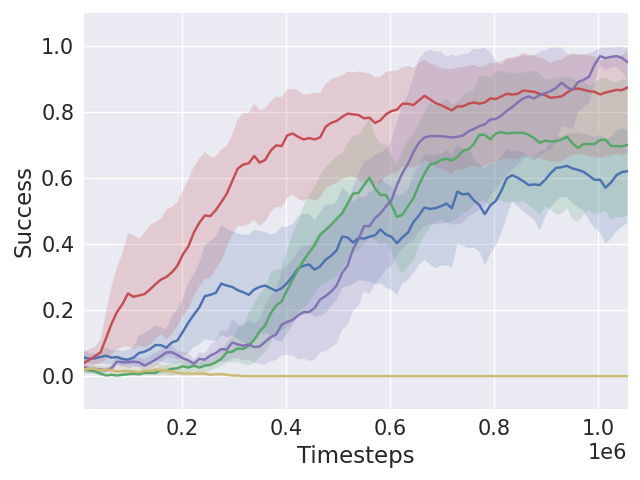}}\\
\subfigure[PPO evaluated without RTA Average Return]{\includegraphics[width=0.45\linewidth]{figures/docking2d_explicit_simplex/ppo_no_rta_eval.png}}\qquad
\subfigure[PPO evaluated without RTA Average Success]{\includegraphics[width=0.45\linewidth]{figures/docking2d_explicit_simplex/ppo_no_rta_success.png}}
\caption{Results collected from experiments run in the 2D Spacecraft Docking environment with an explicit simplex RTA. Each curve represents the average 10 trials and the shaded region is the $95\%$ confidence interval about the mean.}
\label{fig:docking2d_ppo_exp_sim_results}
\end{figure}

\begin{table}[hb]
\caption{PPO 2D Spacecraft Docking Explicit Simplex}
\label{tab:ppo2dexplicitsimplex}
\centering
\resizebox{1.0\linewidth}{!}{
\begin{tabular}{lcccccc}
\toprule
       Configuration & RTA & Return & Length & Success &      Interventions/Violations &   Correction \\
\midrule
        baseline & on &  -25.5058 $\pm$ 7.0824 & 451.3160 $\pm$ 228.9647 & 0.5960 $\pm$ 0.4907 &  269.1270 $\pm$ 68.7516 &  0.8521 $\pm$ 0.2730 \\
        & off & -19.5788 $\pm$ 7.5443 & 261.0470 $\pm$ 255.9112 & 0.4740 $\pm$ 0.4993 & 132.0560 $\pm$ 59.6531 & - \\
 baseline punishment & on &    1.7415 $\pm$ 0.7302 & 763.2820 $\pm$ 197.3288 & 0.6720 $\pm$ 0.4695 &     1.0440 $\pm$ 2.4430 &  0.0501 $\pm$ 0.0980 \\
 & off &   1.7350 $\pm$ 0.7505 & 758.7400 $\pm$ 198.3093 & 0.6830 $\pm$ 0.4653 &    1.1420 $\pm$ 2.7484 & - \\
   RTA no punishment & on & -22.5140 $\pm$ 10.7681 &  348.6550 $\pm$ 85.4651 & 0.8530 $\pm$ 0.3541 & 242.0030 $\pm$ 100.1411 &  0.9448 $\pm$ 0.3900 \\
   & off & -18.3351 $\pm$ 4.1843 &  230.2570 $\pm$ 68.7432 & 0.8320 $\pm$ 0.3739 & 154.8820 $\pm$ 42.6876 & - \\
      RTA punishment & on &    2.0770 $\pm$ 0.3609 & 710.4490 $\pm$ 157.6271 & 0.8820 $\pm$ 0.3226 &     1.0830 $\pm$ 1.9236 &  0.0595 $\pm$ 0.0981 \\
      & off &   2.0637 $\pm$ 0.3659 & 703.4900 $\pm$ 155.3529 & 0.8920 $\pm$ 0.3104 &    1.2450 $\pm$ 2.2545 & - \\
RTA Corrected Action & on & -38.8204 $\pm$ 23.3258 & 370.8200 $\pm$ 230.2876 & 0.0 $\pm$ 0.0 & 370.8060 $\pm$ 230.2876 & 12.8179 $\pm$ 3.2593 \\
& off & -22.7426 $\pm$ 9.9165 &    18.0930 $\pm$ 4.9153 & 0.0 $\pm$ 0.0 &   18.0720 $\pm$ 4.8995 & - \\
\bottomrule
\end{tabular}
}
\end{table}

\begin{figure}[ht]
\centering
\subfigure[SAC evaluated with RTA Average Return]{\includegraphics[width=0.45\linewidth]{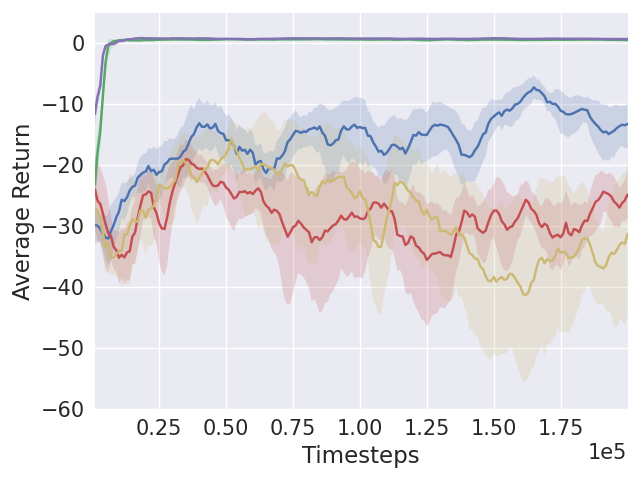}}\qquad
\subfigure[SAC evaluated with RTA Average Success]{\includegraphics[width=0.45\linewidth]{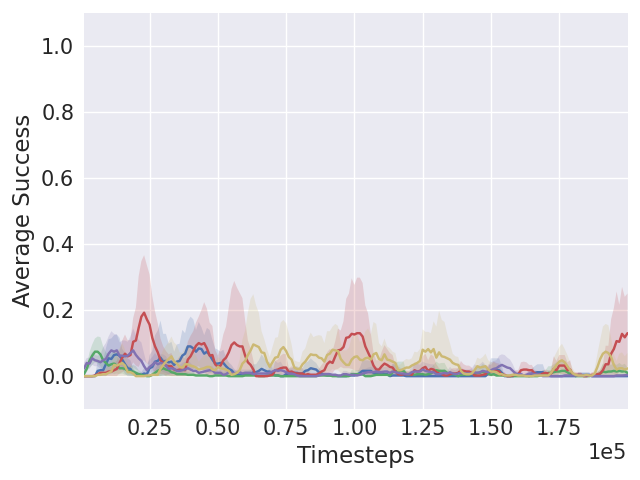}}\\
\subfigure[SAC evaluated without RTA Average Return]{\includegraphics[width=0.45\linewidth]{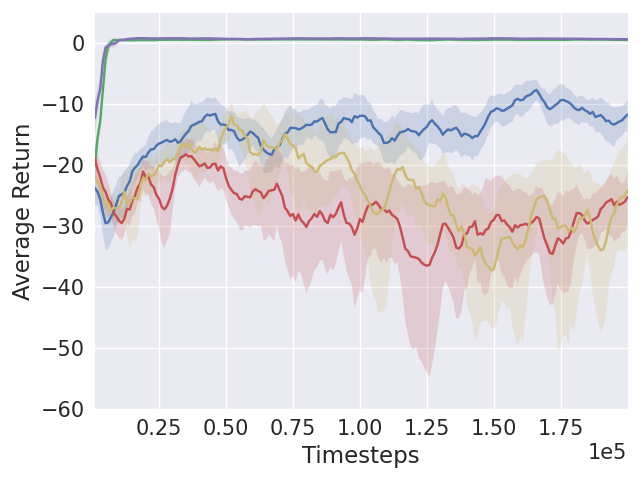}}\qquad
\subfigure[SAC evaluated without RTA Average Success]{\includegraphics[width=0.45\linewidth]{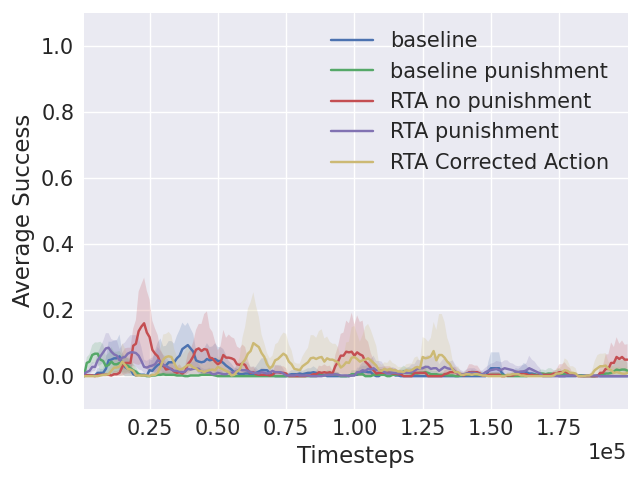}}
\caption{Results collected from experiments run in the 2D Spacecraft Docking environment with an explicit simplex RTA. Each curve represents the average 10 trials and the shaded region is the $95\%$ confidence interval about the mean.}
\label{fig:docking2d_sac_exp_sim_results}
\end{figure}

\begin{table}[hb]
\caption{SAC 2D Spacecraft Docking Explicit Simplex}
\label{tab:sac2dexplicitsimplex}
\centering
\resizebox{1.0\linewidth}{!}{
\begin{tabular}{lcccccc}
\toprule
       Configuration & RTA & Return & Length & Success &      Interventions/Violations &   Correction \\
\midrule
            baseline & on &  -12.5070 $\pm$ 5.3815 &  980.3370 $\pm$ 95.3160 & 0.0170 $\pm$ 0.1293 &  129.1350 $\pm$ 54.4923 & 0.4160 $\pm$ 0.0389 \\
            & off & -44.4387 $\pm$ 18.4755 & 936.0760 $\pm$ 201.4800 & 0.0 $\pm$ 0.0 & 375.4500 $\pm$ 150.4911 & - \\
 baseline punishment & on &   -0.7163 $\pm$ 1.1067 &  998.6170 $\pm$ 18.4751 & 0.0030 $\pm$ 0.0547 &   10.8460 $\pm$ 12.2911 & 0.2622 $\pm$ 0.1204 \\
 & off &   -1.4651 $\pm$ 2.3989 &  995.6790 $\pm$ 41.1139 & 0.0050 $\pm$ 0.0705 &   17.8680 $\pm$ 24.0389 & - \\
   RTA no punishment & on &  -14.5861 $\pm$ 4.6865 & 943.9070 $\pm$ 160.2614 & 0.0380 $\pm$ 0.1912 &  148.0320 $\pm$ 50.7698 & 0.4289 $\pm$ 0.0300 \\
   & off & -55.0964 $\pm$ 28.7844 & 639.9500 $\pm$ 320.4161 & 0.0080 $\pm$ 0.0891 & 397.9490 $\pm$ 210.2359 & - \\
      RTA punishment & on &   -2.2407 $\pm$ 1.5318 &  994.1500 $\pm$ 53.0907 & 0.0010 $\pm$ 0.0316 &   25.4220 $\pm$ 16.9404 & 0.3336 $\pm$ 0.0755 \\
      & off &   -5.9313 $\pm$ 5.0539 &  996.6070 $\pm$ 35.4610 & 0.0040 $\pm$ 0.0631 &   59.2540 $\pm$ 48.6995 & - \\
RTA Corrected Action & on & -19.8202 $\pm$ 14.5191 & 927.6350 $\pm$ 179.7466 & 0.0580 $\pm$ 0.2337 & 199.4710 $\pm$ 144.4070 & 0.4516 $\pm$ 0.0666 \\
& off & -43.6724 $\pm$ 20.6484 & 588.6050 $\pm$ 361.0214 & 0.0050 $\pm$ 0.0705 & 295.2280 $\pm$ 154.5340 & - \\
\bottomrule
\end{tabular}
}
\end{table}





\FloatBarrier \subsection{2D Spacecraft Docking Explicit ASIF}
\label{app:docking2dexplicitasif}

\begin{figure}[ht]
\centering
\subfigure[PPO evaluated with RTA Average Return]{\includegraphics[width=0.45\linewidth]{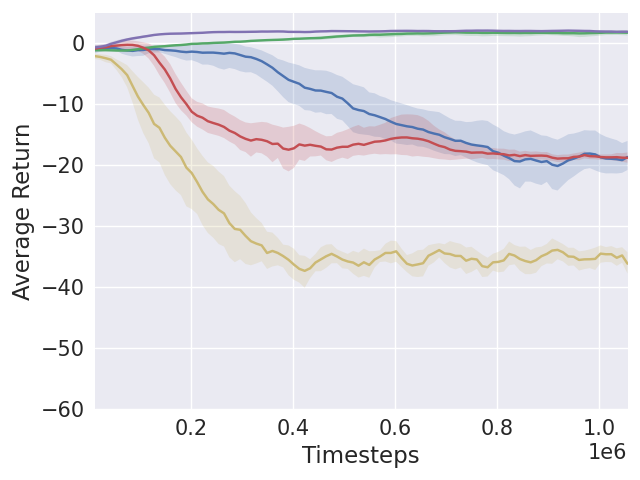}}\qquad
\subfigure[PPO evaluated with RTA Average Success]{\includegraphics[width=0.45\linewidth]{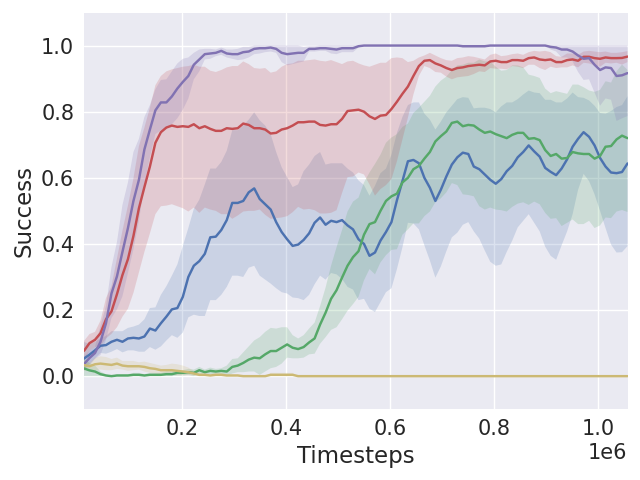}}\\
\subfigure[PPO evaluated without RTA Average Return]{\includegraphics[width=0.45\linewidth]{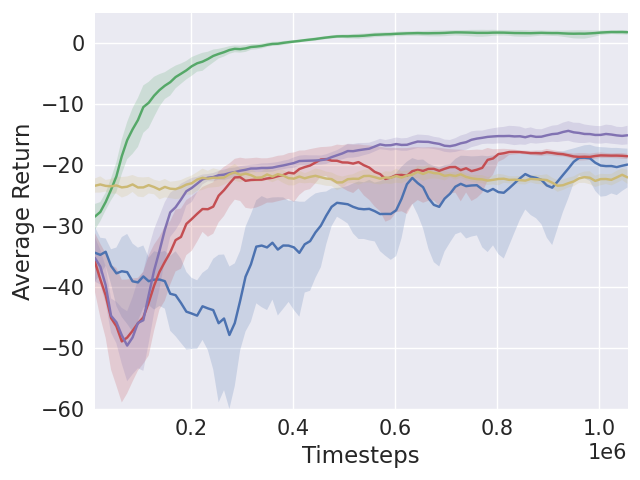}}\qquad
\subfigure[PPO evaluated without RTA Average Success]{\includegraphics[width=0.45\linewidth]{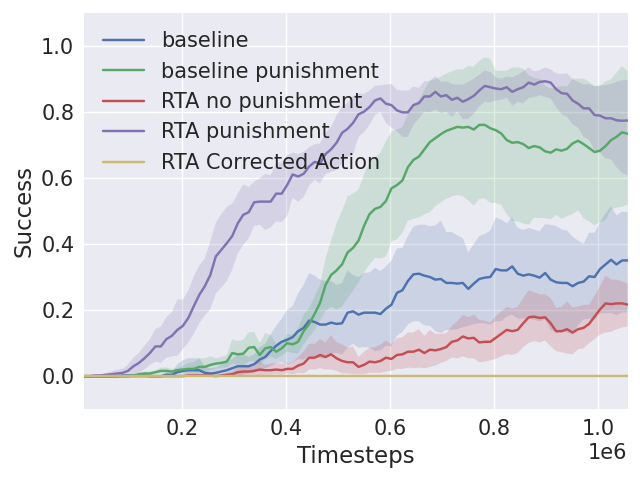}}
\caption{Results collected from experiments run in the 2D Spacecraft Docking environment with an explicit ASIF RTA. Each curve represents the average 10 trials and the shaded region is the $95\%$ confidence interval about the mean. }
\label{fig:docking2d_ppo_exp_asif_results}
\end{figure}

\begin{table}[hb]
\caption{PPO 2D Spacecraft Docking Explicit ASIF}
\label{tab:ppo2dexplicitasif}
\centering
\resizebox{1.0\linewidth}{!}{
\begin{tabular}{lcccccc}
\toprule
       Configuration & RTA & Return & Length & Success &      Interventions/Violations &   Correction \\
\midrule
            baseline & on &  -19.3937 $\pm$ 8.3859 & 496.2690 $\pm$ 276.8729 & 0.6100 $\pm$ 0.4877 &  320.1170 $\pm$ 76.3203 & 0.9750 $\pm$ 0.3698 \\
            & off & -19.6417 $\pm$ 8.6105 & 288.4830 $\pm$ 318.4605 & 0.3710 $\pm$ 0.4831 & 126.4670 $\pm$ 64.2468 & - \\
 baseline punishment & on &    1.6885 $\pm$ 0.7824 & 794.0 $\pm$ 174.0669 & 0.6830 $\pm$ 0.4653 &  135.3380 $\pm$ 47.6325 & 0.2399 $\pm$ 0.0411 \\
 & off &   1.7063 $\pm$ 0.8210 & 770.9680 $\pm$ 187.9315 & 0.6920 $\pm$ 0.4617 &    1.1970 $\pm$ 2.4409 & - \\
   RTA no punishment & on &  -18.8412 $\pm$ 3.9873 &  327.1660 $\pm$ 35.9863 & 0.9570 $\pm$ 0.2029 &  287.7570 $\pm$ 32.6175 & 0.9100 $\pm$ 0.1651 \\
   & off & -18.4948 $\pm$ 2.2503 &  137.2860 $\pm$ 31.5151 & 0.2320 $\pm$ 0.4221 & 126.7610 $\pm$ 19.5246 & - \\
      RTA punishment & on &    1.8265 $\pm$ 0.7208 & 512.1800 $\pm$ 150.3579 & 0.9330 $\pm$ 0.2500 &  198.6380 $\pm$ 45.1417 & 0.2718 $\pm$ 0.0470 \\
      & off & -15.3082 $\pm$ 5.7940 & 391.2710 $\pm$ 193.5872 & 0.7970 $\pm$ 0.4022 & 152.7720 $\pm$ 47.8269 & - \\
RTA Corrected Action & on & -35.2004 $\pm$ 20.0159 & 321.1060 $\pm$ 181.7095 & 0.0 $\pm$ 0.0 & 321.1060 $\pm$ 181.7095 & 9.7873 $\pm$ 3.5195 \\
& off & -22.4530 $\pm$ 9.9569 &    20.7170 $\pm$ 6.3494 & 0.0 $\pm$ 0.0 &   20.5450 $\pm$ 6.2174 & - \\
\bottomrule
\end{tabular}
}
\end{table}

\begin{figure}[ht]
\centering
\subfigure[SAC evaluated with RTA Average Return]{\includegraphics[width=0.45\linewidth]{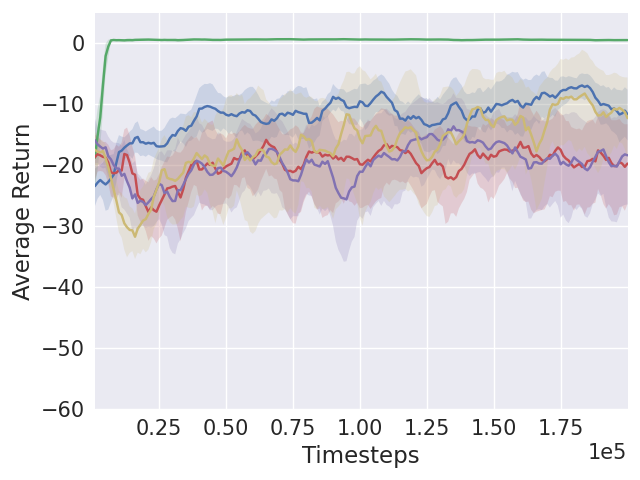}}\qquad
\subfigure[SAC evaluated with RTA Average Success]{\includegraphics[width=0.45\linewidth]{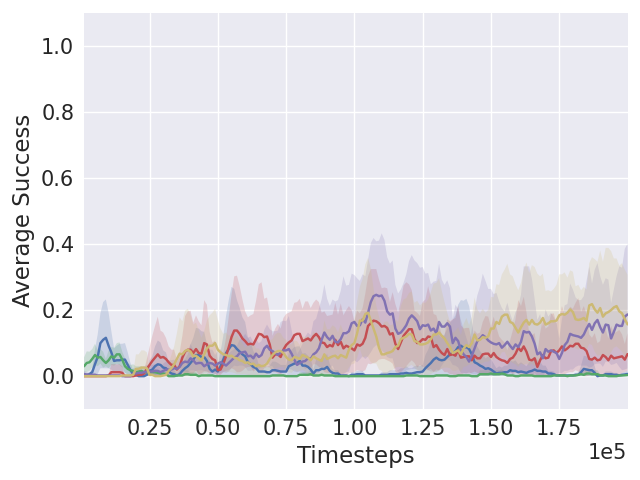}}\\
\subfigure[SAC evaluated without RTA Average Return]{\includegraphics[width=0.45\linewidth]{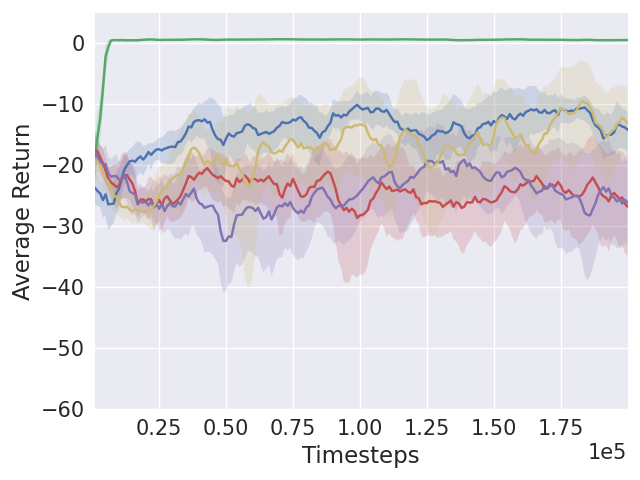}}\qquad
\subfigure[SAC evaluated without RTA Average Success]{\includegraphics[width=0.45\linewidth]{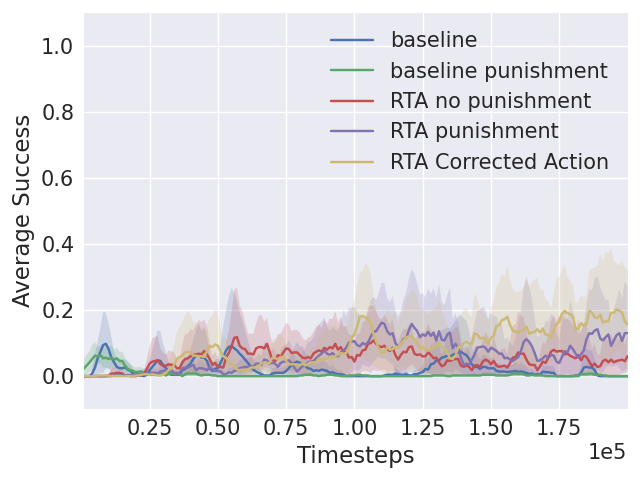}}
\caption{Results collected from experiments run in the 2D Spacecraft Docking environment with an explicit ASIF RTA. Each curve represents the average 10 trials and the shaded region is the $95\%$ confidence interval about the mean.}
\label{fig:docking2d_sac_exp_asif_results}
\end{figure}

\begin{table}[hb]
\caption{SAC 2D Spacecraft Docking Explicit ASIF}
\label{tab:sac2dexplicitasif}
\centering
\resizebox{1.0\linewidth}{!}{
\begin{tabular}{lcccccc}
\toprule
       Configuration & RTA & Return & Length & Success &      Interventions/Violations &   Correction \\
\midrule
            baseline & on &  0.0216 $\pm$ 0.9744 & 968.7900 $\pm$ 127.7141 & 0.0320 $\pm$ 0.1760 & 280.0150 $\pm$ 76.8260 & 0.3916 $\pm$ 0.0454 \\
            & off & -45.9935 $\pm$ 21.5150 & 890.9630 $\pm$ 269.6588 & 0.0010 $\pm$ 0.0316 & 376.5710 $\pm$ 170.1958 & - \\
 baseline punishment & on &  0.1026 $\pm$ 0.2979 &  999.5220 $\pm$ 10.2286 & 0.0 $\pm$ 0.0 & 151.7450 $\pm$ 25.1203 & 0.3134 $\pm$ 0.0211 \\
 & off &   -0.9064 $\pm$ 1.0850 &  995.0900 $\pm$ 51.4919 & 0.0 $\pm$ 0.0 &   12.0070 $\pm$ 11.5531 & - \\
   RTA no punishment & on & -0.1290 $\pm$ 0.6667 & 974.0790 $\pm$ 107.3582 & 0.0010 $\pm$ 0.0316 & 257.1470 $\pm$ 44.1387 & 0.3755 $\pm$ 0.0204 \\
   & off & -37.6844 $\pm$ 22.5865 & 374.3540 $\pm$ 225.3516 & 0.0 $\pm$ 0.0 & 250.7620 $\pm$ 155.8610 & - \\
      RTA punishment & on & -0.0374 $\pm$ 0.5376 &  987.6740 $\pm$ 73.9589 & 0.0020 $\pm$ 0.0447 & 264.8470 $\pm$ 47.4221 & 0.3788 $\pm$ 0.0252 \\
      & off & -41.9909 $\pm$ 23.4668 & 428.7950 $\pm$ 266.7232 & 0.0020 $\pm$ 0.0447 & 282.9180 $\pm$ 171.1889 & - \\
RTA Corrected Action & on &  0.2143 $\pm$ 0.5982 &  990.1150 $\pm$ 56.8130 & 0.0280 $\pm$ 0.1650 & 252.5090 $\pm$ 53.4003 & 0.3716 $\pm$ 0.0291 \\
& off & -36.1541 $\pm$ 18.2568 & 598.6270 $\pm$ 325.5361 & 0.0120 $\pm$ 0.1089 & 269.8170 $\pm$ 136.6618 & - \\
\bottomrule
\end{tabular}
}
\end{table}

\FloatBarrier \subsection{2D Spacecraft Docking Implicit Simplex}
\label{app:docking2dimplicitsimplex}

\begin{figure}[ht]
\centering
\subfigure[PPO evaluated with RTA Average Return]{\includegraphics[width=0.45\linewidth]{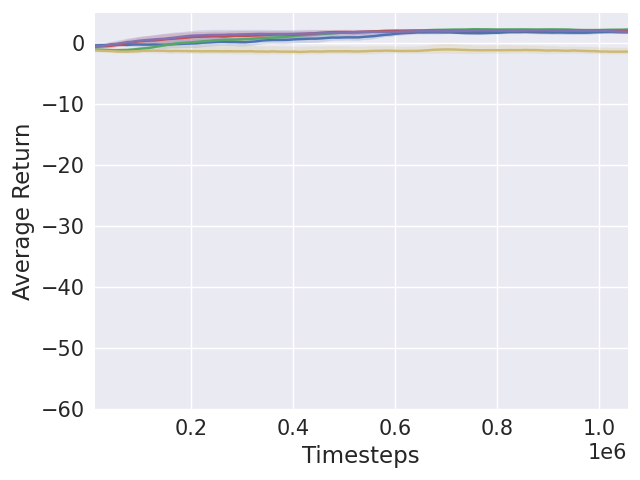}}\qquad
\subfigure[PPO evaluated with RTA Average Success]{\includegraphics[width=0.45\linewidth]{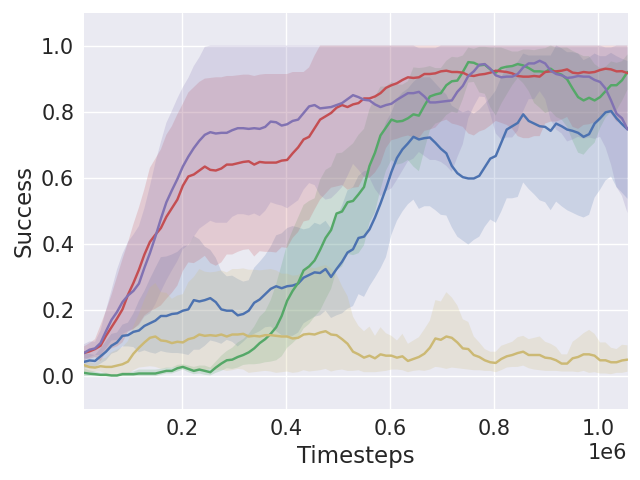}}\\
\subfigure[PPO evaluated without RTA Average Return]{\includegraphics[width=0.45\linewidth]{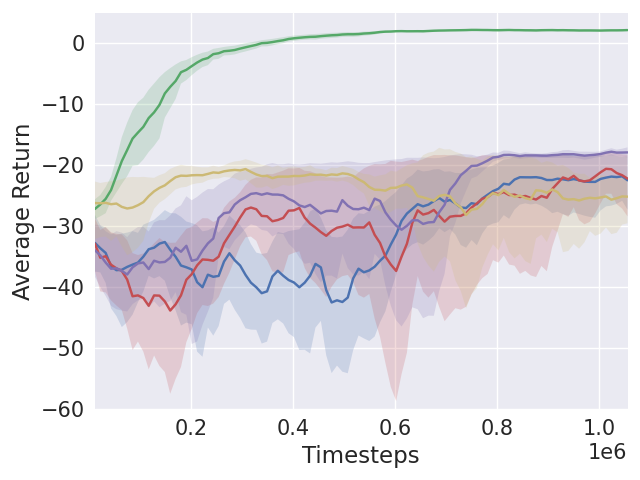}}\qquad
\subfigure[PPO evaluated without RTA Average Success]{\includegraphics[width=0.45\linewidth]{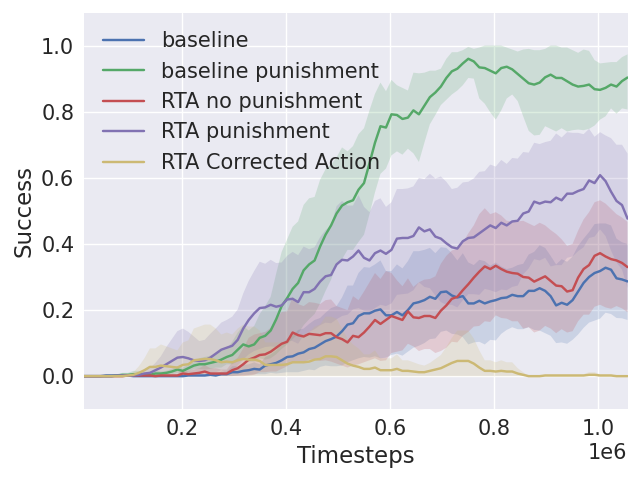}}
\caption{Results collected from experiments run in the 2D Spacecraft Docking environment with an implicit simplex RTA. Each curve represents the average 10 trials and the shaded region is the $95\%$ confidence interval about the mean. }
\label{fig:docking2d_ppo_imp_sim_results}
\end{figure}

\begin{table}[hb]
\caption{PPO 2D Spacecraft Docking Implicit Simplex}
\label{tab:Sppo2dimplicitsimplex}
\centering
\resizebox{1.0\linewidth}{!}{
\begin{tabular}{lcccccc}
\toprule
          Configuration & RTA & Return & Length & Success &      Interventions/Violations &   Correction \\
\midrule
             baseline & on &  1.7324 $\pm$ 0.5743 & 628.6310 $\pm$ 233.1921 & 0.7340 $\pm$ 0.4419 &  190.7160 $\pm$ 87.9536 & 2.5410 $\pm$ 0.4201 \\
             & off & -22.1098 $\pm$ 10.9814 & 320.5170 $\pm$ 360.2277 & 0.2660 $\pm$ 0.4419 &  143.9450 $\pm$ 79.5162 & - \\
 baseline punishment & on &  2.2181 $\pm$ 0.3315 & 709.3360 $\pm$ 158.4366 & 0.9100 $\pm$ 0.2862 &     0.6080 $\pm$ 1.1218 & 0.6372 $\pm$ 0.9141 \\
 & off &    2.1394 $\pm$ 0.3291 & 697.2640 $\pm$ 151.6598 & 0.9270 $\pm$ 0.2601 &     0.9660 $\pm$ 2.2147 & - \\
   RTA no punishment & on &  2.0223 $\pm$ 0.3853 & 483.4820 $\pm$ 159.6581 & 0.9200 $\pm$ 0.2713 &  137.0720 $\pm$ 37.9305 & 2.1929 $\pm$ 0.2068 \\
   & off & -22.3856 $\pm$ 15.3886 & 176.1880 $\pm$ 147.2843 & 0.3410 $\pm$ 0.4740 & 147.8420 $\pm$ 102.7799 & - \\
      RTA punishment & on &  1.8153 $\pm$ 0.5999 & 604.7220 $\pm$ 252.2543 & 0.7310 $\pm$ 0.4434 &  108.0690 $\pm$ 54.2185 & 2.0565 $\pm$ 0.2571 \\
      & off &  -17.9896 $\pm$ 5.1619 & 415.3470 $\pm$ 341.2807 & 0.4600 $\pm$ 0.4984 &  147.5900 $\pm$ 40.1991 & - \\
RTA Corrected Action & on & -1.3983 $\pm$ 1.2348 & 706.2610 $\pm$ 303.1083 & 0.0230 $\pm$ 0.1499 & 456.0030 $\pm$ 333.3891 & 4.8626 $\pm$ 2.1863 \\
& off & -27.0715 $\pm$ 19.0662 & 132.6590 $\pm$ 234.5261 & 0.0 $\pm$ 0.0 &  90.0710 $\pm$ 154.2038 & - \\
\bottomrule
\end{tabular}
}
\end{table}

\begin{figure}[ht]
\centering
\subfigure[SAC evaluated with RTA Average Return]{\includegraphics[width=0.45\linewidth]{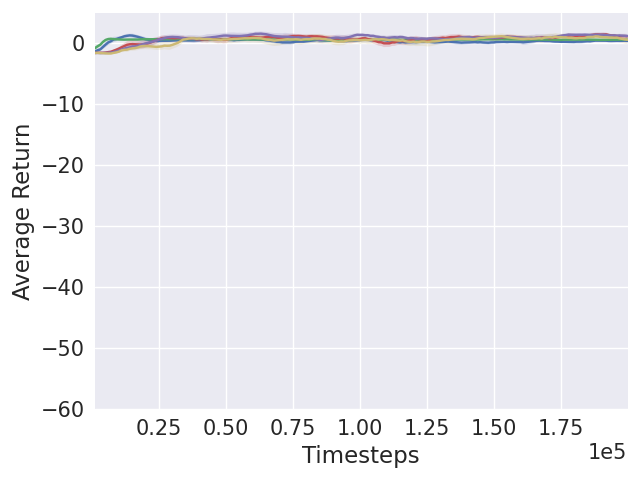}}\qquad
\subfigure[SAC evaluated with RTA Average Success]{\includegraphics[width=0.45\linewidth]{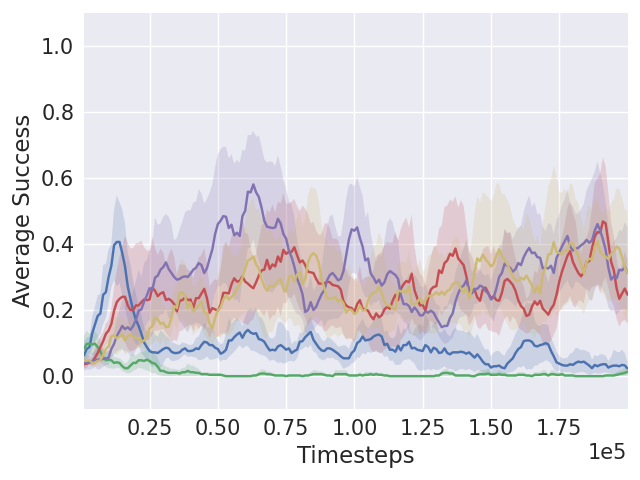}}\\
\subfigure[SAC evaluated without RTA Average Return]{\includegraphics[width=0.45\linewidth]{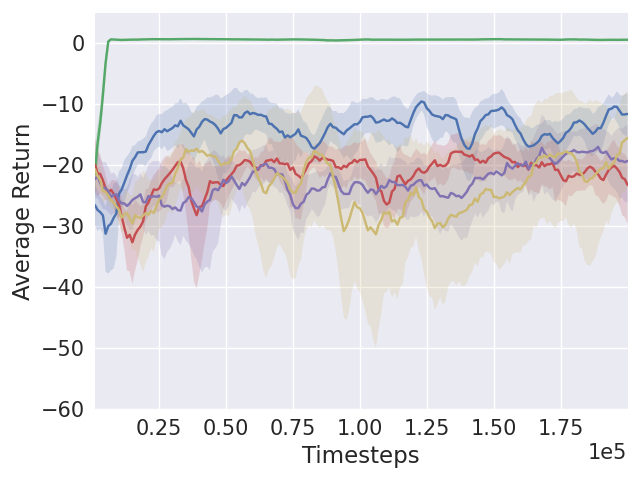}}\qquad
\subfigure[SAC evaluated without RTA Average Success]{\includegraphics[width=0.45\linewidth]{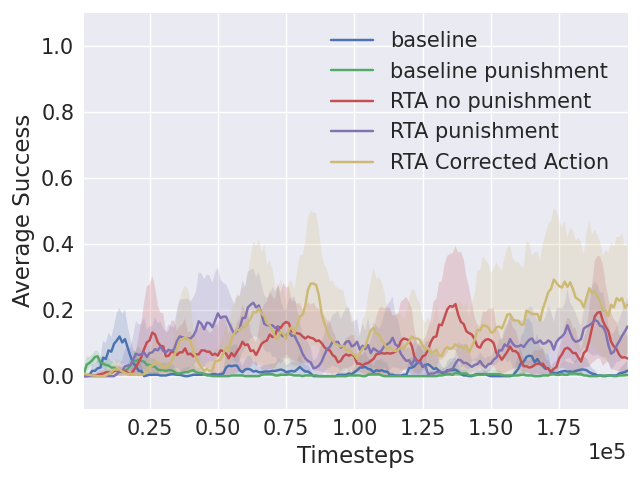}}
\caption{Results collected from experiments run in the 2D Spacecraft Docking environment with an implicit simplex RTA. Each curve represents the average 10 trials and the shaded region is the $95\%$ confidence interval about the mean.}
\label{fig:docking2d_sac_imp_sim_results}
\end{figure}

\begin{table}[hb]
\caption{SAC 2D Spacecraft Docking Implicit Simplex}
\label{tab:sac2dimplicitsimplex}
\centering
\resizebox{1.0\linewidth}{!}{
\begin{tabular}{lcccccc}
\toprule
       Configuration & RTA & Return & Length & Success &      Interventions/Violations &   Correction \\
\midrule
            baseline & on & 0.4104 $\pm$ 0.4594 & 992.1540 $\pm$ 61.4690 & 0.0170 $\pm$ 0.1293 & 40.1990 $\pm$ 29.2195 & 2.0015 $\pm$ 0.1018 \\
            & off & -36.9676 $\pm$ 17.9372 & 963.3310 $\pm$ 160.9444 & 0.0010 $\pm$ 0.0316 & 312.0830 $\pm$ 141.1573 & - \\
 baseline punishment & on & 0.2988 $\pm$ 0.4151 & 996.6570 $\pm$ 34.9140 & 0.0 $\pm$ 0.0 &   5.1930 $\pm$ 4.0115 & 1.8304 $\pm$ 0.5606 \\
 & off &   -1.1718 $\pm$ 1.6220 &  994.9620 $\pm$ 50.3511 & 0.0 $\pm$ 0.0 &   14.5570 $\pm$ 16.7634 & - \\
   RTA no punishment & on & 0.5268 $\pm$ 0.5937 & 985.8930 $\pm$ 77.3721 & 0.0260 $\pm$ 0.1591 & 45.5030 $\pm$ 14.1846 & 2.0088 $\pm$ 0.0616 \\
   & off & -49.2378 $\pm$ 26.5212 & 575.5730 $\pm$ 312.6387 & 0.0090 $\pm$ 0.0944 & 356.9840 $\pm$ 198.7087 & - \\
      RTA punishment & on & 0.4419 $\pm$ 0.6550 & 980.7460 $\pm$ 98.2139 & 0.0050 $\pm$ 0.0705 & 43.3130 $\pm$ 12.2646 & 2.0111 $\pm$ 0.0616 \\
      & off & -47.3890 $\pm$ 26.1510 & 525.6140 $\pm$ 298.1644 & 0.0090 $\pm$ 0.0944 & 341.2960 $\pm$ 193.6923 & - \\
RTA Corrected Action & on & 0.6113 $\pm$ 0.6937 & 982.4190 $\pm$ 82.0556 & 0.0410 $\pm$ 0.1983 & 46.1820 $\pm$ 21.0099 & 2.0038 $\pm$ 0.0626 \\
& off & -38.0627 $\pm$ 19.2859 & 598.8810 $\pm$ 311.1165 & 0.0350 $\pm$ 0.1838 & 286.5530 $\pm$ 145.0794 & - \\
\bottomrule
\end{tabular}
}
\end{table}

\FloatBarrier \subsection{2D Spacecraft Docking Implicit ASIF}
\label{app:docking2dimplicitasif}

\begin{figure}[ht]
\centering
\subfigure[PPO evaluated with RTA Average Return]{\includegraphics[width=0.45\linewidth]{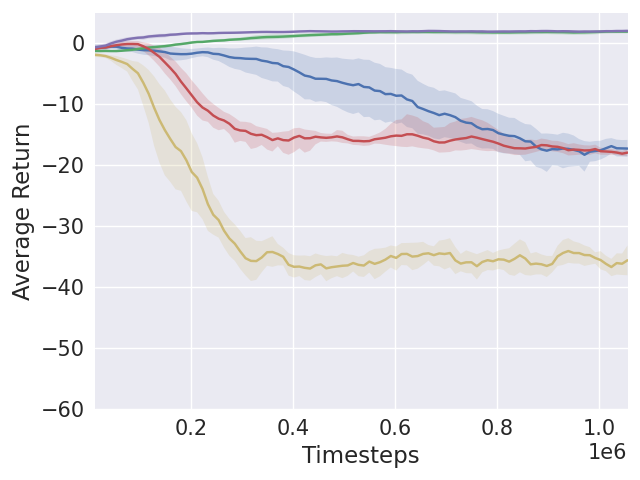}}\qquad
\subfigure[PPO evaluated with RTA Average Success]{\includegraphics[width=0.45\linewidth]{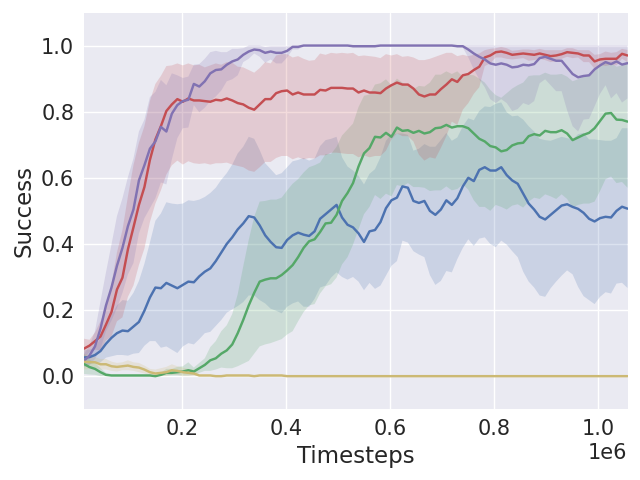}}\\
\subfigure[PPO evaluated without RTA Average Return]{\includegraphics[width=0.45\linewidth]{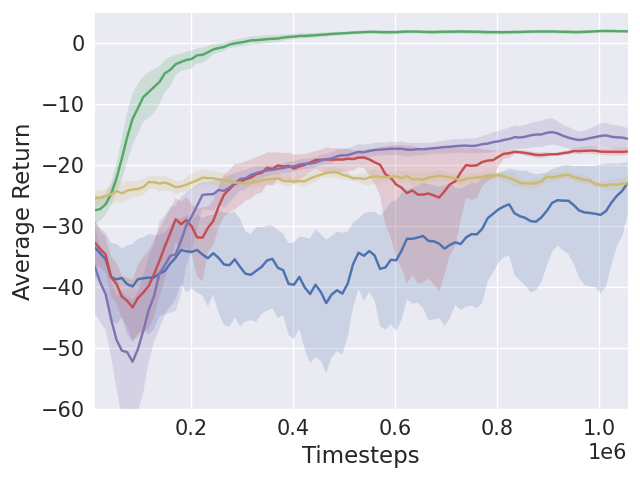}}\qquad
\subfigure[PPO evaluated without RTA Average Success]{\includegraphics[width=0.45\linewidth]{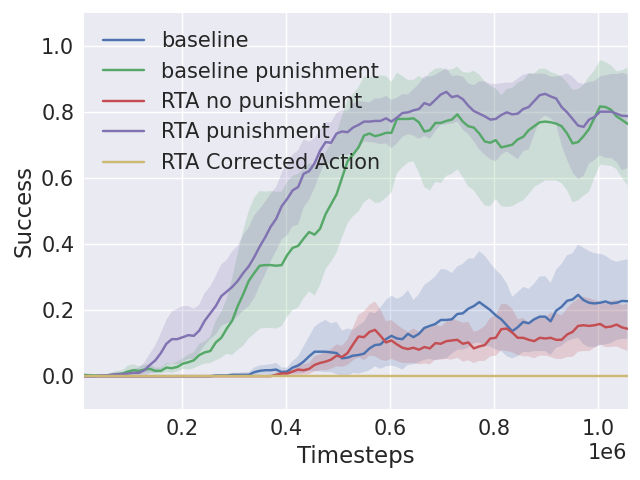}}
\caption{Results collected from experiments run in the 2D Spacecraft Docking environment with an implicit ASIF RTA. Each curve represents the average 10 trials and the shaded region is the $95\%$ confidence interval about the mean. }
\label{fig:docking2d_ppo_imp_asif_results}
\end{figure}

\begin{table}[hb]
\caption{PPO 2D Spacecraft Docking Implicit ASIF}
\label{tab:Sppo2dimplicitasif}
\centering
\resizebox{1.0\linewidth}{!}{
\begin{tabular}{lcccccc}
\toprule
       Configuration & RTA & Return & Length & Success &      Interventions/Violations &   Correction \\
\midrule
            baseline & on &  -17.7031 $\pm$ 8.5565 & 601.9800 $\pm$ 318.1612 & 0.4870 $\pm$ 0.4998 &  337.4270 $\pm$ 80.0928 & 0.8393 $\pm$ 0.2897 \\
            & off & -23.4315 $\pm$ 13.5040 & 385.0710 $\pm$ 393.7963 & 0.1980 $\pm$ 0.3985 & 143.9770 $\pm$ 99.2892 & - \\
 baseline punishment & on &    1.8864 $\pm$ 0.5720 & 751.0750 $\pm$ 157.2639 & 0.7810 $\pm$ 0.4136 &  138.1540 $\pm$ 44.0523 & 0.2437 $\pm$ 0.0398 \\
 & off &    1.8927 $\pm$ 0.5791 & 733.8830 $\pm$ 175.6035 & 0.7550 $\pm$ 0.4301 &    0.9690 $\pm$ 1.8868 & - \\
   RTA no punishment & on &  -17.5369 $\pm$ 4.5902 &  329.2390 $\pm$ 34.7500 & 0.9610 $\pm$ 0.1936 &  290.2000 $\pm$ 32.0983 & 0.8294 $\pm$ 0.1533 \\
   & off &  -17.8740 $\pm$ 2.3042 &  123.9470 $\pm$ 36.0346 & 0.1880 $\pm$ 0.3907 & 114.9080 $\pm$ 25.0518 & - \\
      RTA punishment & on &    1.9883 $\pm$ 0.3713 & 502.9990 $\pm$ 143.2604 & 0.9480 $\pm$ 0.2220 &  204.7170 $\pm$ 37.7863 & 0.2781 $\pm$ 0.0398 \\
      & off &  -15.8753 $\pm$ 5.1271 & 364.4110 $\pm$ 181.5286 & 0.7690 $\pm$ 0.4215 & 156.4490 $\pm$ 41.2289 & - \\
RTA Corrected Action & on & -36.3506 $\pm$ 21.1084 & 335.6310 $\pm$ 192.6612 & 0.0 $\pm$ 0.0 & 335.6310 $\pm$ 192.6612 & 9.6171 $\pm$ 3.2334 \\
& off &  -22.0632 $\pm$ 9.6757 &    21.0460 $\pm$ 7.2165 & 0.0 $\pm$ 0.0 &   20.8590 $\pm$ 7.0182 & - \\
\bottomrule
\end{tabular}
}
\end{table}

\begin{figure}[ht]
\centering
\subfigure[SAC evaluated with RTA Average Return]{\includegraphics[width=0.45\linewidth]{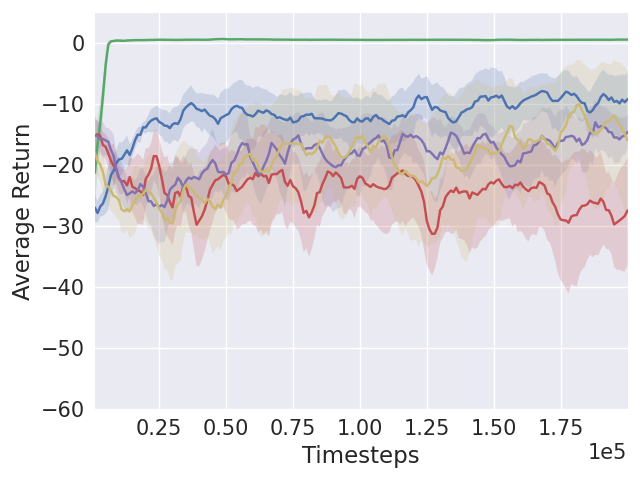}}\qquad
\subfigure[SAC evaluated with RTA Average Success]{\includegraphics[width=0.45\linewidth]{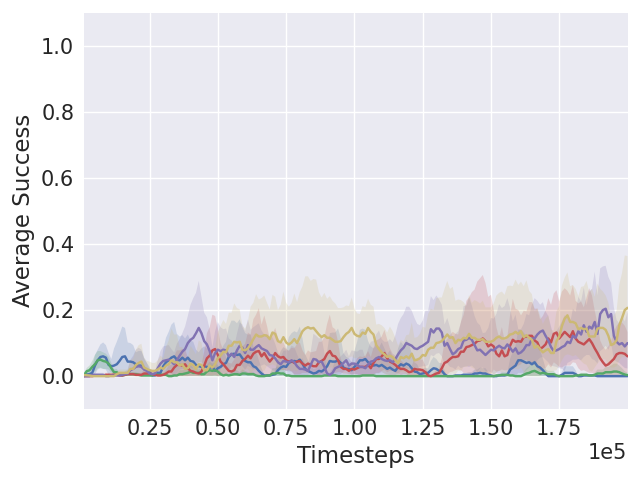}}\\
\subfigure[SAC evaluated without RTA Average Return]{\includegraphics[width=0.45\linewidth]{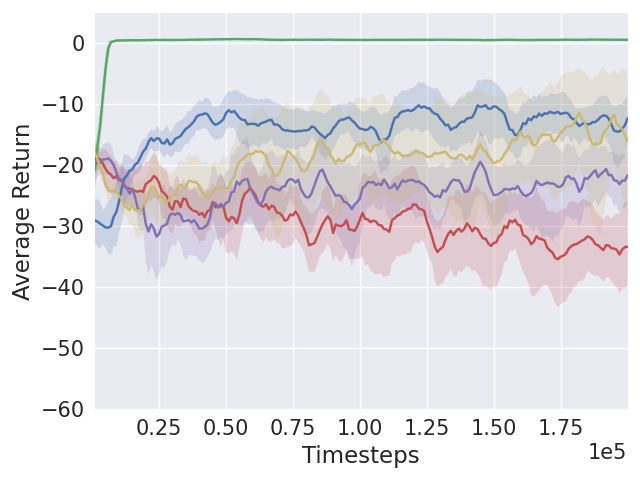}}\qquad
\subfigure[SAC evaluated without RTA Average Success]{\includegraphics[width=0.45\linewidth]{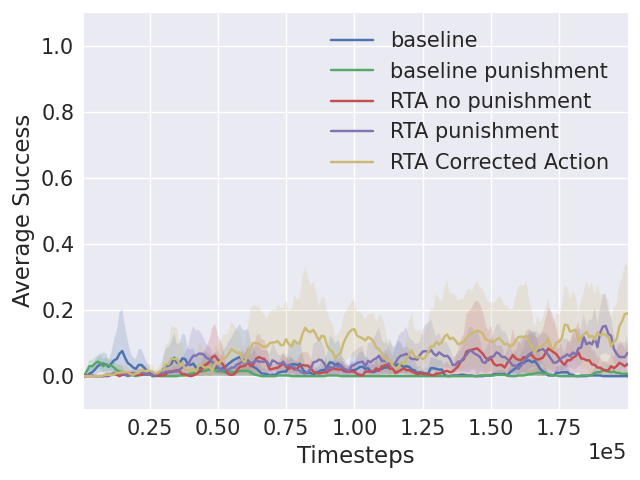}}
\caption{Results collected from experiments run in the 2D Spacecraft Docking environment with an implicit ASIF RTA. Each curve represents the average 10 trials and the shaded region is the $95\%$ confidence interval about the mean.}
\label{fig:docking2d_sac_imp_asif_results}
\end{figure}

\begin{table}[hb]
\caption{SAC 2D Spacecraft Docking Implicit ASIF}
\label{tab:sac2dimplicitasif}
\centering
\resizebox{1.0\linewidth}{!}{
\begin{tabular}{lcccccc}
\toprule
       Configuration & RTA & Return & Length & Success &      Interventions/Violations &   Correction \\
\midrule
            baseline & on &  0.0170 $\pm$ 0.6159 & 988.4880 $\pm$ 66.0788 & 0.0220 $\pm$ 0.1467 & 327.1520 $\pm$ 78.9583 & 0.3659 $\pm$ 0.0418 \\
            & off & -40.4752 $\pm$ 19.6528 & 945.6220 $\pm$ 187.7012 & 0.0030 $\pm$ 0.0547 & 335.9500 $\pm$ 154.2188 & - \\
 baseline punishment & on & -0.0926 $\pm$ 0.3824 & 998.7810 $\pm$ 21.0513 & 0.0 $\pm$ 0.0 & 242.3540 $\pm$ 94.6843 & 0.2916 $\pm$ 0.0280 \\
 & off &   -0.6723 $\pm$ 1.0088 &  998.8650 $\pm$ 18.3413 & 0.0 $\pm$ 0.0 &    9.5740 $\pm$ 10.5377 & - \\
   RTA no punishment & on &  0.0072 $\pm$ 0.5972 & 989.0330 $\pm$ 63.0413 & 0.0120 $\pm$ 0.1089 & 337.4620 $\pm$ 70.4624 & 0.3637 $\pm$ 0.0308 \\
   & off & -40.3231 $\pm$ 22.8186 & 388.8060 $\pm$ 237.8237 & 0.0030 $\pm$ 0.0547 & 266.6250 $\pm$ 160.5296 & - \\
      RTA punishment & on & -0.0248 $\pm$ 0.5306 & 992.9850 $\pm$ 54.3038 & 0.0020 $\pm$ 0.0447 & 322.7700 $\pm$ 69.0018 & 0.3560 $\pm$ 0.0272 \\
      & off & -45.2013 $\pm$ 24.9335 & 500.5040 $\pm$ 285.1929 & 0.0020 $\pm$ 0.0447 & 319.9860 $\pm$ 182.5422 & - \\
RTA Corrected Action & on &  0.0461 $\pm$ 0.7008 & 984.5460 $\pm$ 75.3451 & 0.0210 $\pm$ 0.1434 & 317.3800 $\pm$ 76.3632 & 0.3630 $\pm$ 0.0405 \\
& off & -38.1003 $\pm$ 20.0277 & 474.9430 $\pm$ 292.3608 & 0.0110 $\pm$ 0.1043 & 261.0650 $\pm$ 143.3972 & - \\
\bottomrule
\end{tabular}
}
\end{table}


\FloatBarrier \subsection{3D Spacecraft Docking Explicit Simplex}
\label{app:docking3dexplicitsimplex}

\begin{figure}[ht]
\centering
\subfigure[PPO evaluated with RTA Average Return]{\includegraphics[width=0.45\linewidth]{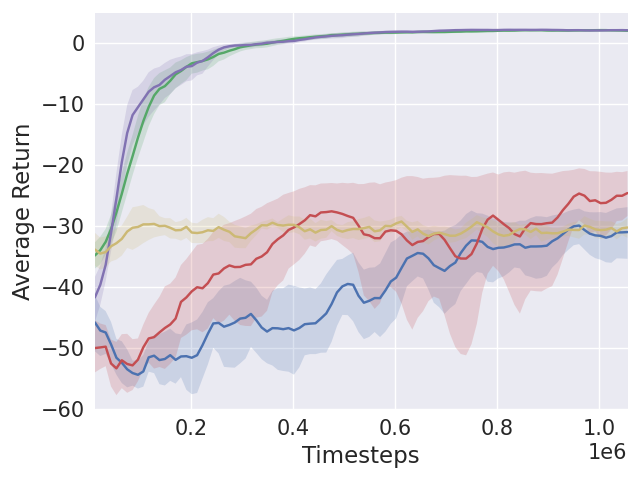}}\qquad
\subfigure[PPO evaluated with RTA Average Success]{\includegraphics[width=0.45\linewidth]{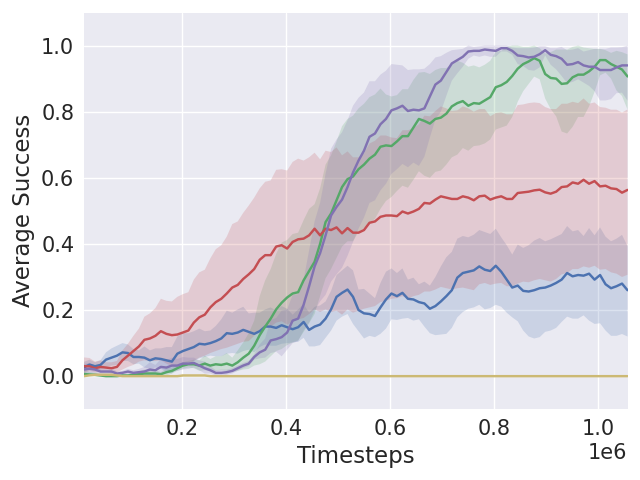}}\\
\subfigure[PPO evaluated without RTA Average Return]{\includegraphics[width=0.45\linewidth]{figures/docking3d_explicit_simplex/ppo_no_rta_eval.png}}\qquad
\subfigure[PPO evaluated without RTA Average Success]{\includegraphics[width=0.45\linewidth]{figures/docking3d_explicit_simplex/ppo_no_rta_success.png}}
\caption{Results collected from experiments run in the Docking3D environment with an explicit simplex RTA. Each curve represents the average 10 trials and the shaded region is the $95\%$ confidence interval about the mean.}
\label{fig:docking3d_ppo_exp_sim_results}
\end{figure}

\begin{table}[hb]
\caption{PPO 3D Spacecraft Docking Explicit Simplex}
\label{tab:ppo3dexplicitsimplex}
\centering
\resizebox{1.0\linewidth}{!}{
\begin{tabular}{lcccccc}
\toprule
       Configuration & RTA & Return & Length & Success &      Interventions/Violations &   Correction \\
\midrule
            baseline & on & -31.0188 $\pm$ 10.6629 & 518.1790 $\pm$ 321.9829 & 0.2820 $\pm$ 0.4500 &  193.3420 $\pm$ 67.8229 &  0.8483 $\pm$ 0.2922 \\
            & off & -23.6606 $\pm$ 10.0484 & 387.3470 $\pm$ 368.7029 & 0.2320 $\pm$ 0.4221 & 153.5120 $\pm$ 81.3324 & - \\
 baseline punishment & on &    2.0348 $\pm$ 0.3485 & 700.8320 $\pm$ 150.7818 & 0.8990 $\pm$ 0.3013 &     0.0 $\pm$ 0.0 &  0.0 $\pm$ 0.0 \\
 & off &    2.0182 $\pm$ 0.3761 & 714.7290 $\pm$ 147.2671 & 0.8870 $\pm$ 0.3166 &    0.9160 $\pm$ 1.9034 & - \\
   RTA no punishment & on & -24.4788 $\pm$ 10.7840 &  280.2370 $\pm$ 88.3712 & 0.5980 $\pm$ 0.4903 & 149.2880 $\pm$ 101.5997 &  0.9711 $\pm$ 0.8016 \\
    & off & -21.8075 $\pm$ 10.9196 & 190.2490 $\pm$ 111.3772 & 0.5510 $\pm$ 0.4974 & 134.1560 $\pm$ 72.1069 & - \\
      RTA punishment & on &    2.0870 $\pm$ 0.3968 & 687.3150 $\pm$ 135.2182 & 0.9360 $\pm$ 0.2448 &     0.0 $\pm$ 0.0 &  0.0 $\pm$ 0.0 \\
      & off &    2.0874 $\pm$ 0.3910 & 685.5820 $\pm$ 136.5485 & 0.9330 $\pm$ 0.2500 &    1.0390 $\pm$ 2.3608 & - \\
RTA Corrected Action & on & -30.8903 $\pm$ 14.7595 & 230.2190 $\pm$ 122.1902 & 0.0 $\pm$ 0.0 & 230.2190 $\pm$ 122.1902 & 19.4718 $\pm$ 3.4466 \\
& off &  -22.6890 $\pm$ 8.2845 &    14.7530 $\pm$ 3.3308 & 0.0 $\pm$ 0.0 &   14.7530 $\pm$ 3.3308 & - \\
\bottomrule
\end{tabular}
}
\end{table}

\begin{figure}[ht]
\centering
\subfigure[SAC evaluated with RTA Average Return]{\includegraphics[width=0.45\linewidth]{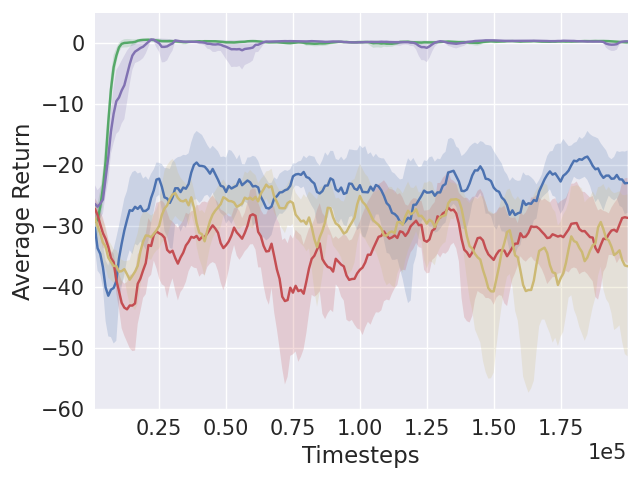}}\qquad
\subfigure[SAC evaluated with RTA Average Success]{\includegraphics[width=0.45\linewidth]{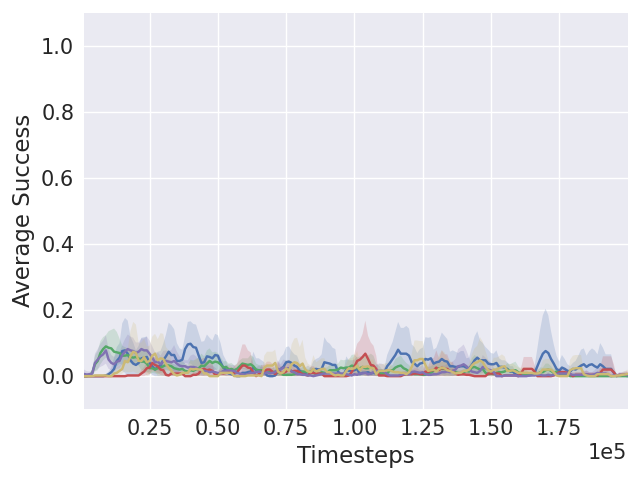}}\\
\subfigure[SAC evaluated without RTA Average Return]{\includegraphics[width=0.45\linewidth]{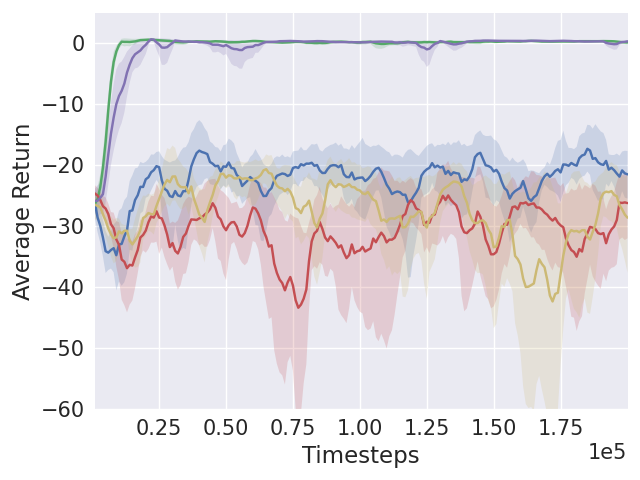}}\qquad
\subfigure[SAC evaluated without RTA Average Success]{\includegraphics[width=0.45\linewidth]{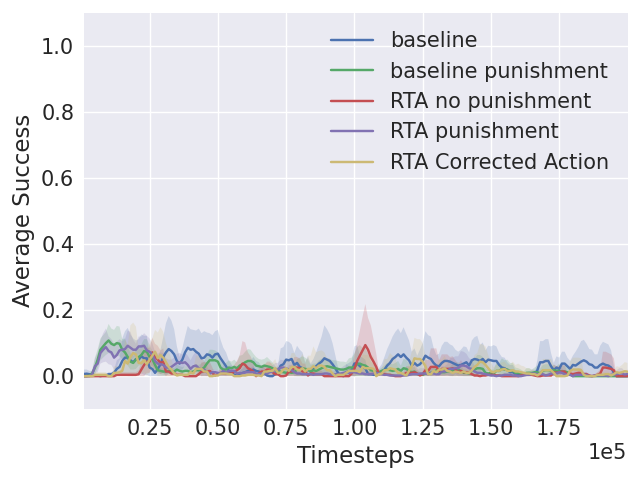}}
\caption{Results collected from experiments run in the Docking3D environment with an explicit simplex RTA. Each curve represents the average 10 trials and the shaded region is the $95\%$ confidence interval about the mean.}
\label{fig:docking3d_sac_exp_sim_results}
\end{figure}

\begin{table}[hb]
\caption{SAC 3D Spacecraft Docking Explicit Simplex}
\label{tab:sac3dexplicitsimplex}
\centering
\resizebox{1.0\linewidth}{!}{
\begin{tabular}{lcccccc}
\toprule
       Configuration & RTA & Return & Length & Success &      Interventions/Violations &   Correction \\
\midrule
            baseline & on & -45.5176 $\pm$ 17.8207 & 889.6050 $\pm$ 233.6027 & 0.0210 $\pm$ 0.1434 &   81.5790 $\pm$ 65.0452 & 0.4107 $\pm$ 0.1040 \\
            & off & -51.7261 $\pm$ 22.6411 & 793.0560 $\pm$ 312.3463 & 0.0050 $\pm$ 0.0705 & 404.6600 $\pm$ 179.4067 & - \\
 baseline punishment & on &   -3.1201 $\pm$ 3.3789 &  984.0890 $\pm$ 78.1228 & 0.0 $\pm$ 0.0 &     0.5910 $\pm$ 2.4831 & 0.0378 $\pm$ 0.1168 \\
 & off &   -3.0951 $\pm$ 3.7081 &  985.4280 $\pm$ 66.5032 & 0.0010 $\pm$ 0.0316 &   29.8100 $\pm$ 35.5730 & - \\
   RTA no punishment & on & -56.2666 $\pm$ 18.2059 & 894.0790 $\pm$ 228.7942 & 0.0340 $\pm$ 0.1812 &  113.1790 $\pm$ 63.4759 & 0.4365 $\pm$ 0.0638 \\
   & off & -65.2064 $\pm$ 30.6350 & 749.4010 $\pm$ 336.3577 & 0.0020 $\pm$ 0.0447 & 478.7300 $\pm$ 229.1926 & - \\
      RTA punishment & on &   -3.2459 $\pm$ 4.9680 &  997.0710 $\pm$ 27.9703 & 0.0050 $\pm$ 0.0705 &     1.6550 $\pm$ 6.6895 & 0.0429 $\pm$ 0.1162 \\
      & off &   -3.5462 $\pm$ 6.0729 &  995.9100 $\pm$ 36.4424 & 0.0050 $\pm$ 0.0705 &   35.5580 $\pm$ 58.1211 & - \\
RTA Corrected Action & on & -57.1892 $\pm$ 23.8735 & 874.9890 $\pm$ 254.3602 & 0.0230 $\pm$ 0.1499 & 134.5720 $\pm$ 111.9885 & 0.4541 $\pm$ 0.1048 \\
& off & -53.6893 $\pm$ 28.7788 & 639.5200 $\pm$ 380.3689 & 0.0030 $\pm$ 0.0547 & 379.5430 $\pm$ 226.4612 & - \\
\bottomrule
\end{tabular}
}
\end{table}

\FloatBarrier \subsection{3D Spacecraft Docking Explicit ASIF}
\label{app:docking3dexplicitasif}

\begin{figure}[ht]
\centering
\subfigure[PPO evaluated with RTA Average Return]{\includegraphics[width=0.45\linewidth]{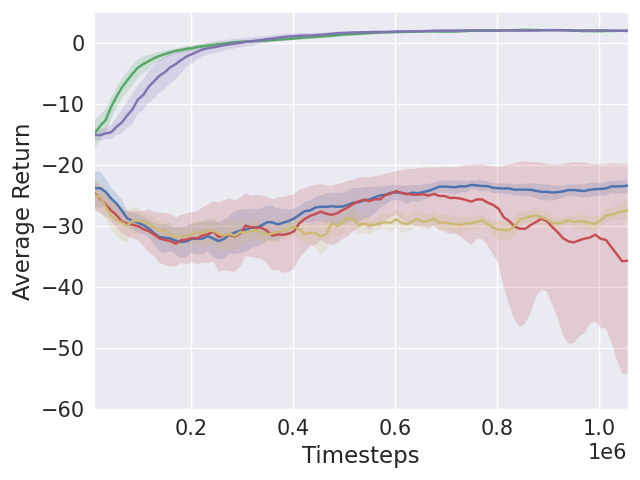}}\qquad
\subfigure[PPO evaluated with RTA Average Success]{\includegraphics[width=0.45\linewidth]{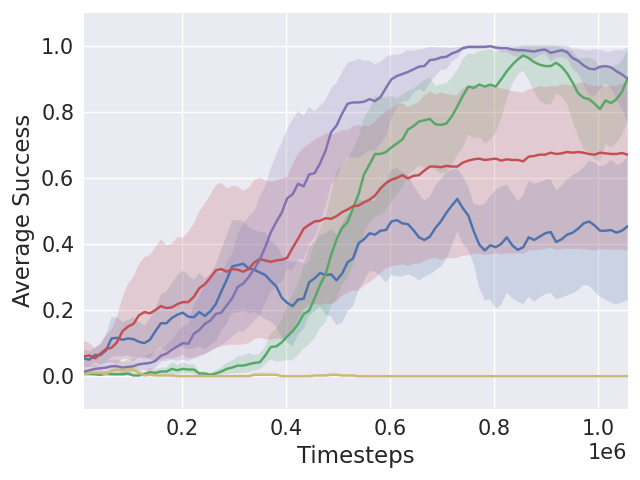}}\\
\subfigure[PPO evaluated without RTA Average Return]{\includegraphics[width=0.45\linewidth]{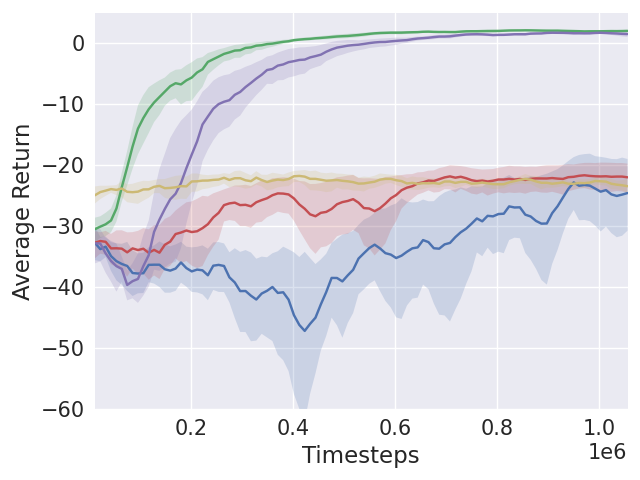}}\qquad
\subfigure[PPO evaluated without RTA Average Success]{\includegraphics[width=0.45\linewidth]{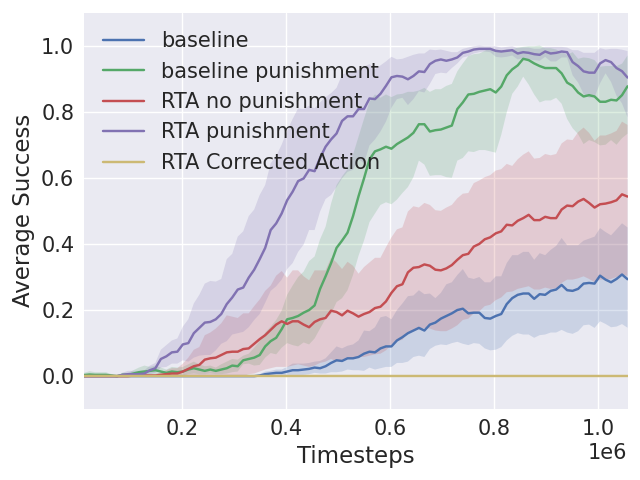}}
\caption{Results collected from experiments run in the Docking3D environment with an explicit ASIF RTA. Each curve represents the average 10 trials and the shaded region is the $95\%$ confidence interval about the mean. }
\label{fig:docking3d_ppo_exp_asif_results}
\end{figure}

\begin{table}[hb]
\caption{PPO 3D Spacecraft Docking Explicit ASIF}
\label{tab:ppo3dexplicitasif}
\centering
\resizebox{1.0\linewidth}{!}{
\begin{tabular}{lcccccc}
\toprule
       Configuration & RTA & Return & Length & Success &      Interventions/Violations &   Correction \\
\midrule
            baseline & on &  -23.5395 $\pm$ 5.5692 & 520.2150 $\pm$ 335.7678 & 0.4380 $\pm$ 0.4961 &  262.1080 $\pm$ 81.3717 &  0.7496 $\pm$ 0.2788 \\
            & off & -25.9051 $\pm$ 15.2182 & 395.0710 $\pm$ 393.6692 & 0.2810 $\pm$ 0.4495 & 174.5970 $\pm$ 127.5959 & - \\
 baseline punishment & on &    2.0252 $\pm$ 0.4779 & 719.1760 $\pm$ 154.5015 & 0.8760 $\pm$ 0.3296 &   61.9650 $\pm$ 14.8473 &  0.1813 $\pm$ 0.0324 \\
 & off &    1.9796 $\pm$ 0.5287 & 699.8420 $\pm$ 156.6811 & 0.8890 $\pm$ 0.3141 &     1.2990 $\pm$ 3.2802 & - \\
   RTA no punishment & on & -35.7737 $\pm$ 31.4269 & 385.1380 $\pm$ 255.2133 & 0.6760 $\pm$ 0.4680 & 295.9280 $\pm$ 214.9787 &  0.8228 $\pm$ 0.4336 \\
   & off &  -22.1710 $\pm$ 6.5638 &  161.8100 $\pm$ 77.1825 & 0.5380 $\pm$ 0.4986 &  133.3550 $\pm$ 59.4011 & - \\
      RTA punishment & on &    1.9614 $\pm$ 0.5390 & 678.1530 $\pm$ 155.3601 & 0.8860 $\pm$ 0.3178 &   65.2410 $\pm$ 17.8011 &  0.1843 $\pm$ 0.0306 \\
      & off &    1.5387 $\pm$ 1.0994 & 667.7730 $\pm$ 160.3803 & 0.8910 $\pm$ 0.3116 &    6.0220 $\pm$ 10.9016 & - \\
RTA Corrected Action & on & -28.6656 $\pm$ 12.8821 &  196.4600 $\pm$ 92.5018 & 0.0 $\pm$ 0.0 &  196.4600 $\pm$ 92.5018 & 18.3608 $\pm$ 2.8570 \\
& off &  -22.9620 $\pm$ 8.5660 &    14.2520 $\pm$ 3.0771 & 0.0 $\pm$ 0.0 &    14.2440 $\pm$ 3.0875 & - \\
\bottomrule
\end{tabular}
}
\end{table}

\begin{figure}[ht]
\centering
\subfigure[SAC evaluated with RTA Average Return]{\includegraphics[width=0.45\linewidth]{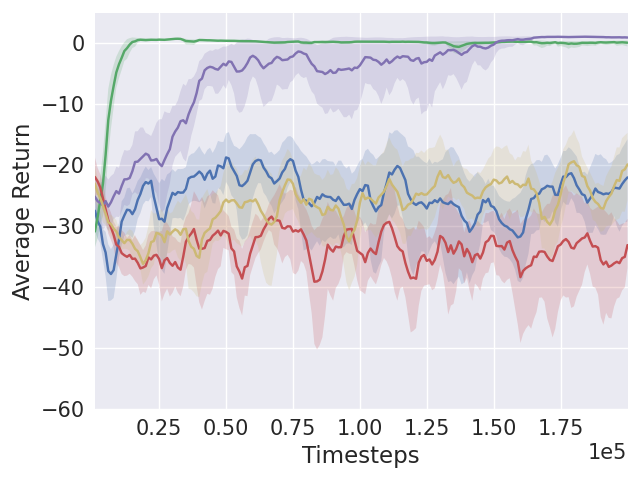}}\qquad
\subfigure[SAC evaluated with RTA Average Success]{\includegraphics[width=0.45\linewidth]{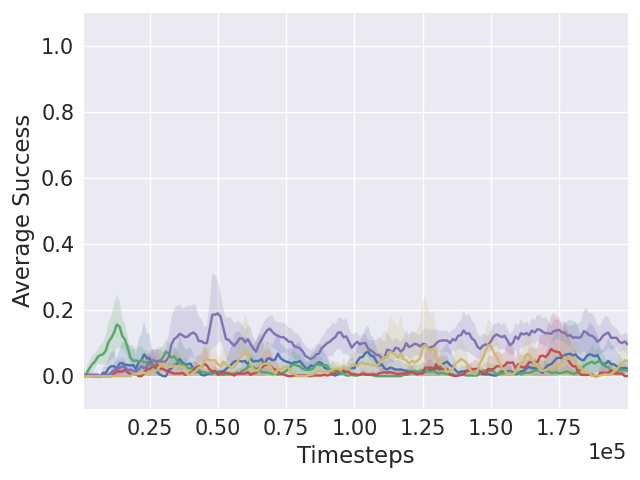}}\\
\subfigure[SAC evaluated without RTA Average Return]{\includegraphics[width=0.45\linewidth]{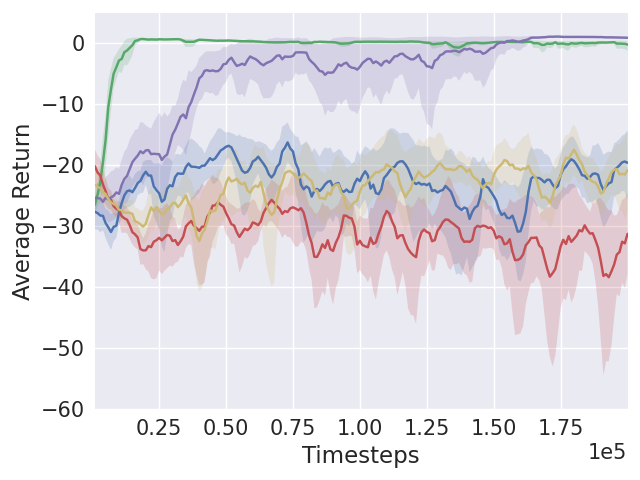}}\qquad
\subfigure[SAC evaluated without RTA Average Success]{\includegraphics[width=0.45\linewidth]{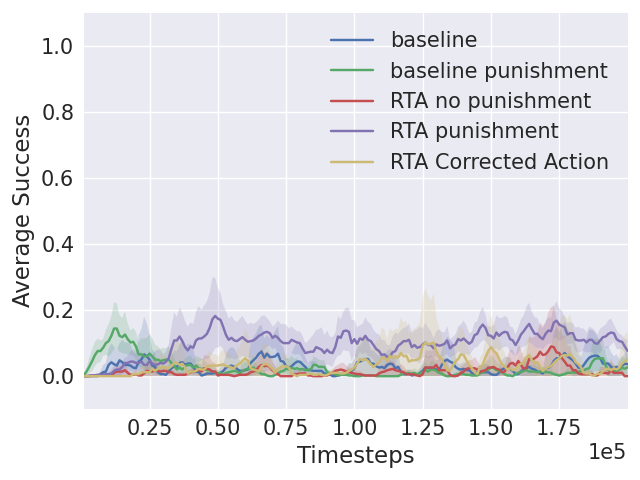}}
\caption{Results collected from experiments run in the Docking3D environment with an explicit ASIF RTA. Each curve represents the average 10 trials and the shaded region is the $95\%$ confidence interval about the mean.}
\label{fig:docking3d_sac_exp_asif_results}
\end{figure}

\begin{table}[hb]
\caption{SAC 3D Spacecraft Docking Explicit ASIF}
\label{tab:sac3ddexplicitasif}
\centering
\resizebox{1.0\linewidth}{!}{
\begin{tabular}{lcccccc}
\toprule
       Configuration & RTA & Return & Length & Success &      Interventions/Violations &   Correction \\
\midrule
            baseline & on & -12.7579 $\pm$ 10.1785 & 915.5070 $\pm$ 202.0146 & 0.1110 $\pm$ 0.3141 & 204.1420 $\pm$ 90.2193 & 0.3788 $\pm$ 0.0620 \\
            & off & -38.0083 $\pm$ 20.4644 & 854.8270 $\pm$ 283.0951 & 0.0410 $\pm$ 0.1983 & 304.2500 $\pm$ 160.4455 & - \\
 baseline punishment & on &   -0.1637 $\pm$ 0.6890 &  987.1430 $\pm$ 68.1994 & 0.0010 $\pm$ 0.0316 &  60.4100 $\pm$ 32.5684 & 0.2518 $\pm$ 0.0390 \\
 & off &   -2.8927 $\pm$ 3.6149 &  986.2250 $\pm$ 70.9115 & 0.0050 $\pm$ 0.0705 &   29.0850 $\pm$ 35.7144 & - \\
   RTA no punishment & on &  -14.5421 $\pm$ 7.3016 & 940.2470 $\pm$ 159.9236 & 0.0780 $\pm$ 0.2682 & 252.0550 $\pm$ 65.3287 & 0.3840 $\pm$ 0.0382 \\
   & off & -56.6587 $\pm$ 29.6890 & 568.9540 $\pm$ 316.5479 & 0.0050 $\pm$ 0.0705 & 399.5180 $\pm$ 211.1678 & - \\
      RTA punishment & on &   -2.7094 $\pm$ 1.9604 &  999.2260 $\pm$ 16.0357 & 0.0 $\pm$ 0.0 & 156.4860 $\pm$ 42.5640 & 0.3229 $\pm$ 0.0292 \\
      & off & -29.8980 $\pm$ 14.0208 & 947.3790 $\pm$ 146.0030 & 0.0180 $\pm$ 0.1330 & 270.2530 $\pm$ 120.9608 & - \\
RTA Corrected Action & on &  -12.3856 $\pm$ 9.0203 & 905.9600 $\pm$ 213.5613 & 0.0410 $\pm$ 0.1983 & 217.5670 $\pm$ 78.7771 & 0.3668 $\pm$ 0.0465 \\
& off & -48.8981 $\pm$ 23.9236 & 687.4480 $\pm$ 355.5327 & 0.0080 $\pm$ 0.0891 & 375.6960 $\pm$ 195.4146 & - \\
\bottomrule
\end{tabular}
}
\end{table}

\FloatBarrier \subsection{3D Spacecraft Docking Implicit Simplex}
\label{app:docking3dimplicitsimplex}

\begin{figure}[ht]
\centering
\subfigure[PPO evaluated with RTA Average Return]{\includegraphics[width=0.45\linewidth]{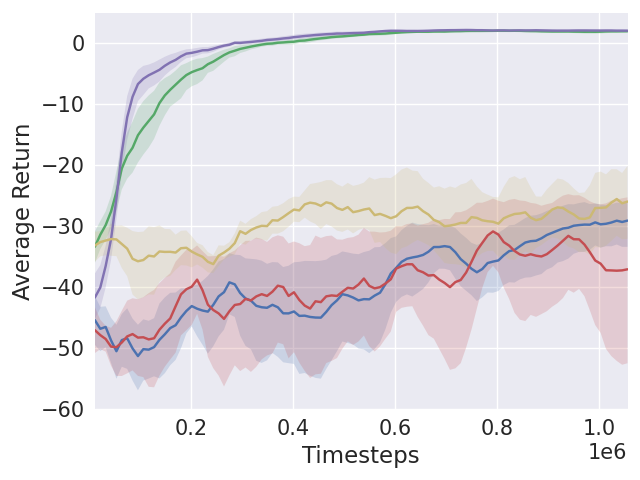}}\qquad
\subfigure[PPO evaluated with RTA Average Success]{\includegraphics[width=0.45\linewidth]{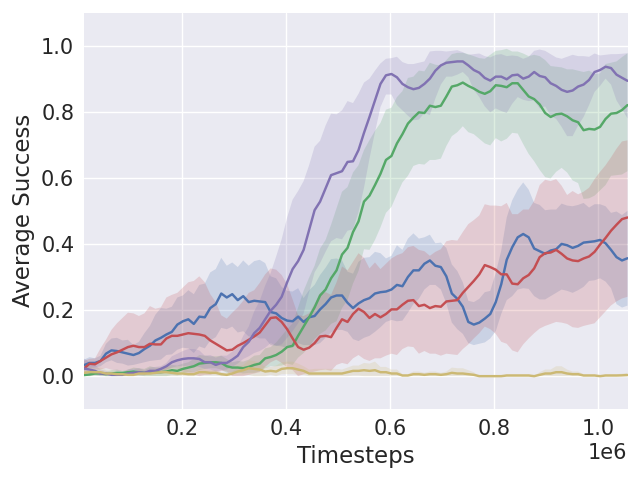}}\\
\subfigure[PPO evaluated without RTA Average Return]{\includegraphics[width=0.45\linewidth]{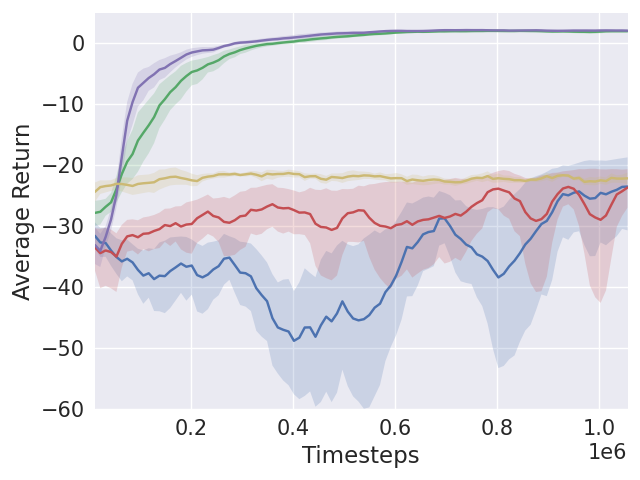}}\qquad
\subfigure[PPO evaluated without RTA Average Success]{\includegraphics[width=0.45\linewidth]{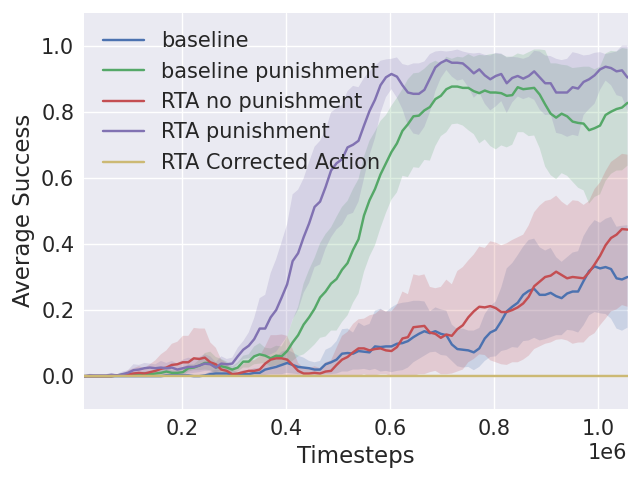}}
\caption{Results collected from experiments run in the Docking3D environment with an implicit simplex RTA. Each curve represents the average 10 trials and the shaded region is the $95\%$ confidence interval about the mean. }
\label{fig:docking3d_ppo_imp_sim_results}
\end{figure}

\begin{table}[hb]
\caption{PPO 3D Spacecraft Docking Implicit Simplex}
\label{tab:ppo3dimplicitsimplex}
\centering
\resizebox{1.0\linewidth}{!}{
\begin{tabular}{lcccccc}
\toprule
       Configuration & RTA & Return & Length & Success &      Interventions/Violations &   Correction \\
\midrule
            baseline & on &  -28.8368 $\pm$ 9.1682 & 444.4520 $\pm$ 311.7484 & 0.3520 $\pm$ 0.4776 &  177.5300 $\pm$ 58.5945 &  1.2214 $\pm$ 0.2807 \\
            & off & -23.7356 $\pm$ 14.1796 & 339.6470 $\pm$ 368.3187 & 0.2710 $\pm$ 0.4445 & 148.9460 $\pm$ 106.5690 & - \\
 baseline punishment & on &    1.9298 $\pm$ 0.4878 & 729.8170 $\pm$ 164.2315 & 0.8050 $\pm$ 0.3962 &     0.0 $\pm$ 0.0 &  0.0 $\pm$ 0.0 \\
 & off &    1.9062 $\pm$ 0.4930 & 741.4000 $\pm$ 163.1066 & 0.7880 $\pm$ 0.4087 &     0.8780 $\pm$ 1.9645 & - \\
   RTA no punishment & on & -33.2941 $\pm$ 22.1803 & 388.1860 $\pm$ 222.9841 & 0.5020 $\pm$ 0.5000 & 152.7050 $\pm$ 104.8435 &  1.2013 $\pm$ 0.4864 \\
   & off &  -23.5098 $\pm$ 8.5427 & 214.6090 $\pm$ 125.9420 & 0.4380 $\pm$ 0.4961 &  152.0270 $\pm$ 57.3112 & - \\
      RTA punishment & on &    2.0360 $\pm$ 0.3864 & 722.1020 $\pm$ 156.9345 & 0.8900 $\pm$ 0.3129 &     0.0010 $\pm$ 0.0316 &  0.0008 $\pm$ 0.0247 \\
      & off &    2.0734 $\pm$ 0.3357 & 703.6070 $\pm$ 152.0538 & 0.9190 $\pm$ 0.2728 &     1.1600 $\pm$ 2.1805 & - \\
RTA Corrected Action & on & -12.6732 $\pm$ 15.6686 & 705.8490 $\pm$ 305.7374 & 0.0290 $\pm$ 0.1678 & 705.4710 $\pm$ 305.5783 & 18.1370 $\pm$ 3.7118 \\
& off &  -22.2881 $\pm$ 8.1576 &    15.2590 $\pm$ 3.6905 & 0.0 $\pm$ 0.0 &    15.2590 $\pm$ 3.6905 & - \\
\bottomrule
\end{tabular}
}
\end{table}

\begin{figure}[ht]
\centering
\subfigure[SAC evaluated with RTA Average Return]{\includegraphics[width=0.45\linewidth]{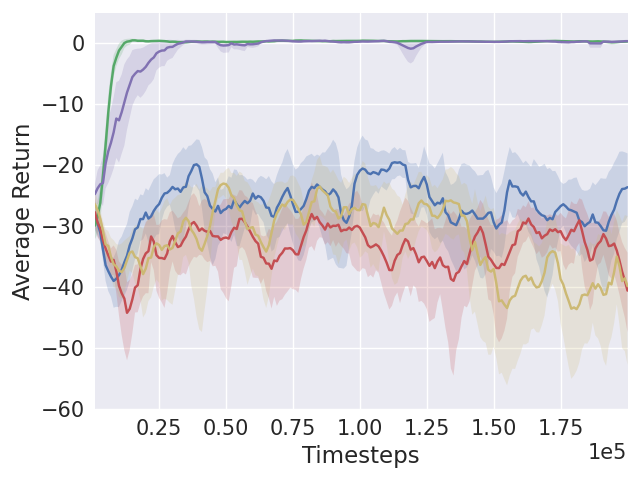}}\qquad
\subfigure[SAC evaluated with RTA Average Success]{\includegraphics[width=0.45\linewidth]{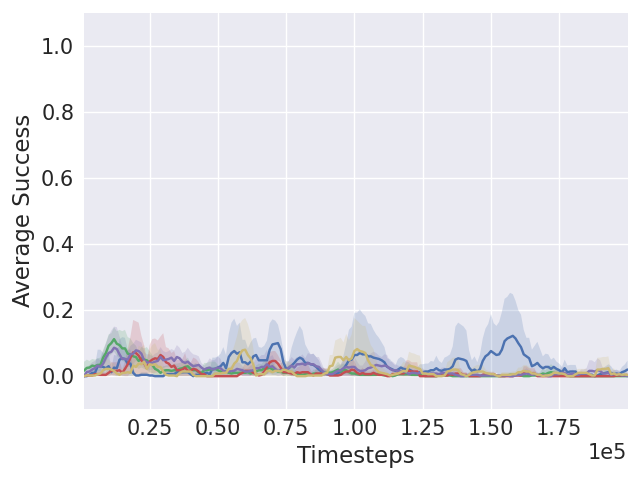}}\\
\subfigure[SAC evaluated without RTA Average Return]{\includegraphics[width=0.45\linewidth]{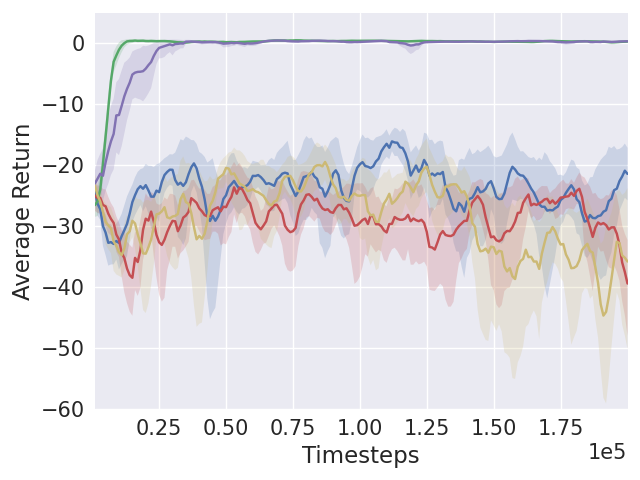}}\qquad
\subfigure[SAC evaluated without RTA Average Success]{\includegraphics[width=0.45\linewidth]{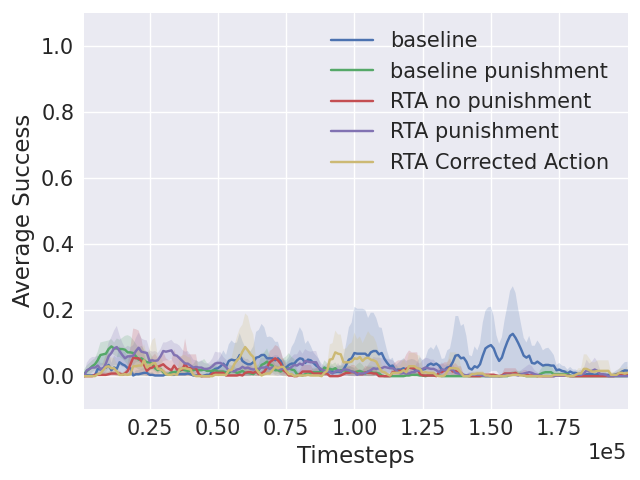}}
\caption{Results collected from experiments run in the Docking3D environment with an implicit simplex RTA. Each curve represents the average 10 trials and the shaded region is the $95\%$ confidence interval about the mean.}
\label{fig:docking3d_sac_imp_sim_results}
\end{figure}

\begin{table}[hb]
\caption{SAC 3D Spacecraft Docking Implicit Simplex}
\label{tab:sac3dimplicitsimplex}
\centering
\resizebox{1.0\linewidth}{!}{
\begin{tabular}{lcccccc}
\toprule
       Configuration & RTA & Return & Length & Success &      Interventions/Violations &   Correction \\
\midrule
            baseline & on & -39.8408 $\pm$ 15.1098 & 853.8830 $\pm$ 271.8064 & 0.0480 $\pm$ 0.2138 &   49.7620 $\pm$ 37.1512 & 1.0700 $\pm$ 0.2004 \\
            & off & -49.9209 $\pm$ 23.8915 & 772.6130 $\pm$ 329.1146 & 0.0090 $\pm$ 0.0944 & 384.6810 $\pm$ 189.9644 & - \\
 baseline punishment & on &   -1.7770 $\pm$ 2.1333 &  996.0120 $\pm$ 39.0965 & 0.0010 $\pm$ 0.0316 &     0.1110 $\pm$ 0.7673 & 0.0623 $\pm$ 0.2595 \\
 & off &   -1.7795 $\pm$ 2.1777 &  996.9430 $\pm$ 31.9320 & 0.0 $\pm$ 0.0 &   18.1090 $\pm$ 21.9634 & - \\
   RTA no punishment & on & -56.7316 $\pm$ 18.9754 & 886.5070 $\pm$ 231.2872 & 0.0530 $\pm$ 0.2240 &   81.5040 $\pm$ 47.9373 & 1.0850 $\pm$ 0.1024 \\
   & off & -69.4304 $\pm$ 35.3073 & 701.8620 $\pm$ 334.5834 & 0.0010 $\pm$ 0.0316 & 489.0240 $\pm$ 244.4675 & - \\
      RTA punishment & on &   -2.3084 $\pm$ 2.7257 &  997.5530 $\pm$ 30.4823 & 0.0020 $\pm$ 0.0447 &     0.4140 $\pm$ 1.8554 & 0.1206 $\pm$ 0.3443 \\
      & off &   -2.5016 $\pm$ 3.2310 &  998.3320 $\pm$ 22.5890 & 0.0030 $\pm$ 0.0547 &   25.6890 $\pm$ 32.5553 & - \\
RTA Corrected Action & on & -54.9041 $\pm$ 21.5688 & 818.4390 $\pm$ 297.3325 & 0.0260 $\pm$ 0.1591 & 119.9790 $\pm$ 103.9273 & 1.1138 $\pm$ 0.0633 \\
& off & -52.5477 $\pm$ 28.3996 & 510.3110 $\pm$ 374.1590 & 0.0 $\pm$ 0.0 & 332.9860 $\pm$ 220.6356 & - \\
\bottomrule
\end{tabular}
}
\end{table}

\clearpage

\FloatBarrier \subsection{3D Spacecraft Docking Implicit ASIF}
\label{app:docking3dimplicitasif}

\begin{figure}[ht]
\centering
\subfigure[PPO evaluated with RTA Average Return]{\includegraphics[width=0.45\linewidth]{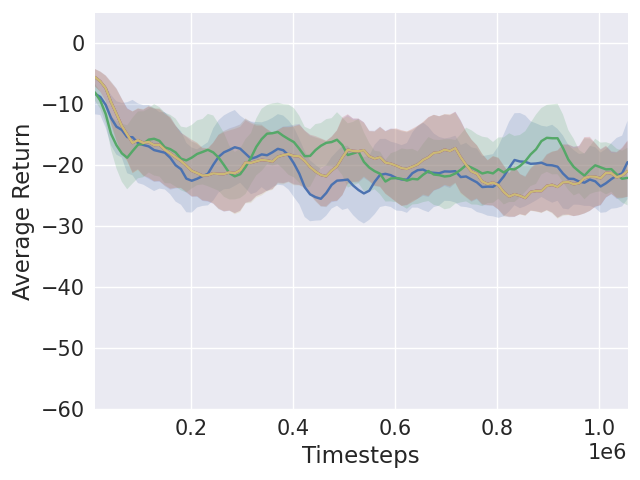}}\qquad
\subfigure[PPO evaluated with RTA Average Success]{\includegraphics[width=0.45\linewidth]{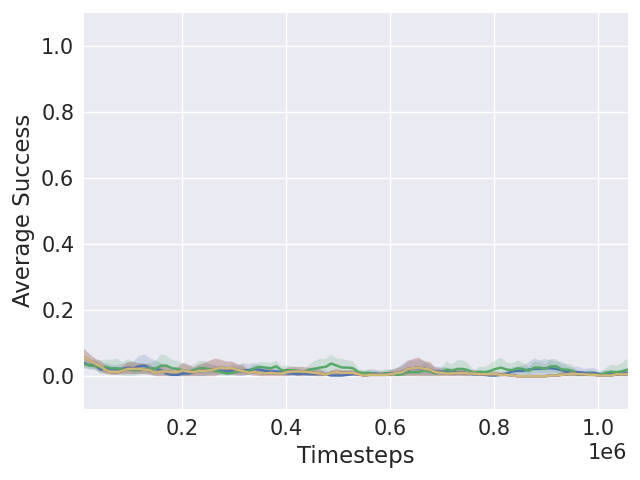}}\\
\subfigure[PPO evaluated without RTA Average Return]{\includegraphics[width=0.45\linewidth]{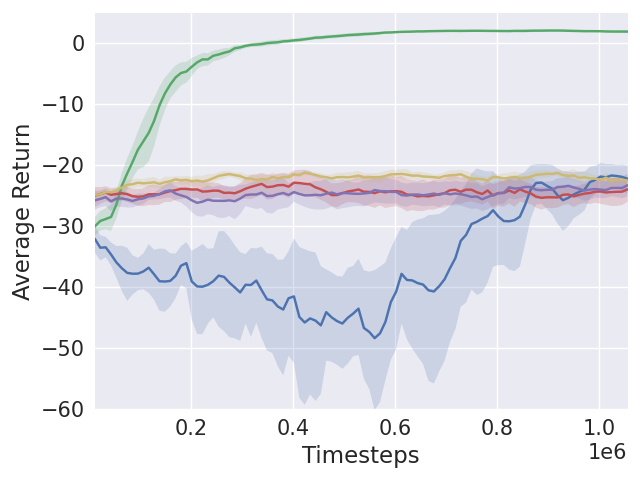}}\qquad
\subfigure[PPO evaluated without RTA Average Success]{\includegraphics[width=0.45\linewidth]{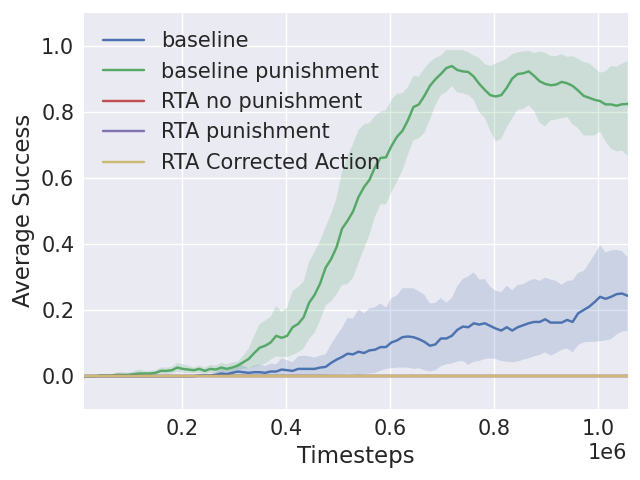}}
\caption{Results collected from experiments run in the Docking3D environment with an implicit ASIF RTA. Each curve represents the average 10 trials and the shaded region is the $95\%$ confidence interval about the mean. }
\label{fig:docking3d_ppo_imp_asif_results}
\end{figure}

\begin{table}[hb]
\caption{PPO 3D Spacecraft Docking Implicit ASIF}
\label{tab:ppo3dimplicitasif}
\centering
\resizebox{1.0\linewidth}{!}{
\begin{tabular}{lcccccc}
\toprule
       Configuration & RTA & Return & Length & Success &      Interventions/Violations &   Correction \\
\midrule
            baseline & on & -11.1833 $\pm$ 14.6859 & 670.6230 $\pm$ 303.9664 & 0.0260 $\pm$ 0.1591 & 670.6230 $\pm$ 303.9664 & 2.1712 $\pm$ 0.6998 \\
            & off & -21.7117 $\pm$ 8.4841 & 358.8620 $\pm$ 359.6777 & 0.2550 $\pm$ 0.4359 & 134.7320 $\pm$ 69.5541 & - \\
 baseline punishment & on & -11.2320 $\pm$ 15.6139 & 723.6400 $\pm$ 297.1046 & 0.0340 $\pm$ 0.1812 & 723.6400 $\pm$ 297.1046 & 2.0077 $\pm$ 0.7648 \\
 & off &   1.8275 $\pm$ 0.5826 & 740.5620 $\pm$ 177.9170 & 0.7960 $\pm$ 0.4030 &    1.2820 $\pm$ 2.5820 & - \\
   RTA no punishment & on & -10.4772 $\pm$ 14.6715 & 727.0780 $\pm$ 304.3238 & 0.0290 $\pm$ 0.1678 & 727.0770 $\pm$ 304.3229 & 0.9596 $\pm$ 0.1598 \\
   & off & -24.7680 $\pm$ 8.6542 &  101.9490 $\pm$ 38.2915 & 0.0 $\pm$ 0.0 &  89.9680 $\pm$ 33.4132 & - \\
      RTA punishment & on & -10.4772 $\pm$ 14.6715 & 727.0780 $\pm$ 304.3238 & 0.0290 $\pm$ 0.1678 & 727.0740 $\pm$ 304.3218 & 0.7655 $\pm$ 0.1558 \\
      & off & -24.4860 $\pm$ 8.6466 &   88.4590 $\pm$ 32.2835 & 0.0 $\pm$ 0.0 &  77.6300 $\pm$ 30.2741 & - \\
RTA Corrected Action & on & -10.4772 $\pm$ 14.6715 & 727.0780 $\pm$ 304.3238 & 0.0290 $\pm$ 0.1678 & 727.0780 $\pm$ 304.3238 & 5.3637 $\pm$ 2.2937 \\
& off & -21.8312 $\pm$ 8.9370 &    27.9090 $\pm$ 8.6748 & 0.0 $\pm$ 0.0 &   27.3040 $\pm$ 8.4486 & - \\
\bottomrule
\end{tabular}
}
\end{table}

\begin{figure}[ht]
\centering
\subfigure[SAC evaluated with RTA Average Return]{\includegraphics[width=0.45\linewidth]{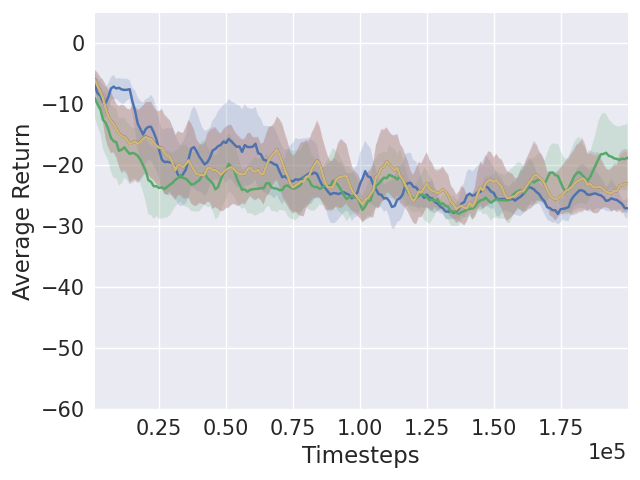}}\qquad
\subfigure[SAC evaluated with RTA Average Success]{\includegraphics[width=0.45\linewidth]{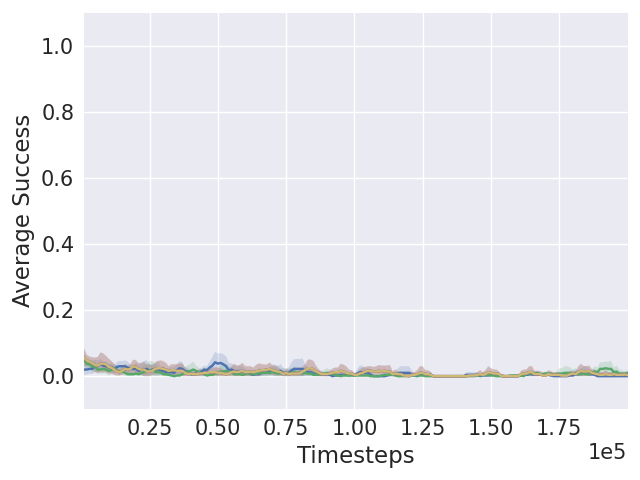}}\\
\subfigure[SAC evaluated without RTA Average Return]{\includegraphics[width=0.45\linewidth]{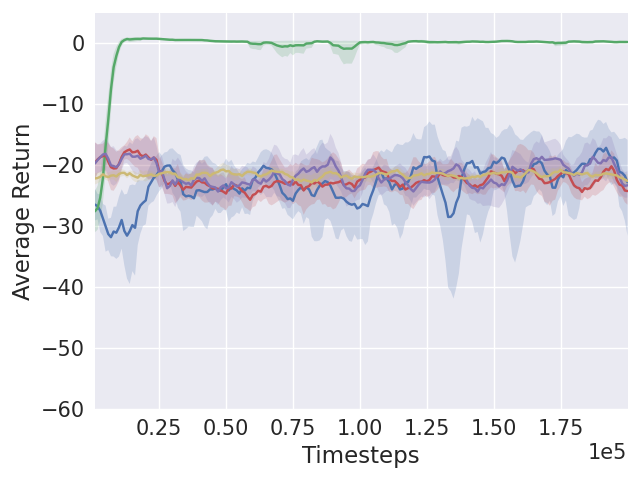}}\qquad
\subfigure[SAC evaluated without RTA Average Success]{\includegraphics[width=0.45\linewidth]{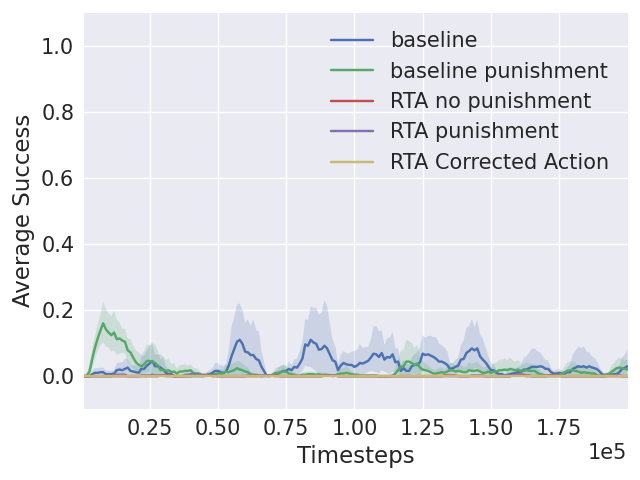}}
\caption{Results collected from experiments run in the Docking3D environment with an implicit ASIF RTA. Each curve represents the average 10 trials and the shaded region is the $95\%$ confidence interval about the mean.}
\label{fig:docking3d_sac_imp_asif_results}
\end{figure}

\begin{table}[hb]
\caption{SAC 3D Spacecraft Docking Implicit ASIF}
\label{tab:sac3dimplicitasif}
\centering
\resizebox{1.0\linewidth}{!}{
\begin{tabular}{lcccccc}
\toprule
       Configuration & RTA & Return & Length & Success &      Interventions/Violations &   Correction \\
\midrule
            baseline & on & -10.1428 $\pm$ 14.3804 & 721.7680 $\pm$ 289.5521 & 0.0350 $\pm$ 0.1838 & 721.7680 $\pm$ 289.5521 & 1.0391 $\pm$ 0.0697 \\
            & off & -49.4634 $\pm$ 24.9526 & 805.9750 $\pm$ 312.1980 & 0.0050 $\pm$ 0.0705 & 384.3810 $\pm$ 188.4363 & - \\
 baseline punishment & on & -13.8266 $\pm$ 16.6181 & 643.0790 $\pm$ 310.5455 & 0.0220 $\pm$ 0.1467 & 643.0790 $\pm$ 310.5455 & 1.0190 $\pm$ 0.0795 \\
 & off &   -3.7864 $\pm$ 5.3072 & 974.0050 $\pm$ 109.7366 & 0.0020 $\pm$ 0.0447 &   36.7950 $\pm$ 50.6100 & - \\
   RTA no punishment & on &  -9.2941 $\pm$ 13.4528 & 740.2390 $\pm$ 294.4016 & 0.0440 $\pm$ 0.2051 & 740.2390 $\pm$ 294.4016 & 0.9780 $\pm$ 0.0124 \\
   & off & -28.8167 $\pm$ 13.9128 & 233.3260 $\pm$ 108.2606 & 0.0 $\pm$ 0.0 &  180.0350 $\pm$ 88.3977 & - \\
      RTA punishment & on &  -9.2941 $\pm$ 13.4528 & 740.2390 $\pm$ 294.4016 & 0.0440 $\pm$ 0.2051 & 740.2390 $\pm$ 294.4016 & 0.9796 $\pm$ 0.0123 \\
      & off & -29.0804 $\pm$ 13.9550 & 232.2120 $\pm$ 107.8407 & 0.0 $\pm$ 0.0 &  180.9220 $\pm$ 87.4997 & - \\
RTA Corrected Action & on &  -9.2941 $\pm$ 13.4528 & 740.2390 $\pm$ 294.4016 & 0.0440 $\pm$ 0.2051 & 740.2390 $\pm$ 294.4016 & 1.6682 $\pm$ 0.1922 \\
& off &  -21.4260 $\pm$ 7.9279 &   48.9950 $\pm$ 12.3313 & 0.0 $\pm$ 0.0 &   46.4630 $\pm$ 11.9167 & - \\
\bottomrule
\end{tabular}
}
\end{table}

\clearpage